# Specific-to-General Learning for Temporal Events with Application to Learning Event Definitions from Video

**Alan Fern** AFERN@PURDUE.EDU
**Robert Givan** GIVAN@PURDUE.EDU
**Jeffrey Mark Siskind** QOBI@PURDUE.EDU
*School of Electrical and Computer Engineering*
*Purdue University, West Lafayette, IN 47907 USA*

## Abstract

We develop, analyze, and evaluate a novel, supervised, specific-to-general learner for a simple temporal logic and use the resulting algorithm to learn visual event definitions from video sequences. First, we introduce a simple, propositional, temporal, event-description language called AMA that is sufficiently expressive to represent many events yet sufficiently restrictive to support learning. We then give algorithms, along with lower and upper complexity bounds, for the subsumption and generalization problems for AMA formulas. We present a positive-examples–only specific-to-general learning method based on these algorithms. We also present a polynomial-time–computable "syntactic" subsumption test that implies semantic subsumption without being equivalent to it. A generalization algorithm based on syntactic subsumption can be used in place of semantic generalization to improve the asymptotic complexity of the resulting learning algorithm. Finally, we apply this algorithm to the task of learning relational event definitions from video and show that it yields definitions that are competitive with hand-coded ones.

## 1. Introduction

Humans conceptualize the world in terms of objects and events. This is reflected in the fact that we talk about the world using nouns and verbs. We perceive events taking place between objects, we interact with the world by performing events on objects, and we reason about the effects that actual and hypothetical events performed by us and others have on objects. We also *learn* new object and event types from novel experience. In this paper, we present and evaluate novel implemented techniques that allow a computer to learn new event types from examples. We show results from an application of these techniques to learning new event types from automatically constructed relational, force-dynamic descriptions of video sequences.

We wish the acquired knowledge of event types to support multiple modalities. Humans can observe someone *fax*ing a letter for the first time and quickly be able to recognize future occurrences of faxing, perform faxing, and reason about faxing. It thus appears likely that humans use and learn event representations that are sufficiently general to support fast and efficient use in multiple modalities. A long-term goal of our research is to allow similar cross-modal learning and use of event representations. We intend the same learned representations to be used for vision (as described in this paper), planning (something that we are beginning to investigate), and robotics (something left to the future).

A crucial requirement for event representations is that they capture the *invariants* of an event type. Humans classify both picking up a cup off a table and picking up a dumbbell off the floor as *picking up*. This suggests that human event representations are *relational*. We have an abstract relational notion of *picking up* that is parameterized by the participant objects rather than distinct propositional notions instantiated for specific objects. Humans also classify an event as *picking up* no matter whether the hand is moving slowly or quickly, horizontally or vertically, leftward or rightward, or along a straight path or circuitous one. It appears that it is not the characteristics of participant-object motion that distinguish *picking up* from other event types. Rather, it is the fact that the object being picked up changes from being supported by resting on its initial location to being supported by being grasped by the agent. This suggests that the primitive relations used to build event representations are *force dynamic* (Talmy, 1988).

Another desirable property of event representations is that they be *perspicuous*. Humans can introspect and describe the defining characteristics of event types. Such introspection is what allows us to create dictionaries. To support such introspection, we prefer a representation language that allows such characteristics to be explicitly manifest in event definitions and not emergent consequences of distributed parameters as in neural networks or hidden Markov models.

We develop a supervised learner for an event representation possessing these desired characteristics as follows. First, we present a simple, propositional, temporal logic called AMA that is a sublanguage of a variety of familiar temporal languages (e.g. linear temporal logic, or LTL Bacchus & Kabanza, 2000, event logic Siskind, 2001). This logic is expressive enough to describe a variety of interesting temporal events, but restrictive enough to support an effective learner, as we demonstrate below. We proceed to develop a specific-to-general learner for the AMA logic by giving algorithms and complexity bounds for the subsumption and generalization problems involving AMA formulas. While we show that semantic subsumption is intractable, we provide a weaker syntactic notion of subsumption that implies semantic subsumption but can be checked in polynomial time. Our implemented learner is based upon this syntactic subsumption.

We next show means to adapt this (propositional) AMA learner to learn relational concepts. We evaluate the resulting relational learner in a complete system for learning force-dynamic event definitions from positive-only training examples given as real video sequences. This is not the first system to perform visual-event recognition from video. We review prior work and compare it to the current work later in the paper. In fact, two such prior systems have been built by one of the authors. HOWARD (Siskind & Morris, 1996) learns to classify events from video using temporal, relational representations. But these representations are not force dynamic. LEONARD (Siskind, 2001) classifies events from video using temporal, relational, force-dynamic representations but does not learn these representations. It uses a library of hand-code representations. This work adds a learning component to LEONARD, essentially duplicating the performance of the hand-coded definitions automatically.

While we have demonstrated the utility of our learner in the visual-event–learning domain, we note that there are many domains where interesting concepts take the form of structured temporal sequences of events. In machine planning, macro-actions represent useful temporal patterns of action. In computer security, typical application behavior, represented perhaps as temporal patterns of system calls, must be differentiated from compromised application behavior (and likewise authorized-user behavior from intrusive behavior).

In what follows, Section 2 introduces our application domain of recognizing visual events and provides an informal description of our system for learning event definitions from video. Section 3 introduces the AMA language, syntax and semantics, and several concepts needed in our analysis of the language. Section 4 develops and analyzes algorithms for the subsumption and generalization problems in the language, and introduces the more practical notion of syntactic subsumption. Sec-tion 5 extends the basic propositional learner to handle relational data and negation, and to control exponential run-time growth. Section 6 presents our results on visual-event learning. Sections 7 and 8 compare to related work and conclude.

## 2. System Overview

This section provides an overview of our system for learning to recognize visual events from video. The aim is to provide an intuitive picture of our system before providing technical details. A formal presentation of our event-description language, algorithms, and both theoretical and empirical results appears in Sections 3–6. We first introduce the application domain of visual-event recognition and the LEONARD system, the event recognizer upon which our learner is built. Second, we describe how our positive-only learner fits into the overall system. Third, we informally introduce the AMA event-description language that is used by our learner. Finally, we give an informal presentation of the learning algorithm.

### 2.1 Recognizing Visual Events

LEONARD (Siskind, 2001) is a system for recognizing visual events from video camera input—an example of a simple visual event is "a hand picking up a block." This research was originally motivated by the problem of adding a learning component to LEONARD—allowing LEONARD to learn to recognize an event by viewing example events of the same type. Below, we give a high-level description of the LEONARD system.

LEONARD is a three-stage pipeline depicted in Figure 1. The raw input consists of a video-frame image sequence depicting events. First, a segmentation-and-tracking component transforms this input into a polygon movie: a sequence of frames, each frame being a set of convex polygons placed around the tracked objects in the video. Figure 2a shows a partial video sequence of a *pick up* event that is overlaid with the corresponding polygon movie. Next, a model-reconstruction component transforms the polygon movie into a force-dynamic model. This model describes the changing support, contact, and attachment relations between the tracked objects over time. Constructing this model is a somewhat involved process as described in Siskind (2000). Figure 2b shows a visual depiction of the force-dynamic model corresponding to the *pick up* event. Finally, an event-recognition component armed with a library of event definitions determines which events occurred in the model and, accordingly, in the video. Figure 2c shows the text output and input of the event-recognizer for the *pick up* event. The first line corresponds to the output which indicates the interval(s) during which a *pick up* occurred. The remaining lines are the text encoding of the event-recognizer input (model-reconstruction output), indicating the time intervals in which various force-dynamic relations are true in the video.

The event-recognition component of LEONARD represents event types with event-logic formulas like the following simplified example, representing $x$ picking up $y$ off of $z$.

$$\textsc{PickUp}(x, y, z) \stackrel{\triangle}{=} (\textsc{Supports}(z, y) \wedge \textsc{Contacts}(z, y)); (\textsc{Supports}(x, y) \wedge \textsc{Attached}(x, y))$$

This formula asserts that an event of $x$ picking up $y$ off of $z$ is defined as a sequence of two states where $z$ supports $y$ by way of contact in the first state and $x$ supports $y$ by way of attachment in the second state. SUPPORTS, CONTACTS, and ATTACHED are primitive force-dynamic relations. This formula is a specific example of the more general class of AMA formulas that we use in our learning.

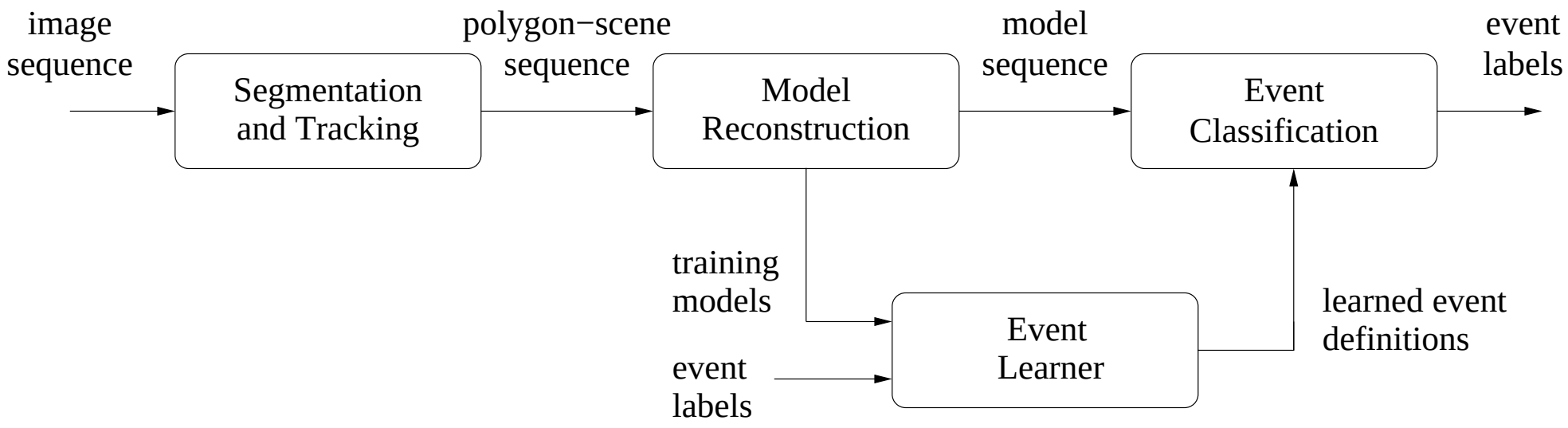

Figure 1: The upper boxes represent the three primary components of LEONARD's pipeline. The lower box depicts the event-learning component described in this paper. The input to the learning component consists of training models of target events (e.g., movies of *pick up* events) along with event labels (e.g., PICKUP(**hand**, **red**, **green**)) and the output is an event definition (e.g., a temporal logic formula defining PICKUP$(x, y, z)$).

### 2.2 Adding a Learning Component

Prior to the work reported in this paper, the definitions in LEONARD's event-recognition library were hand coded. Here, we add a learning component to LEONARD so that it can learn to recognize events. Figure 1 shows how the event learner fits into the overall system. The input to the event learner consists of force-dynamic models from the model-reconstruction stage, along with *event labels*, and its output consists of event definitions which are used by the event recognizer. We take a supervised-learning approach where the force-dynamic model-reconstruction process is applied to training videos of a target event type. The resulting force-dynamic models along with labels indicating the target event type are then given to the learner which induces a candidate definition of the event type.

For example, the input to our learner might consist of two models corresponding to two videos, one of a hand picking up a red block off of a green block with label PICKUP(**hand**, **red**, **green**) and one of a hand picking up a green block off of a red block with label PICKUP(**hand**, **green**, **red**)—the output would be a candidate definition of PICKUP$(x, y, z)$ that is applicable to previously unseen *pick up* events. Note that our learning component is positive-only in the sense that when learning a target event type it uses only positive training examples (where the target event occurs) and does not use negative examples (where the target event does not occur). The positive-only setting is of interest as it appears that humans are able to learn many event definitions given primarily or only positive examples. From a practical standpoint, a positive-only learner removes the often difficult task of collecting negative examples that are representative of what is not the event to be learned (e.g., what is a typical "non-pickup" event?).

The construction of our learner involves two primary design choices. First, we must choose an event representation language to serve as the learner's hypothesis space (i.e., the space of event definitions it may output). Second, we must design an algorithm for selecting a "good" event definition from the hypothesis space given a set of training examples of an event type.

### 2.3 The AMA Hypothesis Space

The full event logic supported by LEONARD is quite expressive, allowing the specification of a wide variety of temporal patterns (formulas). To help support successful learning, we use a more

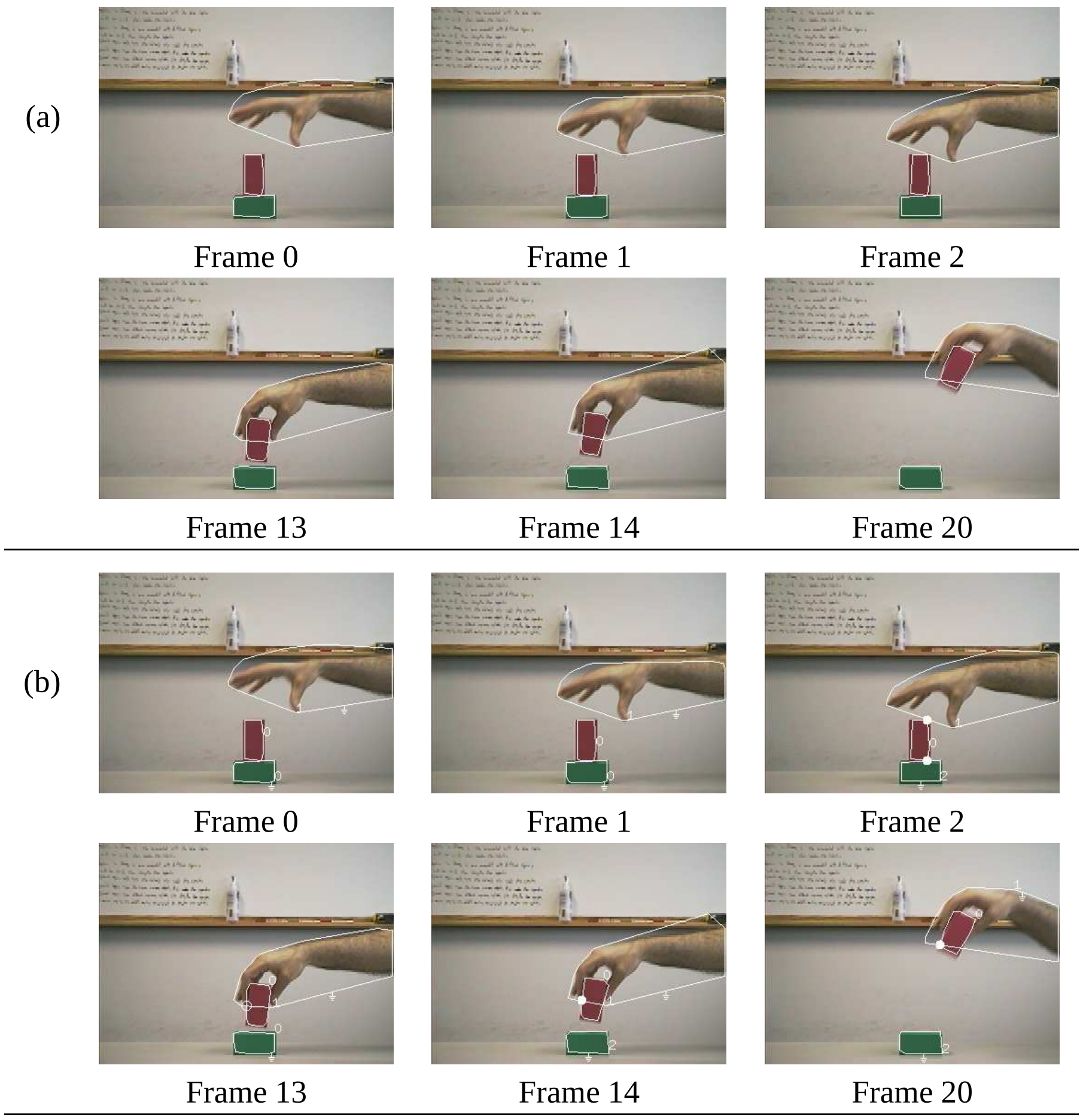

(c)

```
(PICK-UP HAND RED GREEN)@{[[0,1],[14,22])}

(SUPPORTED? RED)@{[[0:22])}
(SUPPORTED? HAND)@{[[1:13]), [[24:26])}
(SUPPORTS? RED HAND)@{[[1:13]), [[24:26])}
(SUPPORTS? HAND RED)@{[[13:22])}
(SUPPORTS? GREEN RED)@{[[0:14])}
(SUPPORTS? GREEN HAND)@{[[1:13])}
(CONTACTS? RED GREEN)@{[[0:2]), [[6:14])}
(ATTACHED? RED HAND)@{[[1:26])}
(ATTACHED? RED GREEN)@{[[1:6])}
```

Figure 2: LEONARD recognizes a *pick up* event. (a) Frames from the raw video input with the automatically generated polygon movie overlaid. (b) The same frames with a visual depiction of the automatically generated force-dynamic properties. (c) The text input/output of the event classifier corresponding to the depicted movie. The top line is the output and the remaining lines make up the input that encodes the changing force-dynamic properties. GREEN represents the block on the table and RED represents the block being picked up.

restrictive subset of event logic, called *AMA*, as our learner's hypothesis space. This subset excludes many practically useless formulas that may "confuse" the learner, while still retaining substantial expressiveness, thus allowing us to represent and learn many useful event types. Our restriction to AMA formulas is a form of syntactic learning bias.

The most basic AMA formulas are called *states* which express constant properties of time intervals of arbitrary duration. For example, $\textsc{Supports}(z, y) \wedge \textsc{Contacts}(z, y)$ is a state which tells us that $z$ must support and be in contact with $y$. In general, a state can be the conjunction of any number of *primitive propositions* (in this case force-dynamic relations). Using AMA we can also describe sequences of states. For example, $(\textsc{Supports}(z, y) \wedge \textsc{Contacts}(z, y))\ ;\ (\textsc{Supports}(x, y) \wedge \textsc{Attached}(x, y))$ is a sequence of two states, with the first state as given above and the second state indicating that $x$ must support and be attached to $y$. This formula is true whenever the first state is true for some time interval, followed immediately by the second state being true for some time interval "meeting" the first time interval. Such sequences are called *MA timelines* since they are the *M*eets of *A*nds. In general, MA timelines can contain any number of states. Finally, we can conjoin MA timelines to get AMA formulas (*A*nds of *MA*'s). For example, the AMA formula

$$[(\textsc{Supports}(z, y) \wedge \textsc{Contacts}(z, y)) ; (\textsc{Supports}(x, y) \wedge \textsc{Attached}(x, y))] \wedge$$
$$[(\textsc{Supports}(u, v) \wedge \textsc{Attached}(u, v)) ; (\textsc{Supports}(w, v) \wedge \textsc{Contacts}(w, v))]$$

defines an event where two MA timelines must be true simultaneously over the same time interval. Using AMA formulas we can represent events by listing various property sequences (MA timelines), all of which must occur in parallel as an event unfolds. It is important to note, however, that the transitions between states of different timelines in an AMA formula can occur in any relation to one another. For example, in the above AMA formula, the transition between the two states of the first timeline can occur before, after, or exactly at the transition between states of the second timeline.

An important assumption leveraged by our learner is that the primitive propositions used to construct states describe *liquid properties* (Shoham, 1987). For our purposes, we say that a property is liquid if when it holds over a time-interval it holds over all of its subintervals. The force-dynamic properties produced by LEONARD are liquid—e.g., if a hand SUPPORTS a block over an interval then clearly the hand supports the block over all subintervals. Because primitive propositions are liquid, properties described by states (conjunctions of primitives) are also liquid. However, properties described by MA and AMA formulas are not, in general, liquid.

### 2.4 Specific-to-General Learning from Positive Data

Recall that the examples that we wish to classify and learn from are force-dynamic models, which can be thought of (and are derived from) movies depicting temporal events. Also recall that our learner outputs definitions from the AMA hypothesis space. Given an AMA formula, we say that it *covers* an example model if it is true in that model. For a particular target event type (such as PICKUP), the ultimate goal is for the learner to output an AMA formula that covers an example model if and only if the model depicts an instance of the target event type. To understand our learner, it is useful to define a generality relationship between AMA formulas. We say that AMA formula $\Psi_1$ is more general (less specific) than AMA formula $\Psi_2$ if and only if $\Psi_2$ covers every example that $\Psi_1$ covers (and possibly more).[1]

1. In our formal analysis, we will use two different notions of generality (semantic and syntactic). In this section, we ignore such distinctions. We note, however, that the algorithm we informally describe later in this section is based on the syntactic notion of generality.

If the only learning goal is to find an AMA formula that is consistent with a set of positive-only training data, then one result can be the trivial solution of returning the formula that covers all examples. Rather than fix this problem by adding negative training examples (which will rule out the trivial solution), we instead change the learning goal to be that of finding the *least-general* formula that covers all of the positive examples.[2] This learning approach has been pursued for a variety of different languages within the machine-learning literature, including clausal first-order logic (Plotkin, 1971), definite clauses (Muggleton & Feng, 1992), and description logic (Cohen & Hirsh, 1994). It is important to choose an appropriate hypothesis space as a bias for this learning approach or the hypothesis returned may simply be (or resemble) one of two extremes, either the disjunction of the training examples or the universal hypothesis that covers all examples. In our experiments, we have found that, with enough training data, the least-general AMA formula often converges usefully.

We take a standard specific-to-general machine-learning approach to finding the least-general AMA formula that covers a set of positive examples. The approach relies on the computation of two functions: the least-general covering formula (LGCF) of an example model and the least-general generalization (LGG) of a set of AMA formulas. The LGCF of an example model is the least general AMA formula that covers the example. Intuitively, the LGCF is the AMA formula that captures the most information about the model. The LGG of any set of AMA formulas is the least-general AMA formula that is more general than each formula in the set. Intuitively, the LGG of a formula set is the AMA formula that captures the largest amount of common information among the formulas. Viewed differently, the LGG of a formula set covers all of the examples covered by those formulas, but covers as few other examples as possible (while remaining in AMA).[3]

The resulting specific-to-general learning approach proceeds as follows. First, use the LGCF function to transform each positive training model into an AMA formula. Second, return the LGG of the resulting formulas. The result represents the least-general AMA formula that covers all of the positive training examples. Thus, to specify our learner, all that remains is to provide algorithms for computing the LGCF and LGG for the AMA language. Below we informally describe our algorithms for computing these functions, which are formally derived and analyzed in Sections 3.4 and 4.

### 2.5 Computing the AMA LGCF

To increase the readability of our presentation, in what follows, we dispense with presenting examples where the primitive properties are meaningfully named force-dynamic relations. Rather, our examples will utilize abstract propositions such as $a$ and $b$. In our current application, these propositions correspond exclusively to force-dynamic properties, but may not for other applications. We now demonstrate how our system computes the LGCF of an example model.

Consider the following example model: $\{a@[1,4], b@[3,6], c@[6,6], d@[1,3], d@[5,6]\}$. Here, we take each number (1, …, 6) to represent a time interval of arbitrary (possibly varying with the number) duration during which nothing changes, and then each fact $p@[i,j]$ indicates that proposition $p$ is continuously true throughout the time intervals numbered $i$ through $j$. This model can be depicted graphically, as shown in Figure 3. The top four lines in the figure indicate the time

2. This avoids the need for negative examples and corresponds to finding the specific boundary of the version space (Mitchell, 1982).
3. The existence and uniqueness of the LGCF and LGG defined here is a formal property of the hypothesis space and is proven for AMA in Sections 3.4 and 4, respectively.

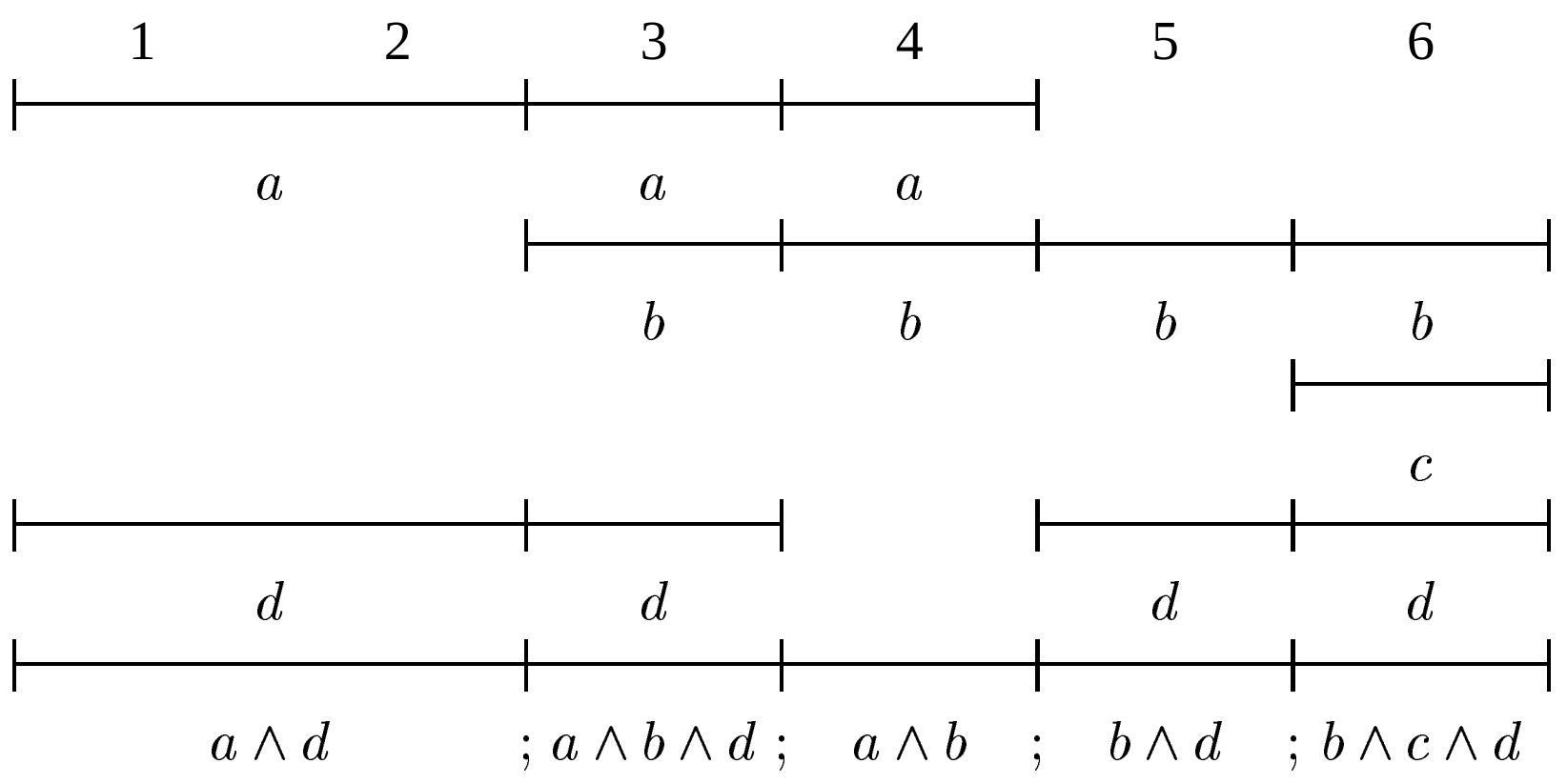

Figure 3: LGCF Computation. The top four horizontal lines of the figure indicate the intervals over which the propositions $a, b, c$ and $d$ are true in the model given by $\{a@[1,4], b@[3,6], c@[6,6], d@[1,3], d@[5,6]\}$. The bottom line shows how the model can be divided into intervals where no transitions occur. The LGCF is an MA timeline, shown at the bottom of the figure, with a state for each of the *no-transition* intervals. Each state simply contains the true propositions within the corresponding interval.

intervals over which each of the propositions $a, b, c$, and $d$ are true in the model. The bottom line in the figure shows how the model can be divided into five time intervals where no propositions change truth value. This division is possible because of the assumption that our propositions are liquid. This allows us, for example, to break up the time-interval where $a$ is true into three consecutive subintervals where $a$ is true. After dividing the model into intervals with no transitions, we compute the LGCF by simply treating each of those intervals as a state of an MA timeline, where the states contain only those propositions that are true during the corresponding time interval. The resulting five-state MA timeline is shown at the bottom of the figure. We show later that this simple computation returns the LGCF for any model. Thus, we see that the LGCF of a model is always an MA timeline.

### 2.6 Computing the AMA LGG

We now describe our algorithm for computing the LGG of two AMA formulas—the LGG of $m$ formulas can be computed via a sequence of $m - 1$ pairwise LGG applications, as discussed later.

Consider the two MA timelines: $\Phi_1 = (a \wedge b \wedge c); (b \wedge c \wedge d); e$ and $\Phi_2 = (a \wedge b \wedge e); a; (e \wedge d)$. It is useful to consider the various ways in which both timelines can be true simultaneously along an arbitrary time interval. To do this, we look at the various ways in which the two timelines can be aligned along a time interval. Figure 4a shows one of the many possible alignments of these timelines. We call such alignments *interdigitations*—in general, there are exponentially many interdigitations, each one ordering the state transitions differently. Note that an interdigitation is allowed to constrain two transitions from different timelines to occur simultaneously (though this is not depicted in the figure).[4]

4. Thus, an interdigitation provides an "ordering" relation on transitions that need not be anti-symmetric, but is reflexive, transitive, and total.

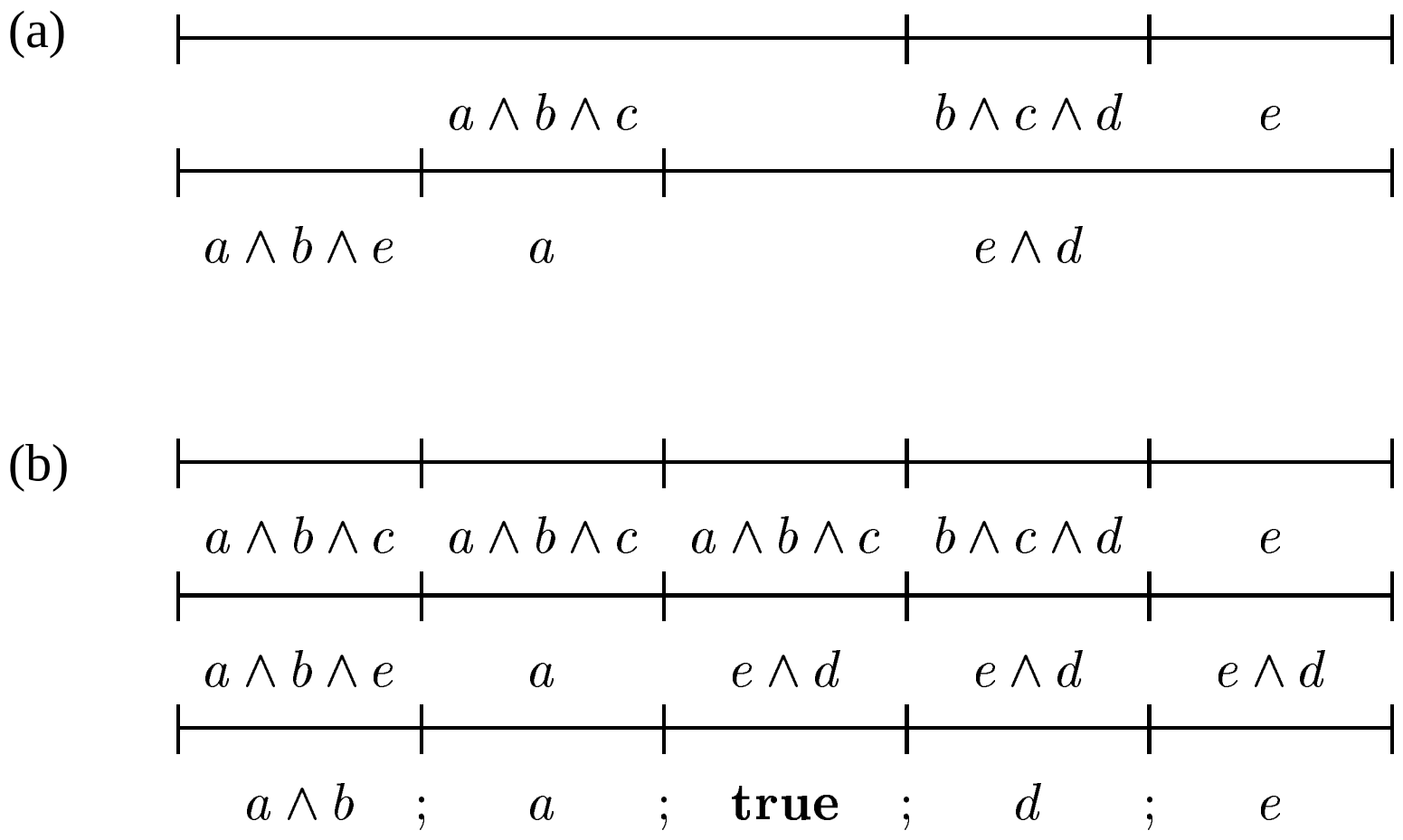

Figure 4: Generalizing the MA timelines $(a \wedge b \wedge c); (b \wedge c \wedge d); e$ and $(a \wedge b \wedge e); a; (e \wedge d)$. (a) One of the exponentially many interdigitations of the two timelines. (b) Computing the interdigitation generalization corresponding to the interdigitation from part (a). States are formed by intersecting aligned states from the two timelines. The state **true** represents a state with no propositions.

Given an interdigitation of two timelines, it is easy to construct a new MA timeline that must be true whenever either of the timelines is true (i.e., to construct a generalization of the two timelines). In Figure 4b, we give this construction for the interdigitation given in Figure 4a. The top two horizontal lines in the figure correspond to the interdigitation, only here we have divided every state on either timeline into two identical states, whenever a transition occurs during that state in the other timeline. The resulting pair of timelines have only simultaneous transitions and can be viewed as a sequence of state pairs, one from each timeline. The bottom horizontal line is then labeled by an MA timeline with one state for each such state pair, with that state being the intersection of the proposition sets in the state pair. Here, **true** represents the empty set of propositions, and is a state that is true anywhere.

We call the resulting timeline an *interdigitation generalization (IG)* of $\Phi_1$ and $\Phi_2$. It should be clear that this IG will be true whenever either $\Phi_1$ or $\Phi_2$ are true. In particular, if $\Phi_1$ holds along a time-interval in a model, then there is a sequence of consecutive (meeting) subintervals where the sequence of states in $\Phi_1$ are true. By construction, the IG can be aligned relative to $\Phi_1$ along the interval so that when we view states as sets, the states in the IG are subsets of the corresponding aligned state(s) in $\Phi_1$. Thus, the IG states are all true in the model under the alignment, showing that the IG is true in the model.

In general, there are exponentially many IGs of two input MA timelines, one for each possible interdigitation between the two. Clearly, since each IG is a generalization of the input timelines, then so is the conjunction of all the IGs. This conjunction is an AMA formula that generalizes the input MA timelines. In fact, we show later in the paper that this AMA formula is the LGG of the two timelines. Below we show the conjunction of all the IGs of $\Phi_1$ and $\Phi_2$ which serves as their LGG.

$$\begin{aligned}
&[(a \wedge b); b; e; \mathbf{true}; e] \wedge \\
&[(a \wedge b); b; \mathbf{true}; e] \wedge \\
&[(a \wedge b); b; \mathbf{true}; \mathbf{true}; e] \wedge \\
&[(a \wedge b); b; \mathbf{true}; e] \wedge \\
&[(a \wedge b); b; \mathbf{true}; d; e] \wedge \\
&[(a \wedge b); \mathbf{true}; \mathbf{true}; e] \wedge \\
&[(a \wedge b); \mathbf{true}; e] \wedge \\
&[(a \wedge b); \mathbf{true}; d; e] \wedge \\
&[(a \wedge b); a; \mathbf{true}; \mathbf{true}; e] \wedge \\
&[(a \wedge b); a; \mathbf{true}; e] \wedge \\
&[(a \wedge b); a; \mathbf{true}; d; e] \wedge \\
&[(a \wedge b); a; d; e] \wedge \\
&[(a \wedge b); a; \mathbf{true}; d; e]
\end{aligned}$$

While this formula is an LGG, it contains redundant timelines that can be pruned. First, it is clear that different IGs can result in the same MA timelines, and we can remove all but one copy of each timeline from the LGG. Second, note that if a timeline $\Phi'$ is more general than a timeline $\Phi$, then $\Phi \wedge \Phi'$ is equivalent to $\Phi$—thus, we can prune away timelines that are generalizations of others. Later in the paper, we show how to efficiently test whether one timeline is more general than another. After performing these pruning steps, we are left with only the first and next to last timelines in the above formula—thus, $[(a \wedge b); a; d; e] \wedge [(a \wedge b); b; e; \mathbf{true}; e]$ is an LGG of $\Phi_1$ and $\Phi_2$.

We have demonstrated how to compute the LGG of pairs of MA timelines. We can use this procedure to compute the LGG of pairs of AMA formulas. Given two AMA formulas we compute their LGG by simply conjoining the LGGs of all pairs of timelines (one from each AMA formula)—i.e., the formula

$$\bigwedge_i^m \bigwedge_j^n \mathrm{LGG}(\Phi_i, \Phi'_j)$$

is an LGG of the two AMA formulas $\Phi_1 \wedge \cdots \wedge \Phi_m$ and $\Phi'_1 \wedge \cdots \wedge \Phi'_n$, where the $\Phi_i$ and $\Phi'_j$ are MA timelines.

We have now informally described the LGCF and LGG operations needed to carry out the specific-to-general learning approach described above. In what follows, we more formally develop these operations and analyze the theoretical properties of the corresponding problems, then discuss the needed extensions to bring these (exponential, propositional, and negation-free) operations to practice.

## 3. Representing Events with AMA

Here we present a formal account of the AMA hypothesis space and an analytical development of the algorithms needed for specific-to-general learning for AMA. Readers that are primarily interested in a high-level view of the algorithms and their empirical evaluation may wish to skip Sections 3 and 4 and instead proceed directly to Sections 5 and 6, where we discuss several practical extensions to the basic learner and then present our empirical evaluation.

We study a subset of an interval-based logic called *event logic* (Siskind, 2001) utilized by LEONARD for event recognition in video sequences. This logic is interval-based in explicitly rep-resenting each of the possible interval relationships given originally by Allen (1983) in his calculus of interval relations (e.g., "overlaps," "meets," "during"). Event-logic formulas allow the definition of event types which can specify static properties of intervals directly and dynamic properties by hierarchically relating sub-intervals using the Allen relations. In this paper, the formal syntax and semantics of full event logic are needed only for Proposition 4 and are given in Appendix A.

Here we restrict our attention to a much simpler subset of event logic we call AMA, defined below. We believe that our choice of event logic rather than first-order logic, as well as our restriction to the AMA fragment of event logic, provide a useful learning bias by ruling out a large number of "practically useless" concepts while maintaining substantial expressive power. The practical utility of this bias is demonstrated via our empirical results in the visual-event–recognition application. AMA can also be seen as a restriction of LTL (Bacchus & Kabanza, 2000) to conjunction and "Until," with similar motivations. Below we present the syntax and semantics of AMA along with some of the key technical properties of AMA that will be used throughout this paper.

### 3.1 AMA Syntax and Semantics

It is natural to describe temporal events by specifying a sequence of properties that must hold over consecutive time intervals. For example, "a hand picking up a block" might become "the block is not supported by the hand and then the block is supported by the hand." We represent such sequences with *MA timelines*[5], which are sequences of conjunctive state restrictions. Intuitively, an MA timeline is given by a sequence of propositional conjunctions, separated by semicolons, and is taken to represent the set of events that temporally match the sequence of consecutive conjunctions. An AMA formula is then the conjunction of a number of MA timelines, representing events that can be simultaneously viewed as satisfying each of the conjoined timelines. Formally, the syntax of AMA formulas is given by,

$$\begin{aligned} \textit{state} &::= \mathtt{true} \mid \textit{prop} \mid \textit{prop} \wedge \textit{state} \\ \textit{MA} &::= (\textit{state}) \mid (\textit{state}); \textit{MA} \qquad \textit{// may omit parens} \\ \textit{AMA} &::= \textit{MA} \mid \textit{MA} \wedge \textit{AMA} \end{aligned}$$

where *prop* is any primitive proposition (sometimes called a primitive event type). We take this grammar to formally define the terms *MA timeline*, *MA formula*, *AMA formula*, and *state*. A $k$-MA formula is an MA formula with at most $k$ states, and a $k$-AMA formula is an AMA formula all of whose MA timelines are $k$-MA timelines. We often treat states as proposition sets with $\mathtt{true}$ the empty set and AMA formulas as MA-timeline sets. We may also treat MA formulas as sets of states—it is important to note, however, that MA formulas may contain duplicate states, and the duplication can be significant. For this reason, when treating MA timelines as sets, we formally intend sets of *state-index pairs* (where the index gives a states position in the formula). We do not indicate this explicitly to avoid encumbering our notation, but the implicit index must be remembered whenever handling duplicate states.

The semantics of AMA formulas is defined in terms of temporal models. A temporal model $\mathcal{M} = \langle M, I \rangle$ over the set PROP of propositions is a pair of a mapping $M$ from the natural numbers (representing time) to the truth assignments over PROP, and a closed natural-number interval $I$. We note that Siskind (2001) gives a continuous-time semantics for event logic where the models

5. MA stands for "Meets/And," an MA timeline being the "Meet" of a sequence of conjunctively restricted intervals.

are defined in terms of real-valued time intervals. The temporal models defined here use discrete natural-number time-indices. However, our results here still apply under the continuous-time semantics. (That semantics bounds the number of state changes in the continuous timeline to a countable number.) It is important to note that the natural numbers in the domain of $M$ are representing time discretely, but that there is no prescribed unit of continuous time represented by each natural number. Instead, each number represents an arbitrarily long period of continuous time during which nothing changed. Similarly, the states in our MA timelines represent arbitrarily long periods of time during which the conjunctive restriction given by the state holds. The satisfiability relation for AMA formulas is given as follows:

- A state $s$ is satisfied by a model $\langle M, I \rangle$ iff $M[x]$ assigns $P$ true for every $x \in I$ and $P \in s$.
- An MA timeline $s_1; s_2; \ldots; s_n$ is satisfied by a model $\langle M, [t, t'] \rangle$ iff there exists some $t''$ in $[t, t']$ such that $\langle M, [t, t''] \rangle$ satisfies $s_1$ and either $\langle M, [t'', t'] \rangle$ or $\langle M, [t''+1, t'] \rangle$ satisfies $s_2; \ldots; s_n$.
- An AMA formula $\Phi_1 \wedge \Phi_2 \wedge \cdots \wedge \Phi_n$ is satisfied by $\mathcal{M}$ iff each $\Phi_i$ is satisfied by $\mathcal{M}$.

The condition defining satisfaction for MA timelines may appear unintuitive at first due to the fact that there are two ways that $s_2; \ldots; s_n$ can be satisfied. The reason for this becomes clear by recalling that we are using the natural numbers to represent continuous time intervals. Intuitively, from a continuous-time perspective, an MA timeline is satisfied if there are consecutive continuous-time intervals satisfying the sequence of consecutive states of the MA timeline. The transition between consecutive states $s_i$ and $s_{i+1}$ can occur either within an interval of constant truth assignment (that happens to satisfy both states) or exactly at the boundary of two time intervals of constant truth value. In the above definition, these cases correspond to $s_2; \ldots; s_n$ being satisfied during the time intervals $[t'', t']$ and $[t''+1, t']$ respectively.

When $\mathcal{M}$ satisfies $\Phi$ we say that $\mathcal{M}$ *is a model of* $\Phi$ or that $\Phi$ *covers* $\mathcal{M}$. We say that AMA $\Psi_1$ *subsumes* AMA $\Psi_2$ iff every model of $\Psi_2$ is a model of $\Psi_1$, written $\Psi_2 \leq \Psi_1$, and we say that $\Psi_1$ *properly subsumes* $\Psi_2$, written $\Psi_2 < \Psi_1$, when we also have $\Psi_1 \not\leq \Psi_2$. Alternatively, we may state $\Psi_2 \leq \Psi_1$ by saying that $\Psi_1$ is *more general (or less specific) than* $\Psi_2$ or that $\Psi_1$ *covers* $\Psi_2$. Siskind (2001) provides a method to determine whether a given model satisfies a given AMA formula.

Finally, it will be useful to associate a distinguished MA timeline to a model. The *MA projection* of a model $\mathcal{M} = \langle M, [i, j] \rangle$ written as MAP($\mathcal{M}$) is an MA timeline $s_0; s_1; \ldots; s_{j-i}$ where state $s_k$ gives the true propositions in $M(i+k)$ for $0 \leq k \leq j - i$. Intuitively, the MA projection gives the sequence of propositional truth assignments from the beginning to the end of the model. Later we show that the MA projection of a model can be viewed as representing that model in a precise sense.

The following two examples illustrate some basic behaviors of AMA formulas:

**Example 1 (Stretchability).** *$S_1; S_2; S_3$, $S_1; S_2; S_2; \ldots; S_2; S_3$, and $S_1; S_1; S_1; S_2; S_3; S_3; S_3$ are all equivalent MA timelines. In general, MA timelines have the property that duplicating any state results in a formula equivalent to the original formula. Recall that, given a model $\langle M, I \rangle$, we view each truth assignment $M[x]$ as representing a continuous time-interval. This interval can conceptually be divided into an arbitrary number of subintervals. Thus if state $S$ is satisfied by $\langle M, [x, x] \rangle$, then so is the state sequence $S; S; \ldots; S$.*

**Example 2 (Infinite Descending Chains).** *Given propositions $A$ and $B$, the MA timeline $\Phi = (A \wedge B)$ is subsumed by each of the formulas $A; B,\ A; B; A; B,\ A; B; A; B; A; B,\ \ldots$. This is intuitively clear when our semantics are viewed from a continuous-time perspective. Any interval in which both $A$ and $B$ are true can be broken up into an arbitrary number of subintervals where both $A$ and $B$ hold. This example illustrates that there can be infinite descending chains of AMA formulas where the entire chain subsumes a given formula (but no member is equivalent to the given formula). In general, any AMA formula involving only the propositions $A$ and $B$ will subsume $\Phi$.*

### 3.2 Motivation for AMA

MA timelines are a very natural way to capture stretchable sequences of state constraints. But why consider the conjunction of such sequences, i.e., AMA? We have several reasons for this language enrichment. First of all, we show below that the AMA least-general generalization (LGG) is unique—this is not true for MA. Second, and more informally, we argue that parallel conjunctive constraints can be important to learning efficiency. In particular, the space of MA formulas of length $k$ grows in size exponentially with $k$, making it difficult to induce long MA formulas. However, finding several shorter MA timelines that each characterize *part* of a long sequence of changes is exponentially easier. (At least, the space to search is exponentially smaller.) The AMA conjunction of these timelines places these shorter constraints simultaneously and often captures a great deal of the concept structure. For this reason, we analyze AMA as well as MA and, in our empirical work, we consider $k$-AMA.

The AMA language is propositional. But our intended applications are relational, or first-order, including visual-event recognition. Later in this paper, we show that the propositional AMA learning algorithms that we develop can be effectively applied in relational domains. Our approach to first-order learning is distinctive in automatically constructing an object correspondence across examples (cf. Lavrac, Dzeroski, & Grobelnik, 1991; Roth & Yih, 2001). Similarly, though AMA does not allow for negative state constraints, in Section 5.4 we discuss how to extend our results to incorporate negation into our learning algorithms, which is crucial in visual-event recognition.

### 3.3 Conversion to First-Order Clauses

We note that AMA formulas can be translated in various ways into first-order clauses. It is not straightforward, however, to then use existing clausal generalization techniques for learning. In particular, to capture the AMA semantics in clauses, it appears necessary to define subsumption and generalization relative to a background theory that restricts us to a "continuous-time" first-order–model space.

For example, consider the AMA formulas $\Phi_1 = A \wedge B$ and $\Phi_2 = A; B$ where $A$ and $B$ are propositions—from Example 2 we know that $\Phi_1 \leq \Phi_2$. Now, consider a straightforward clausal translation of these formulas giving $C_1 = A(I) \wedge B(I)$ and $C_2 = A(I_1) \wedge B(I_2) \wedge \textsc{Meets}(I_1, I_2) \wedge I = \textsc{Span}(I_1, I_2)$, where the $I$ and $I_j$ are variables that represent time intervals, MEETS indicates that two time intervals meet each other, and SPAN is a function that returns a time interval equal to the union of its two time-interval arguments. The meaning we intend to capture is for satisfying assignments of $I$ in $C_1$ and $C_2$ to indicate intervals over which $\Phi_1$ and $\Phi_2$ are satisfied, respectively. It should be clear that, contrary to what we want, $C_1 \not\leq C_2$ (i.e., $\not\models C_1 \rightarrow C_2$), since it is easy to find unintended first-order models that satisfy $C_1$, but not $C_2$. Thus such a translation, and other similar translations, do not capture the continuous-time nature of the AMA semantics.

In order to capture the AMA semantics in a clausal setting, one might define a first-order theory that restricts us to continuous-time models—for example, allowing for the derivation "if property $B$ holds over an interval, then that property also holds over all sub-intervals." Given such a theory $\Sigma$, we have that $\Sigma \models C_1 \rightarrow C_2$, as desired. However, it is well known that least-general generalizations relative to such background theories need not exist (Plotkin, 1971), so prior work on clausal generalization does not simply subsume our results for the AMA language.

We note that for a particular training set, it may be possible to compile a continuous-time background theory $\Sigma$ into a finite but adequate set of ground facts. Relative to such ground theories, clausal LGGs are known to always exist and thus could be used for our application. However, the only such compiling approaches that look promising to us require exploiting an analysis similar to the one given in this paper—i.e., understanding the AMA generalization and subsumption problem separately from clausal generalization and exploiting that understanding in compiling the background theory. We have not pursued such compilations further.

Even if we are given such a compilation procedure, there are other problems with using existing clausal generalization techniques for learning AMA formulas. For the clausal translations of AMA we have found, the resulting generalizations typically fall outside of the (clausal translations of formulas in the) AMA language, so that the language bias of AMA is lost. In preliminary empirical work in our video-event recognition domain using clausal inductive-logic-programming (ILP) systems, we found that the learner appeared to lack the necessary language bias to find effective event definitions. While we believe that it would be possible to find ways to build this language bias into ILP systems, we chose instead to define and learn within the desired language bias directly, by defining the class of AMA formulas, and studying the generalization operation on that class.

### 3.4 Basic Concepts and Properties of AMA

We use the following convention in naming our results: "propositions" and "theorems" are the key results of our work, with theorems being those results of the most technical difficulty, and "lemmas" are technical results needed for the later proofs of propositions or theorems. We number all the results in one sequence, regardless of type. Proofs of theorems and propositions are provided in the main text—omitted proofs of lemmas are provided in the appendix.

We give pseudo-code for our methods in a non-deterministic style. In a non-deterministic language functions can return more than one value non-deterministically, either because they contain non-deterministic choice points, or because they call other non-deterministic functions. Since a non-deterministic function can return more than one possible value, depending on the choices made at the choice points encountered, specifying such a function is a natural way to specify a richly structured set (if the function has no arguments) or relation (if the function has arguments). To actually enumerate the values of the set (or the relation, once arguments are provided) one can simply use a standard backtracking search over the different possible computations corresponding to different choices at the choice points.

#### 3.4.1 Subsumption and Generalization for States

The most basic formulas we deal with are states (conjunctions of propositions). In our propositional setting computing subsumption and generalization at the state level is straightforward. A state $S_1$ subsumes $S_2$ ($S_2 \leq S_1$) iff $S_1$ is a subset of $S_2$, viewing states as sets of propositions. From this, we derive that the intersection of states is the least-general subsumer of those states and that the union of states is likewise the most general subsumee.

### 3.4.2 INTERDIGITATIONS

Given a set of MA timelines, we need to consider the different ways in which a model could simultaneously satisfy the timelines in the set. At the start of such a model (i.e., the first time point), the initial state from each timeline must be satisfied. At some time point in the model, one or more of the timelines can transition so that the second state in those timelines must be satisfied in place of the initial state, while the initial state of the other timelines remains satisfied. After a sequence of such transitions in subsets of the timelines, the final state of each timeline holds. Each way of choosing the transition sequence constitutes a different *interdigitation* of the timelines.

Viewed differently, each model simultaneously satisfying the timelines induces a *co-occurrence relation* on tuples of timeline states, one from each timeline, identifying which tuples co-occur at some point in the model. We represent this concept formally as a set of tuples of co-occurring states, i.e., a co-occurrence relation. We sometimes think of this set of tuples as ordered by the sequence of transitions. Intuitively, the tuples in an interdigitation represent the maximal time intervals over which no MA timeline has a transition, with those tuples giving the co-occurring states for each such time interval.

A relation $R$ on $X_1 \times \cdots \times X_n$ is *simultaneously consistent* with orderings $\leq_1$,...,$\leq_n$, if, whenever $R(x_1, \ldots, x_n)$ and $R(x'_1, \ldots, x'_n)$, either $x_i \leq_i x'_i$, for all $i$, or $x'_i \leq_i x_i$, for all $i$. We say $R$ is *piecewise total* if the projection of $R$ onto each component is total—i.e., every state in any $X_i$ appears in $R$.

**Definition 1 (Interdigitation).** *An interdigitation $I$ of a set $\{\Phi_1, \ldots, \Phi_n\}$ of MA timelines is a* co-occurrence *relation over $\Phi_1 \times \cdots \times \Phi_n$ (viewing timelines as sets of states*[6]*) that is piecewise total and simultaneously consistent with the state orderings of each $\Phi_i$. We say that two states $s \in \Phi_i$ and $s' \in \Phi_j$ for $i \neq j$ co-occur in $I$ iff some tuple of $I$ contains both $s$ and $s'$. We sometimes refer to $I$ as a sequence of tuples, meaning the sequence lexicographically ordered by the $\Phi_i$ state orderings.*

We note that there are exponentially many interdigitations of even two MA timelines (relative to the total number of states in the timelines). Example 3 on page 396 shows an interdigitation of two MA timelines. Pseudo-code for non-deterministically generating an arbitrary interdigitation for a set of MA timelines can be found in Figure 5. Given an interdigitation $I$ of the timelines $s_1; s_2; \ldots; s_m$ and $t_1; t_2; \ldots; t_n$ (and possibly others), the following basic properties of interdigitations are easily verifiable:

1. For $i < j$, if $s_i$ and $t_k$ co-occur in I then for all $k' < k$, $s_j$ does not co-occur with $t_{k'}$ in $I$.

2. $I(s_1, t_1)$ and $I(s_m, t_n)$.

We first use interdigitations to syntactically characterize subsumption between MA timelines.

**Definition 2 (Witnessing Interdigitation).** *An interdigitation $I$ of two MA timelines $\Phi_1$ and $\Phi_2$ is a* witness *to $\Phi_1 \leq \Phi_2$ iff for every pair of co-occurring states $s_1 \in \Phi_1$ and $s_2 \in \Phi_2$, we have that $s_2$ is a subset of $s_1$ (i.e., $s_1 \leq s_2$).*

The following lemma and proposition establish the equivalence between witnessing interdigitations and MA subsumption.

6. Recall, that, formally, MA timelines are viewed as sets of state-index pairs, rather than just sets of states. We ignore this distinction in our notation, for readability purposes, treating MA timelines as though no state is duplicated.

an-interdigitation$(\{\Phi_1, \Phi_2, \ldots, \Phi_n\})$<br>
 *// Input: MA timelines* $\Phi_1, \ldots, \Phi_n$<br>
 *// Output: an interdigitation of* $\{\Phi_1, \ldots, \Phi_n\}$<br>
 $S_0 := \langle \text{head}(\Phi_1), \ldots, \text{head}(\Phi_n) \rangle$;<br>
 **if** for all $1 \leq i \leq n$, $|\Phi_i| = 1$<br>
  **then return** $\langle S_0 \rangle$;<br>
 $T' := \{\Phi_i$ such that $|\Phi_i| > 1\}$;<br>
 $T'' :=$ a-non-empty-subset-of$(T')$;<br>
 **for** i := 1 **to** n<br>
  **if** $\Phi_i \in T''$<br>
   **then** $\Phi'_i := \text{rest}(\Phi_i)$<br>
   **else** $\Phi'_i := \Phi_i$;<br>
 **return** extend-tuple$(S_0,$ an-interdigitation$(\{\Phi'_1, \ldots, \Phi'_n\}))$;

Figure 5: Pseudo-code for an-interdigitation(), which non-deterministically computes an interdigitation for a set $\{\Phi_1, \ldots, \Phi_n\}$ of MA timelines. The function head($\Phi$) returns the first state in the timeline $\Phi$. rest($\Phi$) returns $\Phi$ with the first state removed. extend-tuple($x$,$I$) extends a tuple $I$ by adding a new first element $x$ to form a longer tuple. a-non-empty-subset-of($S$) non-deterministically returns an arbitrary non-empty subset of $S$.

**Lemma 1.** *For any MA timeline $\Phi$ and any model $\mathcal{M}$, if $\mathcal{M}$ satisfies $\Phi$, then there is a witnessing interdigitation for MAP$(\mathcal{M}) \leq \Phi$.*

**Proposition 2.** *For MA timelines $\Phi_1$ and $\Phi_2$, $\Phi_1 \leq \Phi_2$ iff there is an interdigitation that witnesses $\Phi_1 \leq \Phi_2$.*

**Proof**: We show the backward direction by induction on the number of states $n$ in timeline $\Phi_1$. If $n = 1$, then the existence of a witnessing interdigitation for $\Phi_1 \leq \Phi_2$ implies that every state in $\Phi_2$ is a subset of the single state in $\Phi_1$, and thus that any model of $\Phi_1$ is a model of $\Phi_2$ so that $\Phi_1 \leq \Phi_2$. Now, suppose for induction that the backward direction of the theorem holds whenever $\Phi_1$ has $n$ or fewer states. Given an arbitrary model $\mathcal{M}$ of an $n+1$ state $\Phi_1$ and an interdigitation $W$ that witnesses $\Phi_1 \leq \Phi_2$, we must show that $\mathcal{M}$ is also a model of $\Phi_2$ to conclude $\Phi_1 \leq \Phi_2$ as desired.

Write $\Phi_1$ as $s_1; \ldots; s_{n+1}$ and $\Phi_2$ as $t_1; \ldots; t_m$. As a witnessing interdigitation, $W$ must identify some maximal prefix $t_1; \ldots; t_{m'}$ of $\Phi_2$ made up of states that co-occur with $s_1$ and thus that are subsets of $s_1$. Since $\mathcal{M} = \langle M, [t, t'] \rangle$ satisfies $\Phi_1$, by definition there must exist a $t'' \in [t, t']$ such that $\langle M, [t, t''] \rangle$ satisfies $s_1$ (and thus $t_1; \ldots; t_{m'}$) and $\langle M, I' \rangle$ satisfies $s_2; \ldots; s_{n+1}$ for $I'$ equal to either $[t'', t']$ or $[t''+1, t']$. In either case, it is straightforward to construct, from $W$, a witnessing interdigitation for $s_2; \ldots; s_{n+1} \leq t_{m'+1}; \ldots; t_m$ and use the induction hypothesis to then show that $\langle M, I' \rangle$ must satisfy $t_{m'+1}; \ldots; t_m$. It follows that $\mathcal{M}$ satisfies $\Phi_2$ as desired.

For the forward direction, assume that $\Phi_1 \leq \Phi_2$, and let $\mathcal{M}$ be any model such that $\Phi_1 =$ MAP$(\mathcal{M})$. It is clear that such an $\mathcal{M}$ exists and satisfies $\Phi_1$. It follows that $\mathcal{M}$ satisfies $\Phi_2$. Lemma 1 then implies that there is a witnessing interdigitation for MAP$(\mathcal{M}) \leq \Phi_2$ and thus for $\Phi_1 \leq \Phi_2$. $\square$

#### 3.4.3 LEAST-GENERAL COVERING FORMULA

A logic can discriminate two models if it contains a formula that satisfies one but not the other. It turns out that AMA formulas can discriminate two models exactly when much richer *internal positive event logic* (IPEL) formulas can do so. Internal formulas are those that define event occurrence only in terms of properties within the defining interval. That is, satisfaction by $\langle M, I\rangle$ depends only on the proposition truth values given by $M$ inside the interval $I$. Positive formulas are those that do not contain negation. Appendix A gives the full syntax and semantics of IPEL (which are used only to state and prove Lemma 3 ). The fact that AMA can discriminate models as well as IPEL indicates that our restriction to AMA formulas retains substantial expressive power and leads to the following result which serves as the least-general covering formula (LGCF) component of our specific-to-general learning procedure. Formally, an LGCF of model $\mathcal{M}$ within a formula language $\mathcal{L}$ (e.g. AMA or IPEL) is a formula in $\mathcal{L}$ that covers $\mathcal{M}$ such that no other covering formula in $\mathcal{L}$ is strictly less general. Intuitively, the LGCF of a model, if unique, is the "most representative" formula of that model. Our analysis uses the concept of *model embedding*. We say that model $\mathcal{M}$ embeds model $\mathcal{M}'$ iff MAP$(\mathcal{M}) \leq$ MAP$(\mathcal{M}')$.

**Lemma 3.** *For any $E \in IPEL$, if model $\mathcal{M}$ embeds a model that satisfies $E$, then $\mathcal{M}$ satisfies $E$.*

**Proposition 4.** *The MA projection of a model is its LGCF for internal positive event logic (and hence for AMA), up to semantic equivalence.*

**Proof**: Consider model $\mathcal{M}$. We know that MAP$(\mathcal{M})$ covers $\mathcal{M}$, so it remains to show that MAP$(\mathcal{M})$ is the least general formula to do so, up to semantic equivalence.

Let $E$ be any IPEL formula that covers $\mathcal{M}$. Let $\mathcal{M}'$ be any model that is covered by MAP$(\mathcal{M})$—we want to show that $E$ also covers $\mathcal{M}'$. We know, from Lemma 1, that there is a witnessing interdigitation for MAP$(\mathcal{M}') \leq$ MAP$(\mathcal{M})$. Thus, by Proposition 2, MAP$(\mathcal{M}') \leq$ MAP$(\mathcal{M})$ showing that $\mathcal{M}'$ embeds $\mathcal{M}$. Combining these facts with Lemma 3 it follows that $E$ also covers $\mathcal{M}'$ and hence MAP$(\mathcal{M}) \leq E$. □

Proposition 4 tells us that, for IPEL, the LGCF of a model exists, is unique, and is an MA timeline. Given this property, when an AMA formula $\Psi$ covers all the MA timelines covered by another AMA formula $\Psi'$, we have $\Psi' \leq \Psi$. Thus, for the remainder of this paper, when considering subsumption between formulas, we can abstract away from temporal models and deal instead with MA timelines. Proposition 4 also tells us that we can compute the LGCF of a model by constructing the MA projection of that model. Based on the definition of MA projection, it is straightforward to derive an LGCF algorithm which runs in time polynomial in the size of the model[7]. We note that the MA projection may contain repeated states. In practice, we remove repeated states, since this does not change the meaning of the resulting formula (as described in Example 1).

#### 3.4.4 COMBINING INTERDIGITATION WITH GENERALIZATION OR SPECIALIZATION

Interdigitations are useful in analyzing both conjunctions and disjunctions of MA timelines. When conjoining a set of timelines, any model of the conjunction induces an interdigitation of the timelines such that co-occurring states simultaneously hold in the model at some point (viewing states as sets, the the states resulting from unioning co-occurring states must hold). By constructing an

7. We take the size of a model $\mathcal{M} = \langle M, I\rangle$ to be the sum over $x \in I$ of the number of true propositions in $M(x)$.

interdigitation and taking the union of each tuple of co-occurring states to get a sequence of states, we get an MA timeline that forces the conjunction of the timelines to hold. We call such a sequence an *interdigitation specialization* of the timelines. Dually, an *interdigitation generalization* involving intersections of states gives an MA timeline that holds whenever the disjunction of a set of timelines holds.

**Definition 3.** *An interdigitation generalization (specialization) of a set $\Sigma$ of MA timelines is an MA timeline $s_1; \ldots; s_m$, such that, for some interdigitation $I$ of $\Sigma$ with $m$ tuples, $s_j$ is the intersection (respectively, union) of the components of the j'th tuple of the sequence $I$. The set of interdigitation generalizations (respectively, specializations) of $\Sigma$ is called* $\text{IG}(\Sigma)$ *(respectively,* $\text{IS}(\Sigma)$*).*

**Example 3.** *Suppose that $s_1, s_2, s_3, t_1, t_2$, and $t_3$ are each sets of propositions (i.e., states). Consider the timelines $S = s_1; s_2; s_3$ and $T = t_1; t_2; t_3$. The relation*

$$\{ \ \langle s_1, t_1 \rangle, \langle s_2, t_1 \rangle, \langle s_3, t_2 \rangle, \langle s_3, t_3 \rangle \ \}$$

*is an interdigitation of $S$ and $T$ in which states $s_1$ and $s_2$ co-occur with $t_1$, and $s_3$ co-occurs with $t_2$ and $t_3$. The corresponding* IG *and* IS *members are*

$$\begin{array}{lcl} s_1 \cap t_1;\ s_2 \cap t_1;\ s_3 \cap t_2;\ s_3 \cap t_3 & \in & \text{IG}(\{S, T\}) \\ s_1 \cup t_1;\ s_2 \cup t_1;\ s_3 \cup t_2;\ s_3 \cup t_3 & \in & \text{IS}(\{S, T\}). \end{array}$$

*If $t_1 \subseteq s_1, t_1 \subseteq s_2, t_2 \subseteq s_3$, and $t_3 \subseteq s_3$, then the interdigitation* witnesses $S \leq T$.

Each timeline in $\text{IG}(\Sigma)$ (dually, $\text{IS}(\Sigma)$) subsumes (is subsumed by) each timeline in $\Sigma$—this is easily verified using Proposition 2. For our complexity analyses, we note that the number of states in any member of $\text{IG}(\Sigma)$ or $\text{IS}(\Sigma)$ is bounded from below by the number of states in any of the MA timelines in $\Sigma$ and is bounded from above by the total number of states in all the MA timelines in $\Sigma$. The number of interdigitations of $\Sigma$, and thus of members of $\text{IG}(\Sigma)$ or $\text{IS}(\Sigma)$, is exponential in that same total number of states. The algorithms that we present later for computing LGGs require the computation of both $\text{IG}(\Sigma)$ and $\text{IS}(\Sigma)$. Here we give pseudo-code to compute these quantities. Figure 6 gives pseudo-code for the function an-IG-member that non-deterministically computes an arbitrary member of $\text{IG}(\Sigma)$ (an-IS-member is the same, except that we replace intersection by union). Given a set $\Sigma$ of MA timelines we can compute $\text{IG}(\Sigma)$ by executing all possible deterministic computation paths of the function call an-IG-member($\Sigma$), i.e., computing the set of results obtainable from the non-deterministic function for all possible decisions at non-deterministic choice points.

We now give a useful lemma and a proposition concerning the relationships between conjunctions and disjunctions of MA concepts (the former being AMA concepts). For convenience here, we use disjunction on MA concepts, producing formulas outside of AMA with the obvious interpretation.

**Lemma 5.** *Given an MA formula $\Phi$ that subsumes each member of a set $\Sigma$ of MA formulas, $\Phi$ also subsumes some member $\Phi'$ of $\text{IG}(\Sigma)$. Dually, when $\Phi$ is subsumed by each member of $\Sigma$, we have that $\Phi$ is also subsumed by some member $\Phi'$ of $\text{IS}(\Sigma)$. In each case, the length of $\Phi'$ is bounded by the size of $\Sigma$.*

```
an-IG-member({Φ1, Φ2, ..., Φn})
    // Input: MA timelines Φ1, ..., Φn
    // Output: a member of IG({Φ1, Φ2, ..., Φn})
    return map (intersect-tuple, an-interdigitation ({Φ1, ..., Φn}));
```

Figure 6: Pseudo-code for an-IG-member, which non-deterministically computes a member of IG($T$) where $T$ is a set of MA timelines. The function intersect-tuple($I$) takes a tuple $I$ of sets as its argument and returns their intersection. The higher-order function map($f, I$) takes a function $f$ and a tuple $I$ as arguments and returns a tuple of the same length as $I$ obtained by applying $f$ to each element of $I$ and making a tuple of the results.

**Proposition 6.** *The following hold:*

1. *(and-to-or) The conjunction of a set $\Sigma$ of MA timelines equals the disjunction of the timelines in* $\text{IS}(\Sigma)$.

2. *(or-to-and) The disjunction of a set $\Sigma$ of MA timelines is subsumed by the conjunction of the timelines in* $\text{IG}(\Sigma)$.

**Proof**: To prove or-to-and, recall that, for any $\Phi \in \Sigma$ and any $\Phi' \in \text{IG}(\Sigma)$, we have that $\Phi \leq \Phi'$. From this it is immediate that $(\bigvee \Sigma) \leq (\bigwedge \text{IG}(\Sigma))$. Using a dual argument, we can show that $(\bigvee \text{IS}(\Sigma)) \leq (\bigwedge \Sigma)$. It remains to show that $(\bigwedge \Sigma) \leq (\bigvee \text{IS}(\Sigma))$, which is equivalent to showing that any timeline subsumed by $(\bigwedge \Sigma)$ is also subsumed by $(\bigvee \text{IS}(\Sigma))$ (by Proposition 4). Consider any MA timeline $\Phi$ such that $\Phi \leq (\bigwedge \Sigma)$—this implies that each member of $\Sigma$ subsumes $\Phi$. Lemma 5 then implies that there is some $\Phi' \in \text{IS}(\Sigma)$ such that $\Phi \leq \Phi'$. From this we get that $\Phi \leq (\bigvee \text{IS}(\Sigma))$ as desired. □

Using and-to-or, we can now reduce AMA subsumption to MA subsumption, with an exponential increase in the problem size.

**Proposition 7.** *For AMA $\Psi_1$ and $\Psi_2$, $\Psi_1 \leq \Psi_2$ if and only if for all $\Phi_1 \in \text{IS}(\Psi_1)$ and $\Phi_2 \in \Psi_2, \Phi_1 \leq \Phi_2$.*

**Proof**: For the forward direction we show the contrapositive. Assume there is a $\Phi_1 \in \text{IS}(\Psi_1)$ and a $\Phi_2 \in \Psi_2$ such that $\Phi_1 \not\leq \Phi_2$. Thus, there is an MA timeline $\Phi$ such that $\Phi \leq \Phi_1$ but $\Phi \not\leq \Phi_2$. This tells us that $\Phi \leq (\bigvee \text{IS}(\Psi_1))$ and that $\Phi \not\leq \Psi_2$, thus $(\bigvee \text{IS}(\Psi_1)) \not\leq \Psi_2$ and by "and-to-or" we get that $\Psi_1 \not\leq \Psi_2$.

For the backward direction assume that for all $\Phi_1 \in \text{IS}(\Psi_1)$ and $\Phi_2 \in \Psi_2$ that $\Phi_1 \leq \Phi_2$. This tells us that for each $\Phi_1 \in \text{IS}(\Psi_1)$, that $\Phi_1 \leq \Psi_2$—thus, $\Psi_1 = (\bigvee \text{IS}(\Psi_1)) \leq \Psi_2$. □

## 4. Subsumption and Generalization

In this section we study subsumption and generalization of AMA formulas. First, we give a polynomial-time algorithm for deciding subsumption between MA formulas and then show that deciding subsumption for AMA formulas is coNP-complete. Second we give algorithms and complexity bounds for the construction of least-general generalization (LGG) formulas based on our

MA-subsumes $(\Phi_1, \Phi_2)$
*// Input:* $\Phi_1 = s_1; \ldots; s_m$ *and* $\Phi_2 = t_1; \ldots; t_n$
*// Output:* $\Phi_1 \leq \Phi_2$

1. **if** there is a path from $v_{1,1}$ to $v_{m,n}$ in $SG(\Phi_1, \Phi_2)$ **then return** TRUE. For example,
   (a) Create an array Reachable($i$,$j$) of boolean values, all FALSE, for $0 \leq i \leq m$ and $0 \leq j \leq n$.
   (b) **for** $i := 1$ **to** $m$, Reachable$(i, 0)$ := TRUE;
   **for** $j := 1$ **to** $n$, Reachable$(0, j)$ := TRUE;
   **for** $i := 1$ **to** $m$
       **for** $j := 1$ **to** $n$
           Reachable$(i, j)$ := $(t_i \subseteq s_j \wedge ($Reachable$(i-1, j) \vee$
               Reachable$(i, j-1) \vee$
               Reachable$(i-1, j-1))$;
   (c) **if** Reachable$(m, n)$ **then return** TRUE;
2. Otherwise, **return** FALSE;

Figure 7: Pseudo-code for the MA subsumption algorithm. $SG(\Phi_1, \Phi_2)$ is the subsumption graph defined in the main text.

analysis of subsumption, including existence, uniqueness, lower/upper bounds, and an algorithm for the LGG on AMA formulas. Third, we introduce a polynomial-time–computable syntactic notion of subsumption and an algorithm that computes the corresponding syntactic LGG that is exponentially faster than our semantic LGG algorithm. Fourth, in Section 4.4, we give a detailed example showing the steps performed by our LGG algorithms to compute the semantic and syntactic LGGs of two AMA formulas.

### 4.1 Subsumption

All our methods rely critically on a novel algorithm for deciding the subsumption question $\Phi_1 \leq \Phi_2$ between MA formulas $\Phi_1$ and $\Phi_2$ in polynomial-time. We note that merely searching the possible interdigitations of $\Phi_1$ and $\Phi_2$ for a witnessing interdigitation provides an obvious decision procedure for the subsumption question—however, there are, in general, exponentially many such interdigitations. We reduce the MA subsumption problem to finding a path in a graph on pairs of states in $\Phi_1 \times \Phi_2$, a polynomial-time operation. Pseudo-code for the resulting MA subsumption algorithm is shown in Figure 7. The main data structure used by the MA subsumption algorithm is the subsumption graph.

**Definition 4.** *The subsumption graph of two MA timelines* $\Phi_1 = s_1; \cdots; s_m$ *and* $\Phi_2 = t_1; \cdots; t_n$ *(written* $SG(\Phi_1, \Phi_2)$*) is a directed graph* $G = \langle V, E \rangle$ *with* $V = \{v_{i,j} \mid 1 \leq i \leq m, 1 \leq j \leq n\}$*. The (directed) edge set* $E$ *equals* $\{\langle v_{i,j}, v_{i',j'} \rangle \mid s_i \leq t_j,\ s_{i'} \leq t_{j'},\ i \leq i' \leq i+1, j \leq j' \leq j+1\}$.

To achieve a polynomial-time bound one can simply use any polynomial-time pathfinding algorithm. In our case the special structure of the subsumption graph can be exploited to determine if the desired path exists in $O(mn)$ time, as the example method shown in the pseudo-code illustrates. The following theorem asserts the correctness of the algorithm assuming a correct polynomial-time path-finding method is used.

**Lemma 8.** *Given MA timelines $\Phi_1 = s_1; \ldots; s_m$ and $\Phi_2 = t_1; \ldots; t_n$, there is a witnessing interdigitation for $\Phi_1 \leq \Phi_2$ iff there is a path in the subsumption graph $SG(\Phi_1, \Phi_2)$ from $v_{1,1}$ to $v_{m,n}$.*

**Theorem 9.** *Given MA timelines $\Phi_1$ and $\Phi_2$, MA-subsumes($\Phi_1, \Phi_2$) decides $\Phi_1 \leq \Phi_2$ in polynomial time.*

**Proof**: The algorithm clearly runs in polynomial time. Lemma 8 tells us that line 2 of the algorithm will return TRUE iff there is a witnessing interdigitation. Combining this with Proposition 2 shows that the algorithm returns TRUE iff $\Phi_1 \leq \Phi_2$. □

Given this polynomial-time algorithm for MA subsumption, Proposition 7 immediately suggests an exponential-time algorithm for deciding AMA subsumption—by computing MA subsumption between the exponentially many IS timelines of one formula and the timelines of the other formula. Our next theorem suggests that we cannot do any better than this in the worst case—we argue that AMA subsumption is coNP-complete by reduction from boolean satisfiability. Readers uninterested in the technical details of this argument may skip directly to Section 4.2.

To develop a correspondence between boolean satisfiability problems, which include negation, and AMA formulas, which lack negation, we imagine that each boolean variable has two AMA propositions, one for "true" and one for "false." In particular, given a boolean satisfiability problem over $n$ variables $p_1, \ldots, p_n$, we take the set PROP$_n$ to be the set containing $2n$ AMA propositions True$_k$ and False$_k$ for each $k$ between 1 and $n$. We can now represent a truth assignment $A$ to the $p_i$ variables with an AMA state $s_A$ given as follows:

$$s_A = \{\text{True}_i \mid 1 \leq i \leq n,\ A(p_i) = \text{true}\} \cup \{\text{False}_i \mid 1 \leq i \leq n,\ A(p_i) = \text{false}\}$$

As Proposition 7 suggests, checking AMA subsumption critically involves the exponentially many interdigitation specializations of the timelines of one of the AMA formulas. In our proof, we design an AMA formula whose interdigitation specializations can be seen to correspond to truth assignments[8] to boolean variables, as shown in the following lemma.

**Lemma 10.** *Given some $n$, let $\Psi$ be the conjunction of the timelines*

$$\bigcup_{i=1}^{n} \{(\mathit{PROP}_n; \mathit{True}_i; \mathit{False}_i; \mathit{PROP}_n), (\mathit{PROP}_n; \mathit{False}_i; \mathit{True}_i; \mathit{PROP}_n)\}.$$

*We have the following facts about truth assignments to the Boolean variables $p_1, \ldots, p_n$:*

1. *For any truth assignment $A$, $\mathit{PROP}_n; s_A; \mathit{PROP}_n$ is semantically equivalent to a member of* $\text{IS}(\Psi)$.

2. *For each $\Phi \in \text{IS}(\Psi)$ there is a truth assignment $A$ such that $\Phi \leq \mathit{PROP}_n; s_A; \mathit{PROP}_n$.*

8. A truth assignment is a function mapping boolean variables to true or false.

With this lemma in hand, we can now tackle the complexity of AMA subsumption.

**Theorem 11.** *Deciding AMA subsumption is coNP-complete.*

**Proof**: We first show that deciding the AMA-subsumption of $\Psi_1$ by $\Psi_2$ is in coNP by providing a polynomial-length certificate for any "no" answer. This certificate for non-subsumption is an interdigitation of the timelines of $\Psi_1$ that yields a member of IS$(\Psi_1)$ not subsumed by $\Psi_2$. Such a certificate can be checked in polynomial time: given such an interdigitation, the corresponding member of IS$(\Psi_1)$ can be computed in time polynomial in the size of $\Psi_1$, and we can then test whether the resulting timeline is subsumed by each timeline in $\Psi_2$ using the polynomial-time MA-subsumption algorithm. Proposition 7 guarantees that $\Psi_1 \not\leq \Psi_2$ iff there is a timeline in IS$(\Psi_1)$ that is not subsumed by every timeline in $\Psi_2$, so that such a certificate will exist exactly when the answer to a subsumption query is "no."

To show coNP-hardness we reduce the problem of deciding the satisfiability of a 3-SAT formula $S = C_1 \wedge \cdots \wedge C_m$ to the problem of recognizing non-subsumption between AMA formulas. Here, each $C_i$ is $(l_{i,1} \vee l_{i,2} \vee l_{i,3})$ and each $l_{i,j}$ either a proposition $p$ chosen from $P = \{p_1, \ldots, p_n\}$ or its negation $\neg p$. The idea of the reduction is to construct an AMA formula $\Psi$ for which we view the exponentially many members of IS$(\Psi)$ as representing truth assignments. We then construct an MA timeline $\Phi$ that we view as representing $\neg S$ and show that $S$ is satisfiable iff $\Psi \not\leq \Phi$.

Let $\Psi$ be as defined in Lemma 10. Let $\Phi$ be the formula $s_1; \ldots; s_m$, where

$$\begin{aligned} s_i \quad &= \quad \{\text{False}_j \mid l_{i,k} = p_j \text{ for some } k\} \cup \\ &\qquad \{\text{True}_j \mid l_{i,k} = \neg p_j \text{ for some } k\}. \end{aligned}$$

Each $s_i$ can be thought of as asserting "not $C_i$." We start by showing that if $S$ is satisfiable then $\Psi \not\leq \Phi$. Assume that $S$ is satisfied via a truth assignment $A$—we know from Lemma 10 that there is a $\Phi' \in$ IS$(\Psi)$ that is semantically equivalent to PROP$_n; s_A;$ PROP$_n$. We show that PROP$_n; s_A;$ PROP$_n$ is not subsumed by $\Phi$, to conclude $\Psi \not\leq \Phi$ using Proposition 7, as desired. Suppose for contradiction that PROP$_n; s_A;$ PROP$_n$ is subsumed by $\Phi$—then the state $s_A$ must be subsumed by some state $s_i$ in $\Phi$. Consider the corresponding clause $C_i$ of $S$. Since $A$ satisfies $S$ we have that $C_i$ is satisfied and at least one of its literals $l_{i,k}$ must be true. Assume that $l_{i,k} = p_j$ (a dual argument holds for $l_{i,k} = \neg p_j$), then we have that $s_i$ contains False$_j$ while $s_A$ contains True$_j$ but not False$_j$—thus, we have that $s_A \not\leq s_i$ (since $s_i \not\subseteq s_A$), contradicting our choice of $i$.

To complete the proof, we now assume that $S$ is unsatisfiable and show that $\Psi \leq \Phi$. Using Proposition 7, we consider arbitrary $\Phi'$ in IS$(\Psi)$—we will show that $\Phi' \leq \Phi$. From Lemma 10 we know there is some truth assignment $A$ such that $\Phi' \leq$ PROP$_n; s_A;$ PROP$_n$. Since $S$ is unsatisfiable we know that some $C_i$ is not satisfied by $A$ and hence $\neg C_i$ is satisfied by $A$. This implies that each primitive proposition in $s_i$ is in $s_A$. Let $W$ be the following interdigitation between $T =$ PROP$_n; s_A;$ PROP$_n$ and $\Phi = s_1; \ldots; s_m$:

$$\{\langle \text{PROP}_n, s_1\rangle \, \langle \text{PROP}_n, s_2\rangle \cdots \langle \text{PROP}_n, s_i\rangle \, \langle s_A, s_i\rangle \, \langle \text{PROP}_n, s_i\rangle \, \langle \text{PROP}_n, s_{i+1}\rangle \cdots \langle \text{PROP}_n, s_m\rangle\}$$

We see that in each tuple of co-occurring states given above that the state from $T$ is subsumed by the state from $\Phi$. Thus $W$ is a witnessing interdigitation to PROP$_n; s_A;$ PROP$_n \leq \Phi$, which then holds by Proposition 2—combining this with $\Phi' \leq$ PROP$_n; s_A;$ PROP$_n$ we get that $\Phi' \leq \Phi$. □

Given this hardness result we later define a weaker polynomial-time–computable subsumption notion for use in our learning algorithms.

### 4.2 Least-General Generalization.

An AMA LGG of a set of AMA formulas is an AMA formula that is more general than each formula in the set and not strictly more general than any other such formula. The existence of an AMA LGG is nontrivial as there can be infinite chains of increasingly specific formulas all of which generalize given formulas. Example 2 demonstrated such chains for an MA subsumee and can be extended for AMA subsumees. For example, each member of the chain $P;Q,\ P;Q;P;Q,\ P;Q;P;Q;P;Q,\ \ldots$ covers $\Psi_1 = (P \wedge Q);Q$ and $\Psi_2 = P;(P \wedge Q)$. Despite such complications, the AMA LGG does exist.

**Theorem 12.** *There is an LGG for any finite set $\Sigma$ of AMA formulas that is subsumed by all other generalizations of $\Sigma$.*

**Proof**: Let $\Gamma$ be the set $\bigcup_{\Psi' \in \Sigma} \mathrm{IS}(\Psi')$. Let $\Psi$ be the conjunction of all the MA timelines that generalize $\Gamma$ while having size no larger than $\Gamma$. Since there are only a finite number of primitive propositions, there are only a finite number of such timelines, so $\Psi$ is well defined[9]. We show that $\Psi$ is a least-general generalization of $\Sigma$. First, note that each timeline in $\Psi$ generalizes $\Gamma$ and thus $\Sigma$ (by Proposition 6), so $\Psi$ must generalize $\Sigma$. Now, consider arbitrary generalization $\Psi'$ of $\Sigma$. Proposition 7 implies that $\Psi'$ must generalize each formula in $\Gamma$. Lemma 5 then implies that each timeline of $\Psi'$ must subsume a timeline $\Phi$ that is no longer than the size of $\Gamma$ and that also subsumes the timelines of $\Gamma$. But then $\Phi$ must be a timeline of $\Psi$, by our choice of $\Psi$, so that every timeline of $\Psi'$ subsumes a timeline of $\Psi$. It follows that $\Psi'$ subsumes $\Psi$, and that $\Psi$ is an LGG of $\Sigma$ subsumed by all other LGGs of $\Sigma$, as desired. □

Given that the AMA LGG exists and is unique we now show how to compute it. Our first step is to strengthen "or-to-and" from Proposition 6 to get an LGG for the MA sublanguage.

**Theorem 13.** *For a set $\Sigma$ of MA formulas, the conjunction of all MA timelines in* $\mathrm{IG}(\Sigma)$ *is an AMA LGG of $\Sigma$.*

**Proof**: Let $\Psi$ be the specified conjunction. Since each timeline of $\mathrm{IG}(\Sigma)$ subsumes all timelines in $\Sigma$, $\Psi$ subsumes each member of $\Sigma$. To show $\Psi$ is a least-general such formula, consider an AMA formula $\Psi'$ that also subsumes all members of $\Sigma$. Since each timeline of $\Psi'$ must subsume all members of $\Sigma$, Lemma 5 implies that each timeline of $\Psi'$ subsumes a member of $\mathrm{IG}(\Sigma)$ and thus each timeline of $\Psi'$ subsumes $\Psi$. This implies $\Psi \leq \Psi'$. □

We can now characterize the AMA LGG using IS and IG.

**Theorem 14.** $\mathrm{IG}(\bigcup_{\Psi \in \Sigma} \mathrm{IS}(\Psi))$ *is an AMA LGG of the set $\Sigma$ of AMA formulas.*

**Proof**: Let $\Sigma = \{\Psi_1, \ldots, \Psi_n\}$ and $E = \Psi_1 \vee \cdots \vee \Psi_n$. We know that the AMA LGG of $\Sigma$ must subsume $E$, or it would fail to subsume one of the $\Psi_i$. Using "and-to-or" we can represent $E$ as a disjunction of MA timelines given by $E = (\bigvee \mathrm{IS}(\Psi_1)) \vee \cdots \vee (\bigvee \mathrm{IS}(\Psi_n))$. Any AMA LGG must be a least-general formula that subsumes $E$—i.e., an AMA LGG of the set of MA timelines $\bigcup\{\mathrm{IS}(\Psi) | \Psi \in \Sigma\}$. Theorem 13 tells us that an LGG of these timelines is given by $\mathrm{IG}(\bigcup\{\mathrm{IS}(\Psi) | \Psi \in \Sigma\})$. □

9. There must be at least one such timeline, the timeline where the only state is $\mathbf{true}$

```
1:  semantic-LGG({Ψ1, Ψ2, ..., Ψm})
2:      // Input: AMA formulas Ψ1, ..., Ψm
3:      // Output: LGG of {Ψ1, ..., Ψm}
4:      S := {};
5:      for i := 1 to m
6:          for each Φ in all-values(an-IS-member(Ψi))
7:              if (∀Φ′ ∈ S . Φ ≰ Φ′)
8:                  then S′ := {Φ″ ∈ S | Φ″ ≤ Φ};
9:                       S := (S − S′) ∪ {Φ};

10:     G := {};
11:     for each Φ in all-values(an-IG-member(S))
12:         if (∀Φ′ ∈ G . Φ′ ≰ Φ)
13:             then G′ := {Φ″ ∈ G | Φ ≤ Φ″};
14:                  G := (G − G′) ∪ {Φ};

15:     return (⋀ G)
```

Figure 8: Pseudo-code for computing the semantic AMA LGG of a set of AMA formulas.

Theorem 14 leads directly to an algorithm for computing the AMA LGG—Figure 8 gives pseudo-code for the computation. Lines 4-9 of the pseudo-code correspond to the computation of $\bigcup\{\text{IS}(\Psi) | \Psi \in \Sigma\}$, where timelines are not included in the set if they are subsumed by timelines already in the set (which can be checked with the polynomial time MA subsumption algorithm). This pruning, accomplished by the **if** test in line 7, often drastically reduces the size of the timeline set for which we perform the subsequent IG computation—the final result is not affected by the pruning since the subsequent IG computation is a generalization step. The remainder of the pseudo-code corresponds to the computation of $\text{IG}(\bigcup\{\text{IS}(\Psi) | \Psi \in \Sigma\})$ where we do not include timelines in the final result that subsume some other timeline in the set. This pruning step (the **if** test in line 12) is sound since when one timeline subsumes another, the conjunction of those timelines is equivalent to the most specific one. Section 4.4.1 traces the computations of this algorithm for an example LGG calculation.

Since the sizes of both $\text{IS}(\cdot)$ and $\text{IG}(\cdot)$ are exponential in the sizes of their inputs, the code in Figure 8 is doubly exponential in the input size. We conjecture that we cannot do better than this, but we have not yet proven a doubly exponential lower bound for the AMA case. When the input formulas are MA timelines the algorithm takes singly exponential time, since $\text{IS}(\{\Phi\}) = \Phi$ when $\Phi$ is in MA. We now prove an exponential lower bound when the input formulas are in MA. Again, readers uninterested in the technical details of this proof can safely skip forward to Section 4.3.

For this argument, we take the available primitive propositions to be those in the set $\{p_{i,j} \mid 1 \leq i \leq n,\ 1 \leq j \leq n\}$, and consider the MA timelines

$$\begin{aligned} \Phi_1 &= s_{1,*}; s_{2,*}; \ldots; s_{n,*} \\ \text{and} \quad \Phi_2 &= s_{*,1}; s_{*,2}; \ldots; s_{*,n}, \text{ where} \end{aligned}$$

$$\begin{aligned} s_{i,*} &= p_{i,1} \wedge \cdots \wedge p_{i,n} \\ \text{and} \quad s_{*,j} &= p_{1,j} \wedge \cdots \wedge p_{n,j}. \end{aligned}$$

We will show that any AMA LGG of $\Phi_1$ and $\Phi_2$ must contain an exponential number of timelines. In particular, we will show that any AMA LGG is equivalent to the conjunction of a subset of $\text{IG}(\{\Phi_1, \Phi_2\})$, and that certain timelines may not be omitted from such a subset.

**Lemma 15.** *Any AMA LGG $\Psi$ of a set $\Sigma$ of MA timelines is equivalent to a conjunction $\Psi'$ of timelines from* $\text{IG}(\Sigma)$ *with* $|\Psi'| \leq |\Psi|$

**Proof**: Lemma 5 implies that any timeline $\Phi$ in $\Psi$ must subsume some timeline $\Phi' \in \text{IG}(\Sigma)$. But then the conjunction $\Psi'$ of such $\Phi'$ must be equivalent to $\Psi$, since it clearly covers $\Sigma$ and is covered by the LGG $\Psi$. Since $\Psi'$ was formed by taking one timeline from $\text{IG}(\Sigma)$ for each timeline in $\Psi$, we have $|\Psi'| \leq |\Psi|$. □ We can complete our argument then by showing that exponentially many timelines in $\text{IG}(\{\Phi_1, \Phi_2\})$ cannot be omitted from such a conjunction while it remains an LGG.

Notice that for any $i, j$ we have that $s_{i,*} \cap s_{*,j} = p_{i,j}$. This implies that any state in $\text{IG}(\{\Phi_1, \Phi_2\})$ contains exactly one proposition, since each such state is formed by intersecting a state from $\Phi_1$ and $\Phi_2$. Furthermore, the definition of interdigitation, applied here, implies the following two facts for any timeline $q_1; q_2; \ldots; q_m$ in $\text{IG}(\{\Phi_1, \Phi_2\})$:

1. $q_1 = p_{1,1}$ and $q_m = p_{n,n}$.
2. For consecutive states $q_k = p_{i,j}$ and $q_{k+1} = p_{i',j'}$, $i'$ is either $i$ or $i+1$, $j'$ is either $j$ or $j+1$, and not both $i = i'$ and $j = j'$.

Together these facts imply that any timeline in $\text{IG}(\{\Phi_1, \Phi_2\})$ is a sequence of propositions starting with $p_{1,1}$ and ending with $p_{n,n}$ such that any consecutive propositions $p_{i,j}; p_{i',j'}$ are different with $i'$ equal to $i$ or $i+1$ and $j'$ equal to $j$ or $j+1$. We call a timeline in $\text{IG}(\{\Phi_1, \Phi_2\})$ *square* if and only if each pair of consecutive propositions $p_{i,j}$ and $p_{i',j'}$ have either $i' = i$ or $j' = j$. The following lemma implies that no square timeline can be omitted from the conjunction of timelines in $\text{IG}(\Phi_1, \Phi_2)$ if it is to remain an LGG of $\Phi_1$ and $\Phi_2$.

**Lemma 16.** *Let $\Phi_1$ and $\Phi_2$ be as given above and let $\Psi = \bigwedge \text{IG}(\{\Phi_1, \Phi_2\})$. For any $\Psi'$ whose timelines are a subset of those in $\Psi$ that omits some square timeline, we have $\Psi < \Psi'$.*

The number of square timelines in $\text{IG}(\{\Phi_1, \Phi_2\})$ is equal to $\frac{(2n-2)!}{(n-1)!(n-1)!}$ and hence is exponential in the size of $\Phi_1$ and $\Phi_2$. We have now completed the proof of the following result.

**Theorem 17.** *The smallest LGG of two MA formulas can be exponentially large.*

**Proof**: By Lemma 15, any AMA LGG $\Psi'$ of $\Phi_1$ and $\Phi_2$ is equivalent to a conjunction of the same number of timelines chosen from $\text{IG}(\{\Phi_1, \Phi_2\})$. However, by Lemma 16, any such conjunction must have at least $\frac{(2n-2)!}{(n-1)!(n-1)!}$ timelines, and then so must $\Psi'$, which must then be exponentially large. □

**Conjecture 18.** *The smallest LGG of two AMA formulas can be doubly-exponentially large.*

We now show that our lower-bound on AMA LGG complexity is not merely a consequence of the existence of large AMA LGGs. Even when there is a small LGG, it can be expensive to compute due to the difficulty of testing AMA subsumption:

**Theorem 19.** *Determining whether a formula $\Psi$ is an AMA LGG for two given AMA formulas $\Psi_1$ and $\Psi_2$ is co-NP-hard, and is in co-NEXP, in the size of all three formulas together.*

**Proof**: To show co-NP-hardness we use a straightforward reduction from AMA subsumption. Given two AMA formulas $\Psi_1$ and $\Psi_2$ we decide $\Psi_1 \leq \Psi_2$ by asking whether $\Psi_2$ is an AMA LGG of $\Psi_1$ and $\Psi_2$. Clearly $\Psi_1 \leq \Psi_2$ iff $\Psi_2$ is an LGG of the two formulas.

To show the co-NEXP upper bound, note that we can check in exponential time whether $\Psi_1 \leq \Psi$ and $\Psi_2 \leq \Psi$ using Proposition 7 and the polynomial-time MA subsumption algorithm. It remains to show that we can check whether $\Psi$ is *not* the "least" subsumer. Since Theorem 14 shows that the LGG of $\Psi_1$ and $\Psi_2$ is $\text{IG}(\text{IS}(\Psi_1) \cup \text{IS}(\Psi_2))$, if $\Psi$ is not the LGG then $\Psi \not\leq \text{IG}(\text{IS}(\Psi_1) \cup \text{IS}(\Psi_2))$. Thus, by Proposition 7, if $\Psi$ is not a least subsumer, there must be timelines $\Phi_1 \in \text{IS}(\Psi)$ and $\Phi_2 \in \text{IG}(\text{IS}(\Psi_1) \cup \text{IS}(\Psi_2))$ such that $\Phi_1 \not\leq \Phi_2$. We can then use exponentially long certificates for "No" answers: each certificate is a pair of an interdigitation $I_1$ of $\Psi$ and an interdigitation $I_2$ of $\text{IS}(\Psi_1) \cup \text{IS}(\Psi_2)$, such that the corresponding members $\Phi_1 \in \text{IS}(\Psi)$ and $\Phi_2 \in \text{IG}(\text{IS}(\Psi_1) \cup \text{IS}(\Psi_2))$ have $\Phi_1 \not\leq \Phi_2$. Given the pair of certificates $I_1$ and $I_2$, $\Phi_1$ can be computed in polynomial time, $\Phi_2$ can be computed in exponential time, and the subsumption between them can be checked in polynomial time (relative to their size, which can be exponential). If $\Psi$ is the LGG then $\Psi \leq \text{IG}(\text{IS}(\Psi_1) \cup \text{IS}(\Psi_2))$, so that no such certificates will exist. □

### 4.3 Syntactic Subsumption and Syntactic Least-General Generalization.

Given the intractability results for semantic AMA subsumption, we now introduce a tractable generality notion, syntactic subsumption, and discuss the corresponding LGG problem. The use of syntactic forms of generality for efficiency is familiar in ILP (Muggleton & De Raedt, 1994)—where, for example, $\theta$-subsumption is often used in place of the entailment generality relation. Unlike AMA semantic subsumption, syntactic subsumption requires checking only polynomially many MA subsumptions, each in polynomial time (via Theorem 9).

**Definition 5.** *AMA $\Psi_1$ is syntactically subsumed by AMA $\Psi_2$ (written $\Psi_1 \leq_{syn} \Psi_2$) iff for each MA timeline $\Phi_2 \in \Psi_2$, there is an MA timeline $\Phi_1 \in \Psi_1$ such that $\Phi_1 \leq \Phi_2$.*

**Proposition 20.** *AMA syntactic subsumption can be decided in polynomial time.*

Syntactic subsumption trivially implies semantic subsumption—however, the converse does not hold in general. Consider the AMA formulas $(A; B) \wedge (B; A)$, and $A; B; A$ where $A$ and $B$ are primitive propositions. We have $(A; B) \wedge (B; A) \leq A; B; A$; however, we have neither $A; B \leq A; B; A$ nor $B; A \leq A; B; A$, so that $A; B; A$ does not syntactically subsume $(A; B) \wedge (B; A)$. Syntactic subsumption fails to recognize constraints that are only derived from the interaction of timelines within a formula.

**Syntactic Least-General Generalization.** A *syntactic AMA LGG* is a syntactically least-general AMA formula that syntactically subsumes the input AMA formulas. Here, "least" means that no formula properly syntactically subsumed by a syntactic LGG can syntactically subsume the input formulas. Based on the hardness gap between syntactic and semantic AMA subsumption, one might conjecture that a similar gap exists between the syntactic and semantic LGG problems. Proving such a gap exists requires closing the gap between the lower and upper bounds on AMA LGG shown in Theorem 14 in favor of the upper bound, as suggested by Conjecture 18. While we cannot yet show a hardness gap between semantic and syntactic LGG, we do give a syntactic LGG algorithm that is exponentially more efficient than the best semantic LGG algorithm we have found (that of Theorem 14). First, we show that syntactic LGGs exist and are unique up to mutual syntactic subsumption (and hence up to semantic equivalence).

**Theorem 21.** *There exists a syntactic LGG for any AMA formula set $\Sigma$ that is syntactically subsumed by all syntactic generalizations of $\Sigma$.*

**Proof**: Let $\Psi$ be the conjunction of all the MA timelines that syntactically generalize $\Sigma$ while having size no larger than $\Sigma$. As in the proof of Theorem 12, $\Psi$ is well defined. We show that $\Psi$ is a syntactic LGG for $\Sigma$. First, note that $\Psi$ syntactically generalizes $\Sigma$ because each timeline of $\Psi$ generalizes a timeline in every member of $\Sigma$, by the choice of $\Psi$. Now consider an arbitrary syntactic generalization $\Psi'$ of $\Sigma$. By the definition of syntactic subsumption, each timeline $\Phi$ in $\Psi'$ must subsume some timeline $\Phi_\alpha$ in each member $\alpha$ of $\Sigma$. Lemma 5 then implies that there is a timeline $\Phi'$ of size no larger than $\Sigma$ that subsumes all the $\Phi_\alpha$ while being subsumed by $\Phi$. By our choice of $\Psi$, the timeline $\Phi'$ must be a timeline of $\Psi$. It follows then that $\Psi'$ syntactically subsumes $\Psi$, and that $\Psi$ is a syntactic LGG of $\Sigma$ subsumed by all other syntactic generalizations of $\Sigma$. □

In general, we know that semantic and syntactic LGGs are different, though clearly the syntactic LGG is a semantic generalization and so must subsume the semantic LGG. For example, $(A; B) \wedge (B; A)$, and $A; B; A$ have a semantic LGG of $A; B; A$, as discussed above; but their syntactic LGG is $(A; B; \mathbf{true}) \wedge (\mathbf{true}; B; A)$, which subsumes $A; B; A$ but is not subsumed by $A; B; A$. Even so, for MA formulas:

**Proposition 22.** *For MA $\Phi$ and AMA $\Psi$, $\Phi \leq_{syn} \Psi$ is equivalent to $\Phi \leq \Psi$.*

**Proof**: The forward direction is immediate since we already know syntactic subsumption implies semantic subsumption. For the reverse direction, note that $\Phi \leq \Psi$ implies that each timeline of $\Psi$ subsumes $\Phi$—thus since $\Phi$ is a single timeline each timeline in $\Psi$ subsumes "some timeline" in $\Phi$ which is the definition of syntactic subsumption. □

**Proposition 23.** *Any syntactic AMA LGG for an MA formula set $\Sigma$ is also a semantic LGG for $\Sigma$.*

**Proof**: Now, consider a syntactic LGG $\Psi$ for $\Sigma$. Proposition 22 implies that $\Psi$ is a semantic generalization of $\Sigma$. Consider any semantic LGG $\Psi'$ of $\Sigma$. We show that $\Psi \leq \Psi'$ to conclude that $\Psi$ is a semantic LGG for $\Sigma$. Proposition 22 implies that $\Psi'$ syntactically subsumes $\Sigma$. It follows that $\Psi' \wedge \Psi$ syntactically subsumes $\Sigma$. But, $\Psi' \wedge \Psi$ is syntactically subsumed by $\Psi$, which is a syntactic LGG of $\Sigma$—it follows that $\Psi' \wedge \Psi$ syntactically subsumes $\Psi$, or $\Psi$ would not be a *least* syntactic generalization of $\Sigma$. But then $\Psi \leq (\Psi' \wedge \Psi)$, which implies $\Psi \leq \Psi'$, as desired. □

We note that the stronger result stating that a formula $\Psi$ is a syntactic LGG of a set $\Sigma$ of MA formulas if and only if it is a semantic LGG of $\Sigma$ is *not* an immediate consequence of our results above. At first examination, the strengthening appears trivial, given the equivalence of $\Phi \leq \Psi$ and $\Phi \leq_{\text{syn}} \Psi$ for MA $\Phi$. However, being semantically least is *not* necessarily a stronger condition than being syntactically least—we have not ruled out the possibility that a semantically least generalization $\Psi$ may syntactically subsume another generalization that is semantically (but not syntactically) equivalent. (This question is open, as we have not found an example of this phenomenon either.)

Proposition 23 together with Theorem 21 have the nice consequence for our learning approach that the syntactic LGG of two AMA formulas is a semantic LGG of those formulas, as long as the original formulas are themselves syntactic LGGs of sets of MA timelines. Because our learning approach starts with training examples that are converted to MA timelines using the LGCF operation, the syntactic LGGs computed (whether combining all the training examples at once, or incrementally computing syntactic LGGs of parts of the training data) are always syntactic LGGs of sets of MA timelines and hence are also semantic LGGs, in spite of the fact that syntactic subsumption is weaker than semantic subsumption. We note, however, that the resulting semantic LGGs may be considerably larger than the smallest semantic LGG (which may not be a syntactic LGG at all).

Using Proposition 23, we now show that we cannot hope for a polynomial-time syntactic LGG algorithm.

**Theorem 24.** *The smallest syntactic LGG of two MA formulas can be exponentially large.*

**Proof**: Suppose there is always a syntactic LGG of two MA formulas that is not exponentially large. Since by Proposition 23 each such formula is also a semantic LGG, there is always a semantic LGG of two MA formulas that is not exponentially large. This contradicts Theorem 17. □

While this is discouraging, we have an algorithm for the syntactic LGG whose time complexity matches this lower-bound, unlike the semantic LGG case, where the best algorithm we have is doubly exponential in the worst case. Theorem 14 yields an exponential time method for computing the semantic LGG of a set of MA timelines $\Sigma$—since for a timeline $\Phi$, $\text{IS}(\Phi) = \Phi$, we can simply conjoin all the timelines of $\text{IG}(\Sigma)$. Given a set of AMA formulas, the syntactic LGG algorithm uses this method to compute the polynomially-many semantic LGGs of sets of timelines, one chosen from each input formula, and conjoins all the results.

**Theorem 25.** *The formula* $\bigwedge_{\Phi_i \in \Psi_i} \text{IG}(\{\Phi_1, \ldots, \Phi_n\})$ *is a syntactic LGG of the AMA formulas* $\Psi_1, \ldots, \Psi_n$.

**Proof**: Let $\Psi$ be $\bigwedge_{\Phi_i \in \Psi_i} \text{IG}(\{\Phi_1, \ldots, \Phi_n\})$. Each timeline $\Phi$ of $\Psi$ must subsume each $\Psi_i$ because $\Phi$ is an output of IG on a set containing a timeline of $\Psi_i$—thus $\Psi$ syntactically subsumes each $\Psi_i$. To show that $\Psi$ is a syntactically least such formula, consider a $\Psi'$ that syntactically subsumes every $\Psi_i$. We show that $\Psi \leq_{\text{syn}} \Psi'$ to conclude. Each timeline $\Phi'$ in $\Psi'$ subsumes a timeline $T_i \in \Psi_i$, for each $i$, by our assumption that $\Psi_i \leq_{\text{syn}} \Psi'$. But then by Lemma 5, $\Phi'$ must subsume a member of $\text{IG}(\{T_1, \ldots, T_n\})$—and that member is a timeline of $\Psi$—so each timeline $\Phi'$ of $\Psi'$ subsumes a timeline of $\Psi$. We conclude $\Psi \leq_{\text{syn}} \Psi'$, as desired. □

This theorem yields an algorithm that computes a syntactic AMA LGG in exponential time—pseudo-code for this method is given in Figure 9. The exponential time bound follows from the fact that there are exponentially many ways to choose $\Phi_1, \ldots, \Phi_m$ in line 5, and for each of these there are exponentially many semantic-LGG members in line 6 (since the $\Phi_i$ are all MA timelines)—the product of these two exponentials is still an exponential.

1: syntactic-LGG$(\{\Psi_1, \Psi_2, \ldots, \Psi_m\})$
2:     *// Input: AMA formulas* $\{\Psi_1, \ldots, \Psi_m\}$
3:     *// Output: syntactic LGG of* $\{\Psi_1, \ldots, \Psi_m\}$
4:     $G := \{\}$;
5:     **for** each $\langle \Phi_1, \ldots, \Phi_m \rangle \in \Psi_1 \times \cdots \times \Psi_m$
6:         **for** each $\Phi$ in semantic-LGG$(\{\Phi_1, \ldots, \Phi_m\})$
7:             **if** $(\forall \Phi' \in G \,.\, \Phi' \not\leq \Phi)$
8:                 **then** $G' := \{\Phi'' \in G \mid \Phi \leq \Phi''\}$;
9:                     $G := (G - G') \cup \{\Phi\}$;
10:     **return** $(\bigwedge G)$

Figure 9: Pseudo-code that computes the syntactic AMA LGG of a set of AMA formulas.

The formula returned by the algorithm shown is actually a subset of the syntactic LGG given by Theorem 25. This subset is syntactically (and hence semantically) equivalent to the formula specified by the theorem, but is possibly smaller due to the pruning achieved by the **if** statement in lines 7–9. A timeline is pruned from the set if it is (semantically) subsumed by any other timeline in the set (one timeline is kept from any semantically equivalent group of timelines, at random). This pruning of timelines is sound, since a timeline is pruned from the output only if it subsumes some other formula in the output—this fact allows an easy argument that the pruned formula is syntactically equivalent to (i.e. mutually syntactically subsumed by) the unpruned formula. Section 4.4.2 traces the computations of this algorithm for an example LGG calculation. We note that in our empirical evaluation discussed in Section 6, there was no cost in terms of accuracy for using the more efficient syntactic vs. semantic LGG. We know this because our learned definitions made errors in the direction of being overly specific—thus, since the semantic-LGG is at least as specific as the syntactic-LGG there would be no advantage to using the semantic algorithm.

The method does an exponential amount of work even if the result is small (typically because many timelines can be pruned from the output because they subsume what remains). It is still an open question as to whether there is an output-efficient algorithm for computing the syntactic AMA LGG—this problem is in coNP and we conjecture that it is coNP-complete. One route to settling this question is to determine the output complexity of semantic LGG for MA input formulas. We believe that problem also to be coNP-complete, but have not proven this; if that problem is in P, there is an output-efficient method for computing syntactic AMA LGG based on Theorem 25.

A summary of the algorithmic complexity results from this section can be found in Table 3 in the conclusions section of this paper.

### 4.4 Examples: Least-General Generalization Calculations

Below we work through the details of a semantic and a syntactic LGG calculation. We consider the AMA formulas $\Psi = (A; B) \wedge (B; A)$ and $\Phi = A; B; A$, for which the semantic LGG is $A; B; A$ and the syntactic LGG is $(A; B; \mathbf{true}) \wedge (\mathbf{true}; B; A)$.

#### 4.4.1 SEMANTIC LGG EXAMPLE

The first step in calculating the semantic LGG, according to the algorithm given in Figure 8, is to compute the interdigitation-specializations of the input formulas (i.e., $\mathrm{IS}(\Phi)$ and $\mathrm{IS}(\Psi)$). Trivially, we have that $\mathrm{IS}(\Phi) = \Phi = A; B; A$. To calculate $\mathrm{IS}(\Psi)$, we must consider the possible interdigitations of $\Psi$, for which there are three,

$$\begin{array}{l} \{\ \langle A, B\rangle, \langle B, B\rangle, \langle B, A\rangle\ \} \\ \{\ \langle A, B\rangle, \langle B, A\rangle\ \} \\ \{\ \langle A, B\rangle, \langle A, A\rangle, \langle B, A\rangle\ \} \end{array}$$

Each interdigitation leads to the corresponding member of $\mathrm{IS}(\Psi)$ by unioning (conjoining) the states in each tuple, so $\mathrm{IS}(\Psi)$ is

$$\begin{array}{ll} \{ & (A \wedge B); B; (A \wedge B), \\ & (A \wedge B), \\ & (A \wedge B); A; (A \wedge B)\ \}. \end{array}$$

Lines 5–9 of the semantic LGG algorithm compute the set $S$, which is equal to the union of the timelines in $\mathrm{IS}(\Psi)$ and $\mathrm{IS}(\Phi)$, with all subsumed timelines removed. For our formulas, we see that each timeline in $\mathrm{IS}(\Psi)$ is subsumed by $\Phi$—thus, we have that $S = \Phi = A; B; A$.

After computing $S$, the algorithm returns the conjunction of timelines in $\mathrm{IG}(S)$, with redundant timelines removed (i.e., all subsuming timelines are removed). In our case, $\mathrm{IG}(S) = A; B; A$, trivially, as there is only one timeline in $S$, thus the algorithm correctly computes the semantic LGG of $\Psi$ and $\Phi$ to be $A; B; A$.

#### 4.4.2 SYNTACTIC LGG EXAMPLE

The syntactic LGG algorithm, shown in Figure 9, computes a series of semantic LGGs for MA timeline sets, returning the conjunction of the results (after pruning). Line 5 of the algorithm, cycles through timeline tuples from the cross-product of the input AMA formulas. In our case the tuples in $\Phi \times \Psi$ are $T_1 = \langle A; B; A,\ A; B\rangle$ and $T_2 = \langle A; B; A,\ B; A\rangle$—for each tuple, the algorithm computes the semantic LGG of the tuple's timelines.

The semantic LGG computation for each tuple uses the algorithm given in Figure 8, but the argument is always a set of MA timelines rather than AMA formulas. For this reason, lines 4–9 are superfluous, as for an MA timeline $\Phi'$, $\mathrm{IS}(\Phi') = \Phi'$. In the case of tuple $T_1$, lines 4–9 of the algorithm just compute $S = \{A; B; A,\ A; B\}$. It remains to compute the interdigitation-generalizations of $S$ (i.e., $\mathrm{IG}(S)$), returning the conjunction of those timelines after pruning (lines 10–15 in Figure 8). The set of all interdigitations of $S$ are,

$$\begin{array}{l} \{\ \langle A, A\rangle, \langle B, A\rangle, \langle B, B\rangle, \langle B, A\rangle\ \} \\ \{\ \langle A, A\rangle, \langle B, B\rangle, \langle B, A\rangle\ \} \\ \{\ \langle A, A\rangle, \langle A, B\rangle, \langle B, B\rangle, \langle B, A\rangle\ \} \\ \{\ \langle A, A\rangle, \langle A, B\rangle, \langle B, A\rangle\ \} \\ \{\ \langle A, A\rangle, \langle A, B\rangle, \langle A, A\rangle, \langle B, A\rangle\ \} \end{array}$$

By intersecting states in interdigitation tuples we get $\mathrm{IG}(S)$,

$$\{\ A; \mathbf{true}; B; \mathbf{true},\ A; B; \mathbf{true}, A; \mathbf{true}; B; \mathbf{true},\ A; \mathbf{true}; \mathbf{true},\ A; \mathbf{true}; A; \mathbf{true}\ \}$$

Since the timeline $A; B; \mathbf{true}$ is subsumed by all timelines in $\text{IG}(S)$, all other timelines will be pruned. Thus the semantic LGG algorithm returns $A; B; \mathbf{true}$ as the semantic LGG of the timelines in $T_1$.

Next the syntactic LGG algorithm computes the semantic LGG of the timelines in $T_2$. Following the same steps as for $T_1$, we find that the semantic LGG of the timelines in $T_2$ is $\mathbf{true}; B; A$. Since $A; B; \mathbf{true}$ and $\mathbf{true}; B; A$ do not subsume one another, the set $G$ computed by lines 5–9 of the syntactic LGG algorithm is equal to $\{ A; B; \mathbf{true}, \ \mathbf{true}; B; A \}$. Thus, the algorithm computes the syntactic LGG of $\Phi$ and $\Psi$ to be $(A; B; \mathbf{true}) \wedge (\mathbf{true}; B; A)$. Note that, in this case, the syntactic LGG is more general than the semantic LGG.

## 5. Practical Extensions

We have implemented a specific-to-general AMA learning algorithm based on the LGCF and syntactic LGG algorithms presented earlier. This implementation includes four practical extensions. The first extension aims at controlling the exponential complexity by limiting the length of the timelines we consider. Second we describe an often more efficient LGG algorithm based on a modified algorithm for computing pairwise LGGs. The third extension deals with applying our propositional algorithm to relational data, as is necessary for the application domain of visual event recognition. Fourth, we add negation into the AMA language and show how to compute the corresponding LGCFs and LGGs using our algorithms for AMA (without negation). Adding negation into AMA turns out to be crucial to achieving good performance in our experiments. We end this section with a review of the overall complexity of our implemented system.

### 5.1 k-AMA Least-General Generalization

We have already indicated that our syntactic AMA LGG algorithm takes exponential time relative to the lengths of the timelines in the AMA input formulas. This motivates restricting the AMA language to $k$-AMA in practice, where formulas contain timelines with no more than $k$ states. As $k$ is increased the algorithm is able to output increasingly specific formulas at the cost of an exponential increase in computational time. In the visual-event–recognition experiments shown later, as we increased $k$, the resulting formulas became overly specific before a computational bottleneck is reached—i.e., for that application the best values of $k$ were practically computable and the ability to limit $k$ provided a useful language bias.

We use a $k$-*cover* operator in order to limit our syntactic LGG algorithm to $k$-AMA. A $k$-cover of an AMA formula is a syntactically least general $k$-AMA formula that syntactically subsumes the input—it is easy to show that a $k$-cover for a formula can be formed by conjoining all $k$-MA timelines that syntactically subsume the formula (i.e., that subsume any timeline in the formula) . Figure 10 gives pseudo-code for computing the $k$-cover of an AMA formula. It can be shown that this algorithm correctly computes a $k$-cover for any input AMA formula. The algorithm calculates the set of least general $k$-MA timelines that subsume each timeline in the input—the resulting $k$-MA formulas are conjoined and "redundant" timelines are pruned using a subsumption test. We note that the $k$-cover of an AMA formula may itself be exponentially larger than that formula; however, in practice, we have found $k$-covers not to exhibit undue size growth.

Given the $k$-cover algorithm we restrict our learner to $k$-AMA as follows: 1) Compute the $k$-cover for each AMA input formula. 2) Compute the syntactic AMA LGG of the resulting $k$-AMA formulas. 3) Return the $k$-cover of the resulting AMA formula. The primary bottleneck of

k-cover$(k, \bigwedge_{1 \leq i \leq m} \Phi_i)$
    *// Input: positive natural number $k$, AMA formula $\bigwedge_{1 \leq i \leq m} \Phi_i$*
    *// Output: $k$-cover of $\bigwedge_{1 \leq i \leq m} \Phi_i$*
    $G := \{\}$;
    **for** $i := 1$ **to** $m$
        **for** each $P := \langle P_1, \ldots, P_n \rangle$ in all-values(a-k-partition$(k, \Phi_i)$)
            $\Phi := (\bigcap P_1); \ldots; (\bigcap P_n)$;
            **if** $(\forall \Phi' \in G \,.\, \Phi' \not\leq \Phi)$
                **then** $G' := \{\Phi'' \in G \mid \Phi \leq \Phi''\}$;
                    $G := (G - G') \cup \{\Phi\}$;
    **return** $(\bigwedge G)$

a-k-partition$(k, s_1; \ldots; s_j)$
    *// Input: positive natural number $k$, MA timeline $s_1; \ldots; s_j$*
    *// Output: a tuple of $\leq k$ sets of consecutive states that partitions $s_1, \ldots, s_j$*
    **if** $j \leq k$ **then return** $\langle \{s_1\}, \ldots, \{s_j\} \rangle$;
    **if** $k = 1$ **then return** $\langle \{s_1, \ldots, s_j\} \rangle$;
    $l :=$ a-member-of$(\{1, 2, \ldots, j - k + 1\})$;    *// pick next block size*
    $P_0 = \{s_1, \ldots, s_l\}$;    *// construct next block*
    **return** extend-tuple$(P_0,$ a-k-partition$(k - 1, s_{l+1}; \ldots; s_j))$;

Figure 10: Pseudo-code for non-deterministically computing a k-cover of an AMA formula, along with a non-deterministic helper function for selecting a $\leq k$ block partition of the states of a timeline.

the original syntactic LGG algorithm is computing the exponentially large set of interdigitation-generalizations—the $k$-limited algorithm limits this complexity as it only computes interdigitation-generalizations involving $k$-MA timelines.

### 5.2 Incremental Pairwise LGG Computation

Our implemented learner computes the syntactic k-AMA LGG of AMA formula sets—however, it does not directly use the algorithm describe above. Rather than compute the LGG of formula sets via a single call to the above algorithm, it is typically more efficient to break the computation into a sequence of pairwise LGG calculations. Below we describe this approach and the potential efficiency gains.

It is straightforward to show that for both syntactic and semantic subsumption we have that LGG$(\Psi_1, \ldots, \Psi_m)$ = LGG$(\Psi_1,$ LGG$(\Psi_2, \ldots, \Psi_m))$ where the $\Psi_i$ are AMA formulas. Thus, by recursively applying this transformation we can incrementally compute the LGG of $m$ AMA formulas via a sequence of $m - 1$ pairwise LGG calculations. Note that since the LGG operator is commutative and associative the final result does not depend on the order in which we process the formulas. We will refer to this incremental pairwise LGG strategy as the *incremental approach* and to the strategy that makes a single call to the k-AMA LGG algorithm (passing in the entire formula set) as the *direct approach*.

To simplify the discussion we will consider computing the LGG of an MA formula set $\Sigma$—the argument can be extended easily to AMA formulas (and hence to k-AMA). Recall that the syntactic LGG algorithm of Figure 9 computes $\text{LGG}(\Sigma)$ by conjoining timelines in $\text{IG}(\Sigma)$ that do not subsume any of the others, eliminating subsuming timelines in a form of pruning. The incremental approach applies this pruning step after each pair of input formulas is processed—in contrast, the direct approach must compute the interdigitation-generalization of all the input formulas before any pruning can happen. The resulting savings can be substantial, and typically more than compensates for the extra effort spent checking for pruning (i.e. testing subsumption between timelines as the incremental LGG is computed). A formal approach to describing these savings can be constructed based on the observation that both $\bigcup_{\Phi \in \text{IG}(\{\Phi_1, \Phi_2\})} \text{IG}(\{\Phi\} \cup \Sigma)$ and $\bigcup_{\Phi \in \text{LGG}(\Phi_1, \Phi_2)} \text{IG}(\{\Phi\} \cup \Sigma)$ can be seen to compute the LGG of $\Sigma \cup \{\Phi_1, \Phi_2\}$, but with the latter being possibly much cheaper to compute due to pruning. That is, $\text{LGG}(\Phi_1, \Phi_2)$ typically contains a much smaller number of timelines than $\text{IG}(\{\Phi_1, \Phi_2\})$.

Based on the above observations our implemented system uses the incremental approach to compute the LGG of a formula set. We now describe an optimization used in our system to speedup the computation of pairwise LGGs, compared to directly running the algorithm in Figure 9. Given a pair of AMA formulas $\Psi_1 = \Phi_{1,1} \wedge \cdots \wedge \Phi_{1,m}$ and $\Psi_2 = \Phi_{2,1} \wedge \cdots \wedge \Phi_{2,n}$, let $\Psi$ be their syntactic LGG obtained by running the algorithm in Figure 9. The algorithm constructs $\Psi$ by computing LGGs of all MA timeline pairs (i.e., $\text{LGG}(\Phi_{1,i}, \Phi_{2,j})$ for all $i$ and $j$) and conjoining the results while removing subsuming timelines. It turns out that we can often avoid computing many of these MA LGGs. To see this consider the case when there exists $i$ and $j$ such that $\Phi_{1,i} \leq \Phi_{2,j}$, we know $\text{LGG}(\Phi_{1,i}, \Phi_{2,j}) = \Phi_{2,j}$ which tells us that that $\Phi_{2,j}$ will be considered for inclusion into $\Psi$ (it may be pruned). Furthermore we know that any other LGG involving $\Phi_{2,j}$ will subsume $\Phi_{2,j}$ and thus will be pruned from $\Psi$. This shows that we need not compute any MA LGGs involving $\Phi_{2,j}$, rather we need only to consider adding $\Phi_{2,j}$ when constructing $\Psi$.

The above observation leads to a modified algorithm (used in our system) for computing the syntactic LGG of *a pair* of AMA formulas. The new algorithm only computes LGGs between non-subsuming timelines. Given AMA formulas $\Psi_1$ and $\Psi_2$, the modified algorithm proceeds as follows: 1) Compute the *subsumer set* $S = \{\Phi \in \Psi_1 \mid \exists \Phi' \in \Psi_2 \; s.t. \; \Phi' \leq \Phi\} \cup \{\Phi \in \Psi_2 \mid \exists \Phi' \in \Psi_1 \; s.t. \; \Phi' \leq \Phi\}$. 2) Let AMA $\Psi_1'$ ($\Psi_2'$) be the result of removing timelines from $\Psi_1$ ($\Psi_2$) that are in $S$. 3) Let $\Psi'$ be the syntactic LGG of $\Psi_1'$ and $\Psi_2'$ computed by running the algorithm in Figure 9 (if either $\Psi_i'$ is empty then $\Psi'$ will be empty). 4) Let $S'$ be the conjunction of timelines in $S$ that do not subsume any timeline in $\Psi'$. 5) Return $\Psi = \Psi' \wedge S'$. This method avoids computing MA LGGs involving subsuming timelines (an exponential operation) at the cost of performing polynomially many MA subsumption tests (a polynomial operation). We have noticed a significant advantage to using this procedure in our experiments. In particular, the advantage tends to grow as we process more training examples. This is due to the fact that as we incrementally process training examples the resulting formulas become more general—thus, these more general formulas are likely to have more subsuming timelines. In the best case when $\Psi_1 \leq_{\text{syn}} \Psi_2$ (i.e., all timelines in $\Psi_2$ are subsuming), we see that step 2 produces an empty formula and thus step 3 (the expensive step) performs no work—in this case we return the set $S = \Psi_2$ as desired.

### 5.3 Relational Data

LEONARD produces relational models that involve objects and (force dynamic) relations between those objects. Thus event definitions include variables to allow generalization over objects. For example, a definition for PICKUP$(x, y, z)$ recognizes both PICKUP(**hand**, **block**, **table**) as well as PICKUP(**man**, **box**, **floor**). Despite the fact that our $k$-AMA learning algorithm is propositional, we are still able to use it to learn relational definitions.

We take a straightforward object-correspondence approach to relational learning. We view the models output by LEONARD as containing relations applied to constants. Since we (currently) support only supervised learning, we have a set of distinct training examples for each event type. There is an implicit correspondence between the objects filling the same role across the different training models for a given type. For example, models showing PICKUP(**hand**, **block**, **table**) and PICKUP(**man**, **box**, **floor**) have implicit correspondences given by ⟨**hand**, **man**⟩, ⟨**block**, **box**⟩, and ⟨**table**, **floor**⟩. We outline two relational learning methods that differ in how much object-correspondence information they require as part of the training data.

#### 5.3.1 COMPLETE OBJECT CORRESPONDENCE

This first approach assumes that a complete object correspondence is given, as input, along with the training examples. Given such information, we can propositionalize the training models by replacing corresponding objects with unique constants. The propositionalized models are then given to our propositional $k$-AMA learning algorithm which returns a propositional $k$-AMA formula. We then lift this propositional formula by replacing each constant with a distinct variable. Lavrac et al. (1991) has taken a similar approach.

#### 5.3.2 PARTIAL OBJECT CORRESPONDENCE

The above approach assumes complete object-correspondence information. While it is sometimes possible to provide all correspondences (for example, by color-coding objects that fill identical roles when recording training movies), such information is not always available. When only a partial object correspondence (or even none at all) is available, we can automatically complete the correspondence and apply the above technique.

For the moment, assume that we have an evaluation function that takes two relational models and a candidate object correspondence, as input, and yields an evaluation of correspondence quality. Given a set of training examples with missing object correspondences, we perform a greedy search for the best set of object-correspondence completions over the models. Our method works by storing a set $P$ of propositionalized training examples (initially empty) and a set $U$ of unpropositionalized training examples (initially the entire training set). For the first step, when $P$ is empty, we evaluate all pairs of examples from $U$, under all possible correspondences, select the pair that yields the highest score, remove the examples involved in that pair from $U$, propositionalize them according to the best correspondence, and add them to $P$. For each subsequent step, we use the previously computed values of all pairs of examples, one from $U$ and one from $P$, under all possible correspondences. We then select the example from $U$ and correspondence that yields the highest average score relative to all models in $P$—this example is removed from $U$, propositionalized according to the winning correspondence, and added to $P$. For a fixed number of objects, the effort expended here is polynomial in the size of the training set; however, if the number of objects $b$ that appear in a training example is allowed to grow, the number of correspondences that must be considered grows as $b^b$. For this reason, it is important that the events involved manipulate only a modest number of objects.

Our evaluation function is based on the intuition that object roles for visual events (as well as events from other domains) can often be inferred by considering the changes between the initial and final moments of an event. Specifically, given two models and an object correspondence, we first propositionalize the models according to the correspondence. Next, we compute ADD and DELETE lists for each model. The ADD list is the set of propositions that are true at the final moment but not the initial moment. The DELETE list is the set of propositions that are true at the initial moment but not the final moment. These add and delete lists are motivated by STRIPS action representations (Fikes & Nilsson, 1971). Given such $\text{ADD}_i$ and $\text{DELETE}_i$ lists for models 1 and 2, the evaluation function returns the sum of the cardinalities of $\text{ADD}_1 \cap \text{ADD}_2$ and $\text{DELETE}_1 \cap \text{DELETE}_2$. This heuristic measures the similarity between the ADD and DELETE lists of the two models. The intuition behind this heuristic is similar to the intuition behind the STRIPS action-description language—i.e., that most of the differences between the initial and final moments of an event occurrence are related to the target event, and that event effects can be described by ADD and DELETE lists. We have found that this evaluation function works well in the visual-event domain.

Note, that when full object correspondences are given to the learner (rather than automatically extracted by the learner), the training examples are interpreted as specifying that the target event took place as well as which objects filled the various event roles (e.g., PICKUP(a,b,c)). Rather, when no object correspondences are provided the training examples are interpreted as specifying the existence of a target event occurrence but do not specify which objects fill the roles (i.e., the training example is labeled by PICKUP rather than PICKUP(a,b,c)). Accordingly, the rules learned when no correspondences are provided only allow us to infer that a target event occurred and not which objects filled the event roles. For example when object correspondences are manually provided the learner might produce the rule,

$$\textsc{PickUp}(x, y, z) \triangleq \begin{bmatrix} (\textsc{Supports}(z, y) \wedge \textsc{Contacts}(z, y)); \\ (\textsc{Supports}(x, y) \wedge \textsc{Attached}(x, y)) \end{bmatrix}$$

whereas a learner that automatically extracts the correspondences would instead produce the rule,

$$\textsc{PickUp} \triangleq \begin{bmatrix} (\textsc{Supports}(z, y) \wedge \textsc{Contacts}(z, y)); \\ (\textsc{Supports}(x, y) \wedge \textsc{Attached}(x, y)) \end{bmatrix}$$

Its worth noting, however, that upon producing the second rule the availability of a single training example with correspondence information allows the learner to determine the roles of the variables, upon which it can output the first rule. Thus, under the assumption that the learner can reliably extract object correspondences, we need not label all training examples with correspondence information in order to obtain definitions that explicitly recognize object roles.

## 5.4 Negative Information

The AMA language does not allow negated propositions. Negation, however, is sometimes necessary to adequately define an event type. In this section, we consider the language $\text{AMA}^-$, which is a superset of AMA, with the addition of negated propositions. We first give the syntax and semantics of $\text{AMA}^-$, and extend AMA syntactic subsumption to $\text{AMA}^-$. Next, we describe our approach to learning AMA$^-$ formulas using the above-presented algorithms for AMA. We show that our approach correctly computes the AMA$^-$ LGCF and the syntactic AMA$^-$ LGG. Finally, we discuss an alternative, related approach to adding negation designed to reduce the overfitting that appears to result from the full consideration of negated propositions.

AMA$^-$ has the same syntax as AMA, only with a new grammar for building states with negated propositions:

$$\begin{aligned} \textit{literal} &::= \texttt{true} \mid \textit{prop} \mid \neg\Diamond\textit{prop} \\ \textit{state} &::= \textit{literal} \mid \textit{literal} \wedge \textit{state} \end{aligned}$$

where *prop* is any primitive proposition. The semantics of AMA$^-$ are the same as for AMA except for state satisfaction.

- A *positive literal* $P$ (*negative literal* $\neg\Diamond P$) is satisfied by model $\langle M, I\rangle$ iff $M[x]$ assigns $P$ true (false), for every $x \in I$.[10]
- A state $l_1 \wedge \cdots \wedge l_m$ is satisfied by model $\langle M, I\rangle$ iff each literal $l_i$ is satisfied by $\langle M, I\rangle$.

**Subsumption.** An important difference between AMA and AMA$^-$ is that Proposition 2, establishing the existence of witnessing interdigitations to MA subsumption, is no longer true for MA$^-$. In other words, if we have two timelines $\Phi_1, \Phi_2 \in$ AMA$^-$, such that $\Phi_1 \leq \Phi_2$, there need not be an interdigitation that witnesses $\Phi_1 \leq \Phi_2$. To see this, consider the AMA$^-$ timelines:

$$\begin{aligned} \Phi_1 &= (a \wedge b \wedge c); b; a; b; (a \wedge b \wedge \neg\diamond c) \\ \Phi_2 &= b; a; c; a; b; a; \neg\diamond c; a; b \end{aligned}$$

We can then argue:

1. *There is no interdigitation that witnesses* $\Phi_1 \leq \Phi_2$. To see this, first show that, in any such witness, the second and fourth states of $\Phi_1$ (each just "$b$") must interdigitate to align with either the first and fifth, or the fifth and ninth states of $\Phi_2$ (also, each just "$b$"). But in either of these cases, the third state of $\Phi_1$ will interdigitate with states of $\Phi_2$ that do not subsume it.

2. *Even so, we still have that* $\Phi_1 \leq \Phi_2$. To see this, consider any model $\langle M, I\rangle$ that satisfies $\Phi_1$. There must be an interval $[i_1, i_2]$ within $I$ such that $\langle M, [i_1, i_2]\rangle$ satisfies the third state of $\Phi_1$, that is the state "$a$." We have two cases:

   (a) The proposition $c$ is true at some point in $\langle M, [i_1, i_2]\rangle$. Then, one can verify that $\langle M, I\rangle$ satisfies both $\Phi_1$ and $\Phi_2$ in the following alignment:

$$\begin{array}{lllllll} \Phi_1 & = & (a \wedge b \wedge c); b; & a; & b; & (a \wedge b \wedge \neg\diamond c) \\ \Phi_2 & = & b; & a; c; a; & b; & a; \neg\diamond c; a; b \end{array}$$

10. We note that it is important that we use the notation $\neg\Diamond P$ rather than just $\neg P$. In event-logic, the formula $\neg P$ is satisfied by a model whenever $P$ is false as some instant in the model. Rather, event-logic interprets $\neg\Diamond P$ as indicating that $P$ is never true in the model (as defined above). Notice that the first form of negation does not yield a liquid property—i.e., $\neg P$ can be true along an interval but not necessarily during all subintervals. The second form of negation, however, does yield a liquid property provided that $P$ is liquid. This is important to our learning algorithms, since they all assume states are built from liquid properties.

(b) The proposition $c$ is false everywhere in $\langle M, [i_1, i_2]\rangle$. Then, one can verify that $\langle M, I\rangle$ satisfies both $\Phi_1$ and $\Phi_2$ in the following alignment:

$$\begin{array}{lclllll} \Phi_1 & = & (a \wedge b \wedge c); & b; & a; & b; (a \wedge b \wedge \neg \diamond c) \\ \Phi_2 & = & b; a; c; a; & b; & a; \neg \diamond c; a; & b \end{array}$$

It follows that $\Phi_1 \leq \Phi_2$.

In light of such examples, we conjecture that it is computationally hard to compute AMA$^-$ subsumption even between timelines. For this reason, we extend our definition of syntactic subsumption to AMA$^-$ in a way that provides a clearly tractable subsumption test analogous to that discussed above for AMA.

**Definition 6.** AMA$^-$ $\Psi_1$ *is syntactically subsumed by* AMA$^-$ $\Psi_2$ *(written* $\Psi_1 \leq_{syn} \Psi_2$*) iff for each timeline* $\Phi_2 \in \Psi_2$*, there is a timeline* $\Phi_1 \in \Psi_1$ *such that there is a witnessing interdigitation for* $\Phi_1 \leq \Phi_2$.

The difference between the definition here and the previous one for AMA is that here we only need to test for witnessing interdigitations between timelines rather than subsumption between timelines. For AMA formulas, we note that the new and old definition are equivalent (due to Proposition 2); however, for AMA$^-$ the new definition is weaker, and will result in more general LGG formulas. As one might expect, AMA$^-$ syntactic subsumption implies semantic subsumption and can be tested in polynomial-time using the subsumption graph described in Lemma 8 to test for witnesses.

**Learning.** Rather than design new LGCF and LGG algorithms to directly handle AMA$^-$, we instead compute these functions indirectly by applying our algorithms for AMA to a transformed problem. Intuitively, we do this by adding new propositions to our models (i.e., the training examples) that represent the proposition negations. Assume that the training-example models are over the set of propositions $P = \{p_1, \ldots, p_n\}$. We introduce a new set $\overline{P} = \{\bar{p}_1, \ldots, \bar{p}_n\}$ of propositions and use these to construct new training models over $P \cup \overline{P}$ by assigning true to $\bar{p}_i$ at a time in a model iff $p_i$ is false in the model at that time. After forming the new set of training models (each with twice as many propositions as the original models) we compute the least general AMA formula that covers the new models (by computing the AMA LGCFs and applying the syntactic AMA LGG algorithm), resulting in an AMA formula $\Psi$ over the propositions $P \cup \overline{P}$. Finally we replace each $\bar{p}_i$ in $\Psi$ with $\neg\Diamond p_i$ resulting in an AMA$^-$ formula $\Psi'$ over propositions in $P$—it turns out that under syntactic subsumption $\Psi'$ is the the least general AMA$^-$ formula that covers the original training models.

We now show the correctness of the above transformational approach to computing the AMA$^-$ LGCF and syntactic LGG. First, we introduce some notation. Let $\mathcal{M}$ be the set of all models over $P$. Let $\overline{\mathcal{M}}$ be the set of models over $P \cup \overline{P}$, such that at any time, for each $i$, exactly one of $p_i$ and $\bar{p}_i$ is true. Let $T$ be the following mapping from $\mathcal{M}$ to $\overline{\mathcal{M}}$: for $\langle M, I\rangle \in \mathcal{M}$, $T[\langle M, I\rangle]$ is the unique $\langle M', I\rangle \in \overline{\mathcal{M}}$ such that for all $j \in I$ and all $i$, $M'(j)$ assigns $p_i$ true iff $M(j)$ assigns $p_i$ true. Notice that the inverse of $T$ is a functional mapping from $\overline{\mathcal{M}}$ to $\mathcal{M}$. Our approach to handling negation using purely AMA algorithms begins by applying $T$ to the original training models. In what follows, we consider AMA$^-$ formulas over the propositions in $P$, and AMA formulas over the propositions in $P \cup \overline{P}$.

Let $F$ be a mapping from AMA$^-$ to AMA where for $\Psi \in$ AMA$^-$, $F[\Psi]$ is an AMA formula identical to $\Psi$ except that each $\neg\Diamond p_i$ in $\Psi$ is replaced with $\bar{p}_i$. Notice that the inverse of $F$ is a func-tion from AMA to AMA$^-$ and corresponds to the final step in our approach described above. The following lemma shows that there is a one-to-one correspondence between satisfaction of AMA$^-$ formulas by models in $\mathcal{M}$ and satisfaction of AMA formulas by models in $\overline{\mathcal{M}}$.

**Lemma 26.** *For any model $\langle M, I\rangle \in \mathcal{M}$ and any $\Psi \in$ AMA$^-$, $\Psi$ covers $\langle M, I\rangle$ iff $F[\Psi]$ covers $T[\langle M, I\rangle]$.*

Using this lemma, it is straightforward to show that our transformational approach computes the AMA$^-$ LGCF under semantic subsumption (and hence under syntactic subsumption).

**Proposition 27.** *For any $\langle M, I\rangle \in \mathcal{M}$, let $\Phi$ be the AMA LGCF of the model $T[\langle M, I\rangle]$. Then, $F^{-1}[\Phi]$ is the unique* AMA$^-$ *LGCF of $\langle M, I\rangle$, up to equivalence.*

**Proof**: We know that $\Phi$ covers $T[\langle M, I\rangle]$, therefore by Lemma 26 we know that $F^{-1}[\Phi]$ covers $\langle M, I\rangle$. We now show that $F^{-1}[\Phi]$ is the least-general formula in AMA$^-$ that covers $\langle M, I\rangle$. For the sake of contradiction assume that some $\Phi' \in$ AMA$^-$ covers $\langle M, I\rangle$ but that $\Phi' < F^{-1}[\Phi]$. It follows that there is some model $\langle M', I'\rangle$ that is covered by $F^{-1}[\Phi]$ but not by $\Phi'$. By Lemma 26 we have that $F[\Phi']$ covers $T[\langle M, I\rangle]$ and since $\Phi$ is the unique AMA LGCF of $T[\langle M, I\rangle]$, up to equivalence, we have that $\Phi \leq F[\Phi']$. However, we also have that $T[\langle M', I'\rangle]$ is covered by $\Phi$ but not by $F[\Phi']$ which gives a contradiction. Thus, no such $\Phi'$ can exist. It follows that $\Phi$ is an AMA$^-$ LGCF. The uniqueness of the AMA$^-$ LGCF up to equivalence follows because AMA$^-$ is closed under conjunction; so that if there were any two non-equivalent LGCF formulas, they could be conjoined to get an LGCF formula strictly less than one of them. □

Below we use the fact that the $F$ operator preserves syntactic subsumption. In particular, given two MA$^-$ timelines $\Phi_1, \Phi_2$, it is clear that any witnessing interdigitation of $\Phi_1 \leq \Phi_2$ can be trivially converted into a witness for $F[\Phi_1] \leq F[\Phi_2]$ (and vice versa). Since syntactic subsumption is defined in terms of witnessing interdigitations, it follows that for any $\Psi_1, \Psi_2 \in$ AMA$^-$, $(\Psi_1 \leq_{\text{syn}} \Psi_2)$ iff $(F[\Psi_1] \leq_{\text{syn}} F[\Psi_2])$. Using this property, it is straightforward to show how to compute the syntactic AMA$^-$ LGG using the syntactic AMA LGG algorithm.

**Proposition 28.** *For any* AMA$^-$ *formulas $\Psi_1, \ldots, \Psi_m$, let $\Psi$ be the syntactic AMA LGG of $\{F[\Psi_1], \ldots, F[\Psi_m]\}$. Then, $F^{-1}[\Psi]$ is the unique syntactic* AMA$^-$ *LGG of $\{\Psi_1, \ldots, \Psi_m\}$.*

**Proof**: We know that for each $i$, $F[\Psi_i] \leq_{\text{syn}} \Psi$—thus, since $F^{-1}$ preserves syntactic subsumption, we have that for each $i$, $\Psi_i \leq_{\text{syn}} F^{-1}[\Psi]$. This shows that $F^{-1}[\Psi]$ is a generalization of the inputs. We now show that $F^{-1}[\Psi]$ is the least such formula. For the sake of contradiction assume that $F^{-1}[\Psi]$ is not least. It follows that there must be a $\Psi' \in$ AMA$^-$ such that $\Psi' <_{\text{syn}} F^{-1}[\Psi]$ and for each $i$, $\Psi_i \leq_{\text{syn}} \Psi'$. Combining this with the fact that $F$ preserves syntactic subsumption, we get that $F[\Psi'] <_{\text{syn}} \Psi$ and for each $i$, $F[\Psi_i] \leq F[\Psi']$. But this contradicts the fact that $\Psi$ is an LGG; so we must have that $F^{-1}[\Psi]$ is a syntactic AMA$^-$ LGG. As argued elsewhere, the uniqueness of this LGG follows from the fact that AMA$^-$ is closed under conjunction. □

These propositions ensure the correctness of our transformational approach to computing the syntactic LGG within AMA$^-$. For the case of semantic subsumption, the transformational approach does not correctly compute the AMA$^-$ LGG. To see this, recall that above we have given two timelines $\Phi_1, \Phi_2 \in$ AMA$^-$, such that $\Phi_1 \leq \Phi_2$, but there is no witnessing interdigitation. Clearly under semantic subsumption, the AMA$^-$ LGG of $\Phi_1$ and $\Phi_2$ is $\Phi_2$. However, the semantic AMA LGG of $F[\Phi_1]$ and $F[\Phi_2]$ is not $F[\Phi_2]$. The reason for this is that since there is no witness to $F[\Phi_1] \leq F[\Phi_2]$ (and the $F[\Phi_i]$ are MA timelines), we know by Proposition 2 that $F[\Phi_1] \not\leq F[\Phi_2]$. Thus, $F[\Phi_2]$ cannot be returned as the AMA LGG, since it does not subsume both input formulas—this shows that the transformational approach will not return $\Phi_2 = F^{-1}[F[\Phi_2]]$. Here, the transformational approach will produce an AMA$^-$ formula that is more general than $\Phi_2$.

On the computational side, we note that, since the transformational approach doubles the number of propositions in the training data, algorithms specifically designed for AMA$^-$ may be more efficient. Such algorithms might leverage the special structure of the transformed examples that our AMA algorithms ignore—in particular, that exactly one of $p_i$ or $\bar{p}_i$ is true at any time.

**Boundary Negation.** In our experiments, we actually compare two methods for assigning truth values to the $\bar{p}_i$ propositions in the training data models. The first method, called *full negation*, assigns truth values as described above, yielding the syntactically least-general AMA$^-$ formula that covers the examples. We found, however, that using full negation often results in learning overly specific formulas. To help alleviate this problem, our second method places a bias on the use of negation. Our choice of bias is inspired by the idea that, often, much of the useful information for characterizing an event type is in its pre- and post-conditions. The second method, called *boundary negation*, differs from full negation in that it only allows $\bar{p}_i$ to be true in the initial and final moments of a model (and then only if $p_i$ is false). $\bar{p}_i$ must be false at all other times. That is, we only allow "informative" negative information at the beginnings and ends of the training examples. We have found that boundary negation provides a good trade-off between no negation (i.e., AMA), which often produces overly general results, and full negation (i.e., AMA$^-$), which often produces overly specific and much more complicated results.

### 5.5 Overall Complexity and Scalability

We now review the overall complexity of our visual event learning component and discuss some scalability issues. Given a training set of temporal models (i.e., a set of movies), our system does the following: 1) Propositionalize the training models, translating negation as descried in Section 5.4. 2) Compute the LGCF of each propositional model. 3) Compute the $k$-AMA LGG of the LGCFs. 4) Return a lifted (variablized) version of the LGG. Steps two and four require little computational overhead, being linear in the sizes of the input and output respectively. Steps one and three are the computational bottlenecks of the system—they encompass the inherent exponential complexity arising from the relational and temporal problem structure.

**Step One.** Recall from Section 5.3.2 that our system allows the user to annotate training examples with object correspondence information. Our technique for propositionalizing the models was shown to be exponential in the number of unannotated objects in a training example. Thus, our system requires that the number of objects be relatively small or that correspondence information be given for all but a small number of objects. Often the event class definitions we are interested in do not involve a large number of objects. When this is true, in a controlled learning setting we can manage the relational complexity by generating training examples with only a small number (or zero) irrelevant objects. This is the case for all of the domains studied empirically in this paper.

In a less controlled setting, the number of unannotated objects may prohibit the use of our correspondence technique—there are at least three ways one might proceed. First, we can try to develop efficient domain-specific techniques for filtering objects and finding correspondences. That is, for a particular problem it may be possible to construct a simple filter that removes irrelevant objects from consideration and then to find correspondences for any remaining objects. Second, we can provide the learning algorithm with a set of hand-coded first-order formulas, defining a set of domain-specific features (e.g., in the spirit of Roth & Yih, 2001). These features can then be used to propositionalize the training instances. Third, we can draw upon ideas from relational learning to design a "truly first-order" version of the $k$-AMA learning algorithm. For example, one could use existing first-order generalization algorithms to generalize relational state descriptions. Effectively this approach pushes the object correspondence problem into the $k$-AMA learning algorithm rather than treating it as a preprocessing step. Since it is well known that computing first-order LGGs can be intractable (Plotkin, 1971), practical generalization algorithms retain tractability by constraining the LGGs in various ways (e.g., Muggleton & Feng, 1992; Morales, 1997).

**Step Three.** Our system uses the ideas of Section 5.2 to speedup the $k$-AMA LGG computation for a set of training data. Nevertheless, the computational complexity is still exponential in $k$—thus, in practice we are restricted to using relatively small values of $k$. While this restriction did not limit performance in our visual event experiments, we expect that it will limit the direct applicability of our system to more complex problems. In particular, many event types of interest may not be adequately represented via $k$-AMA when $k$ is small. Such event types, however, often contain significant hierarchical structure—i.e., they can be decomposed into a set of "short" sub-event types. An interesting research direction is to consider using our $k$-AMA learner as a component of a hierarchical learning system—there it could be used to learn $k$-AMA sub-event types. We note that our learner alone cannot be applied hierarchically because it requires liquid primitive events, but learns non-liquid composite event types. Further work is required (and intended) to construct a hierarchical learner based perhaps on non-liquid AMA learning.

Finally, recall that to compute the LGG of $m$ examples, our system uses a sequence of $m - 1$ pairwise LGG calculations. For a fixed $k$, each pairwise calculation takes polynomial time. However, since the size of a pairwise LGG can grow by at least a constant factor with respect to the inputs, the worst-case time complexity of computing the sequence of $m-1$ pairwise LGGs is exponential in $m$. We expect that this worst case will primarily occur when the target event type does not have a compact $k$-AMA representation—in which case a hierarchical approach as described above is more appropriate. When there is a compact representation, our empirical experience indicates that such growth does not occur—in particular, each pairwise LGG tends to yield significant pruning. For such problems, reasonable assumptions about the amount of pruning[11] imply that the time complexity of computing the sequence of $m - 1$ pairwise LGGs is polynomial in $m$.

## 6. Experiments

### 6.1 Data Set

Our data set contains examples of 7 different event types: *pick up*, *put down*, *stack*, *unstack*, *move*, *assemble*, and *disassemble*. Each of these involve a hand and two to three blocks. For a detailed description and sample video sequences of these event types, see Siskind (2001). Key frames from sample video sequences of these event types are shown in Figure 11. The results of segmentation,

11. In particular, assume that the size of a pairwise $k$-AMA LGG is "usually" bounded by the sizes of the $k$-covers of the inputs.

tracking, and model reconstruction are overlaid on the video frames. We recorded $30$ movies for each of the $7$ event classes resulting in a total of $210$ movies comprising $11946$ frames.[12] We replaced one *assemble* movie (assemble-left-qobi-04), with a duplicate copy of another (assemble-left-qobi-11) because of segmentation and tracking errors.

Some of the event classes are hierarchical in that occurrences of events in one class contain occurrences of events in one or more simpler classes. For example, a movie depicting a $\textsc{Move}(a, b, c, d)$ event (i.e. $a$ moves $b$ from $c$ to $d$) contains subintervals where $\textsc{PickUp}(a, b, c)$ and $\textsc{PutDown}(a, b, d)$ events occur. In our experiments, when learning the definition of an event class only the movies for that event class are used in training. We do not train on movies for other event classes that may also depict an occurrence of the event class being learned as a subevent. However, in evaluating the learned definitions, we wish to detect both the events that correspond to an entire movie as well as subevents that correspond to portions of that movie. For example, given a movie depicting a $\textsc{Move}(a, b, c, d)$ event, we wish to detect not only the $\textsc{Move}(a, b, c, d)$ event but also the $\textsc{PickUp}(a, b, c)$ and $\textsc{PutDown}(a, b, d)$ subevents as well. For each movie type in our data set, we have a set of *intended* events and subevents that should be detected. If a definition does not detect an intended event, we deem the error a false negative. If a definition detects an unintended event, we deem the error a false positive. For example, if a movie depicts a $\textsc{Move}(a, b, c, d)$ event, the intended events are $\textsc{Move}(a, b, c, d)$, $\textsc{PickUp}(a, b, c)$, and $\textsc{PutDown}(a, b, c)$. If the definition for *pick up* detects the occurrence of $\textsc{PickUp}(c, b, a)$ and $\textsc{PickUp}(b, a, c)$, but not $\textsc{PickUp}(a, b, c)$, it will be charged two false positives as well as one false negative. We evaluate our definitions in terms of false positive and negative rates as describe below.

### 6.2 Experimental Procedure

For each event type, we evaluate the $k$-AMA learning algorithm using a leave-one-movie-out cross-validation technique with training-set sampling. The parameters to our learning algorithm are $k$ and the degree $D$ of negative information used. The value of $D$ is either P, for *positive propositions only*, BN, for *boundary negation*, or N, for *full negation*. The parameters to our evaluation procedure include the target event type $E$ and the training-set size $N$. Given this information, the evaluation proceeds as follows: For each movie $M$ (the held-out movie) from the $210$ movies, apply the $k$-AMA learning algorithm to a randomly drawn training sample of $N$ movies from the $30$ movies of event type $E$ (or $29$ movies if $M$ is one of the $30$). Use LEONARD to detect all occurrences of the learned event definition in $M$. Based on $E$ and the event type of $M$, record the number of false positives and false negatives in $M$, as detected by LEONARD. Let FP and FN be the total number of false positives and false negatives observed over all $210$ held-out movies respectively. Repeat the entire process of calculating FP and FN $10$ times and record the averages as $\overline{\text{FP}}$ and $\overline{\text{FN}}$.[13]

Since some event types occur more frequently in our data than others because simpler events occur as subevents of more complex events but not vice versa, we do not report $\overline{\text{FP}}$ and $\overline{\text{FN}}$ directly. Instead, we normalize $\overline{\text{FP}}$ by dividing by the total number of times LEONARD detected the target event correctly or incorrectly within all $210$ movies and we normalize $\overline{\text{FN}}$ by dividing by the total

12. The source code and all of the data used for these experiments are available as Online Appendix 1, and also from `ftp://ftp.ecn.purdue.edu/qobi/ama.tar.Z`.

13. While we did not record the times for our experiments, the system is fast enough to give live demos when $N = 29$ and $k = 3$ with boundary negation, giving the best results we show here (though we don't typically record 29 training videos in a live demo for other reasons). Some of the less favorable parameter settings (particularly $k = 4$ and full negation) can take a (real-time) hour or so.

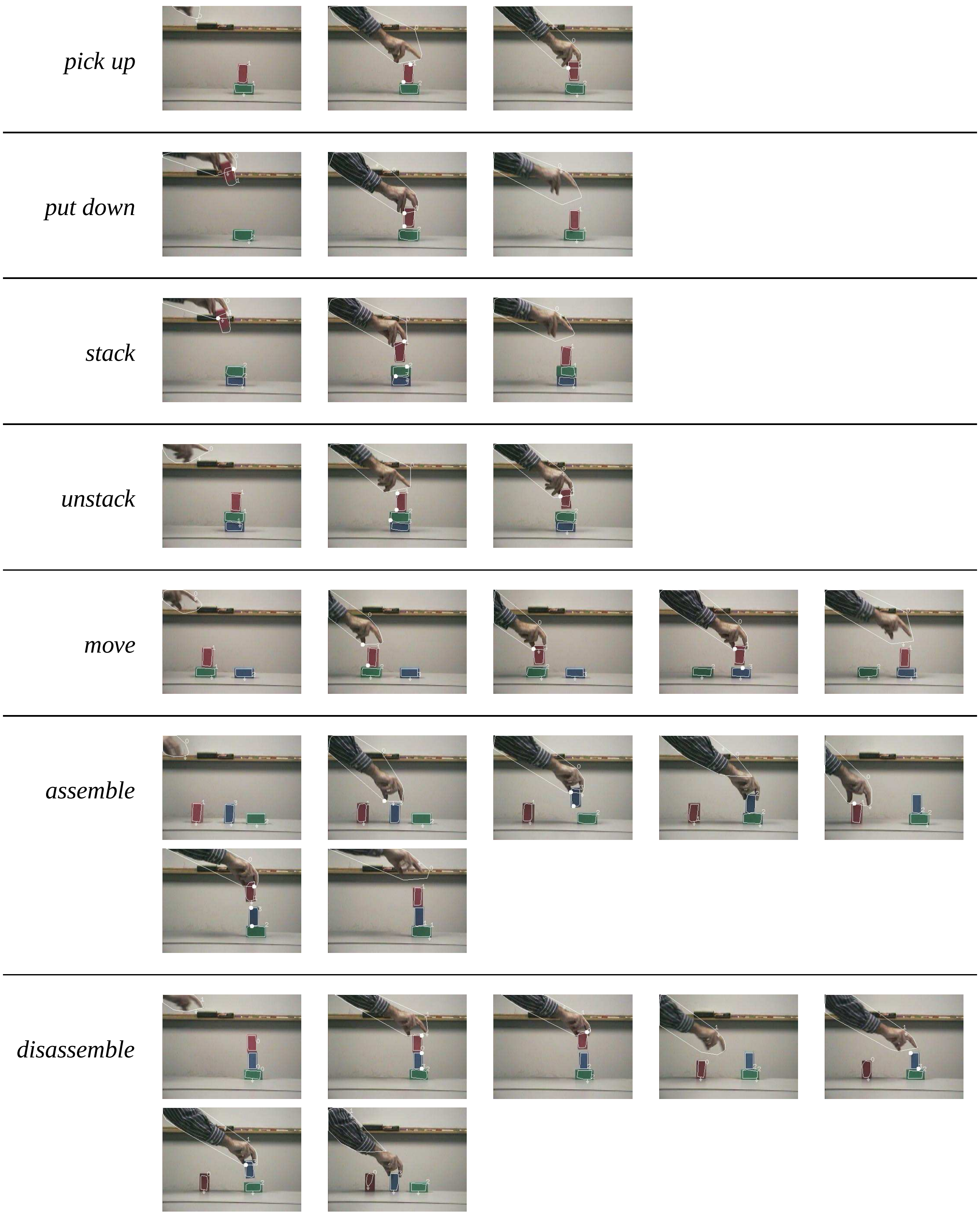

Figure 11: Key frames from sample videos of the 7 event types.

number of correct occurrences of the target event within all 210 movies (i.e., the human assessment of the number of occurrences of the target event). The normalized value of $\overline{\text{FP}}$ estimates the probability that the target event did not occur given that it was predicted to occur, while the normalized value of $\overline{\text{FN}}$ estimates the probability that the event was not predicted to occur given that it did occur.

### 6.3 Results

To evaluate our $k$-AMA learning approach, we ran leave-one-movie-out experiments, as described above, for varying $k$, $D$, and $N$. The 210 example movies were recorded with color-coded objects to provide complete object-correspondence information. We compared our learned event definitions to the performance of two sets of hand-coded definitions. The first set $\text{HD}_1$ of hand-coded definitions appeared in Siskind (2001). In response to subsequent deeper understanding of the behavior of LEONARD's model-reconstruction methods, we manually revised these definitions to yield another set $\text{HD}_2$ of hand-coded definitions that gives a significantly better $\overline{\text{FN}}$ performance at some cost in $\overline{\text{FP}}$ performance. Appendix C gives the event definitions in $\text{HD}_1$ and $\text{HD}_2$ along with a set of machine-generated definitions, produced by the $k$-AMA learning algorithm, given all training data for $k = 30$ and $D = \text{BN}$.

#### 6.3.1 OBJECT CORRESPONDENCE

To evaluate our algorithm for finding object correspondences, we ignored the correspondence information provided by color coding and applied the algorithm to all training models for each event type. The algorithm selected the correct correspondence for all 210 training models. Thus, for this data set, the learning results when no correspondence information is given will be identical to those where the correspondences are manually provided, except that, in the first case, the rules will not specify particular object roles, as discussed in section 5.3.2. Since our evaluation procedure uses role information, the rest of our experiments use the manual correspondence information, provided by color-coding, rather than computing it.

While our correspondence technique was perfect in these experiments, it may not be suited to some event types. Furthermore, it is likely to produce more errors as noise levels increase. Since correspondence errors represent a form of noise and our learner makes no special provisions for handling noise, the results are likely to be poor when such errors are common. For example, in the worst case, it is possible for a single extremely noisy example to cause the the LGG to be trivial (i.e., the formula **true**). In such cases, we will be forced to improve the noise tolerance of our learner.

#### 6.3.2 VARYING $k$

The first three rows of Table 1 show the $\overline{\text{FP}}$ and $\overline{\text{FN}}$ values for all 7 event types for $k \in \{2, 3, 4\}$, $N = 29$ (the maximum), and $D = \text{BN}$. Similar trends were found for $D = \text{P}$ and $D = \text{N}$. The general trend is that, as $k$ increases, $\overline{\text{FP}}$ decreases or remains the same and $\overline{\text{FN}}$ increases or remains the same. Such a trend is a consequence of our $k$-cover approach. This is because, as $k$ increases, the $k$-AMA language contains strictly more formulas. Thus for $k_1 > k_2$, the $k_1$-cover of a formula will never be more general than the $k_2$-cover. This strongly suggests, but does not prove, that $\overline{\text{FP}}$ will be non-increasing with $k$ and $\overline{\text{FN}}$ will be non-decreasing with $k$.

Our results show that 2-AMA is overly general for *put down* and *assemble*, i.e. it gives high $\overline{\text{FP}}$. In contrast, 3-AMA achieves $\overline{\text{FP}} = 0$ for each event type, but pays a penalty in $\overline{\text{FN}}$ compared to 2-AMA. Since 3-AMA achieves $\overline{\text{FP}} = 0$, there is likely no advantage in moving to $k$-AMA for $k > 3$. That is, the expected result is for $\overline{\text{FN}}$ to become larger. This effect is demonstrated for 4-AMA in the table.

| $k$ | $D$ | | *pick up* | *put down* | *stack* | *unstack* | *move* | *assemble* | *disassemble* |
|---|---|---|---|---|---|---|---|---|---|
| 2 | BN | $\overline{\text{FP}}$ | 0 | 0.14 | 0 | 0 | 0 | 0.75 | 0 |
| | | $\overline{\text{FN}}$ | 0 | 0.19 | 0.12 | 0.03 | 0 | 0 | 0 |
| 3 | BN | $\overline{\text{FP}}$ | 0 | 0 | 0 | 0 | 0 | 0 | 0 |
| | | $\overline{\text{FN}}$ | 0 | 0.2 | 0.45 | 0.10 | 0.03 | 0.07 | 0.10 |
| 4 | BN | $\overline{\text{FP}}$ | 0 | 0 | 0 | 0 | 0 | 0 | 0 |
| | | $\overline{\text{FN}}$ | 0 | 0.2 | 0.47 | 0.12 | 0.03 | 0.07 | 0.17 |
| 3 | P | $\overline{\text{FP}}$ | 0.42 | 0.5 | 0 | 0.02 | 0 | 0 | 0 |
| | | $\overline{\text{FN}}$ | 0 | 0.19 | 0.42 | 0.11 | 0.03 | 0.03 | 0.10 |
| 3 | BN | $\overline{\text{FP}}$ | 0 | 0 | 0 | 0 | 0 | 0 | 0 |
| | | $\overline{\text{FN}}$ | 0 | 0.2 | 0.45 | 0.10 | 0.03 | 0.07 | 0.10 |
| 3 | N | $\overline{\text{FP}}$ | 0 | 0 | 0 | 0 | 0 | 0 | 0 |
| | | $\overline{\text{FN}}$ | 0.04 | 0.39 | 0.58 | 0.16 | 0.13 | 0.2 | 0.2 |
| $\text{HD}_1$ | | $\overline{\text{FP}}$ | 0.01 | 0.01 | 0 | 0 | 0 | 0 | 0 |
| | | $\overline{\text{FN}}$ | 0.02 | 0.22 | 0.82 | 0.62 | 0.03 | 1.0 | 0.5 |
| $\text{HD}_2$ | | $\overline{\text{FP}}$ | 0.13 | 0.11 | 0 | 0 | 0 | 0 | 0 |
| | | $\overline{\text{FN}}$ | 0.0 | 0.19 | 0.42 | 0.02 | 0.0 | 0.77 | 0.0 |

Table 1: $\overline{\text{FP}}$ and $\overline{\text{FN}}$ for learned definitions, varying both $k$ and $D$, and for hand-coded definitions.

#### 6.3.3 VARYING $D$

Rows four through six of Table 1 show $\overline{\text{FP}}$ and $\overline{\text{FN}}$ for all 7 event types for $D \in \{\text{P}, \text{BN}, \text{N}\}$, $N = 29$, and $k = 3$. Similar trends were observed for other values of $k$. The general trend is that, as the degree of negative information increases, the learned event definitions become more specific. In other words, $\overline{\text{FP}}$ decreases and $\overline{\text{FN}}$ increases. This makes sense since, as more negative information is added to the training models, more specific structure can be found in the data and exploited by the $k$-AMA formulas. We can see that, with $D = \text{P}$, the definitions for *pick up* and *put down* are overly general, as they produce high $\overline{\text{FP}}$. Alternatively, with $D = \text{N}$, the learned definitions are overly specific, giving $\overline{\text{FP}} = 0$, at the cost of high $\overline{\text{FN}}$. In these experiments, as well as others, we have found that $D = \text{BN}$ yields the best of both worlds: $\overline{\text{FP}} = 0$ for all event types and lower $\overline{\text{FN}}$ than achieved with $D = \text{N}$.

Experiments not shown here have demonstrated that, without negation for *pick up* and *put down*, we can increase $k$ arbitrarily, in an attempt to specialize the learned definitions, and never significantly reduce $\overline{\text{FP}}$. This indicates that negative information plays a particularly important role in constructing definitions for these event types.

#### 6.3.4 COMPARISON TO HAND-CODED DEFINITIONS

The bottom two rows of table 1 show the results for $\text{HD}_1$ and $\text{HD}_2$. We have not yet attempted to automatically select the parameters for learning (i.e. $k$ and $D$). Rather, here we focus on comparing the hand-coded definitions to the parameter set that we judged to be best performing across all event types. We believe, however, that these parameters could be selected reliably using cross-validation techniques applied to a larger data set. In that case, the parameters would be selected on a per-event-type basis and would likely result in an even more favorable comparison to the hand-coded definitions.

The results show that the learned definitions significantly outperform $\text{HD}_1$ on the current data set. The $\text{HD}_1$ definitions were found to produce a large number of false negatives on the current data set. Notice that, although $\text{HD}_2$ produces significantly fewer false negatives for all event types, it produces more false positives for *pick up* and *put down*. This is because the hand definitions utilize *pick up* and *put down* as macros for defining the other events.

The performance of the learned definitions is competitive with the performance of $\text{HD}_2$. The main differences in performance are: (a) for *pick up* and *put down*, the learned and $\text{HD}_2$ definitions achieve nearly the same $\overline{\text{FN}}$ but the learned definitions achieve $\overline{\text{FP}} = 0$ whereas $\text{HD}_2$ has significant $\overline{\text{FP}}$, (b) for *unstack* and *disassemble*, the learned definitions perform moderately worse than $\text{HD}_2$ with respect to $\overline{\text{FN}}$, and (c) the learned definitions perform significantly better than $\text{HD}_2$ on *assemble* events.

We conjecture that further manual revision could improve $\text{HD}_2$ to perform as well as (and perhaps better than) the learned definitions for every event class. Nonetheless, we view this experiment as promising, as it demonstrates that our learning technique is able to compete with, and sometimes outperform, significant hand-coding efforts by one of the authors.

#### 6.3.5 VARYING $N$

It is of practical interest to know how training-set size affects our algorithm's performance. For this application, it is important that our method work well with fairly small data sets, as it can be tedious to collect event data. Table 2 shows the $\overline{\text{FN}}$ of our learning algorithm for each event type, as $N$ is reduced from 29 to 5. For these experiments, we used $k = 3$ and $D = \text{BN}$. Note that $\overline{\text{FP}} = 0$ for all event types and all $N$ and hence is not shown. We expect $\overline{\text{FN}}$ to increase as $N$ is decreased, since, with specific-to-general learning, more data yields more-general definitions. Generally, $\overline{\text{FN}}$ is flat for $N > 20$, increases slowly for $10 < N < 20$, and increases abruptly for $5 < N < 10$. We also see that, for several event types, $\overline{\text{FN}}$ decreases slowly, as $N$ is increased from 20 to 29. This indicates that a larger data set might yield improved results for those event types.

#### 6.3.6 PERSPICUITY OF LEARNED DEFINITIONS

One motivation for using a logic-based event representation is to support perspicuity—in this respect our results are mixed. We note that perspicuity is a fuzzy and subjective concept. Realizing this, we will say that an event definition is *perspicuous* if most humans with knowledge of the language would find the definition to be "natural." Here, we do not assume the human has a detailed knowledge of the model-reconstruction process that our learner is trying to fit. Adding that assumption would presumably make the definitions qualify as more perspicuous, as many of the complex features of the learned definitions appear in fact to be due to idiosyncrasies of the model-reconstruction process. In this sense, we are evaluating the perspicuity of the output of the entire system, not just of the learner itself, so that a key route to improving perspicuity in this sense would be to improve the intuitive properties of the model-reconstruction output without any change to the learner.

While the learned and hand-coded definitions are similar with respect to accuracy, typically the learned definitions are much less perspicuous. For our simplest event types, however, the learned definitions are arguably perspicuous. Below we look at this issue in more detail. Appendix C gives the hand-coded definitions in $HD_1$ and $HD_2$ along with a set of machine-generated definitions. The learned definitions correspond to the output of our $k$-AMA learner when run on all 30 training movies from each event type with $k = 3$ and $D = \text{BN}$ (i.e., our best performing configuration with respect to accuracy).

**Perspicuous Definitions.** The PICKUP$(x, y, z)$ and PUTDOWN$(x, y, z)$ definitions are of particular interest here since short state sequences appear adequate for representing these event types—thus, we can hope for perspicuous 3-AMA definitions. In fact, the hand-coded definitions involve short sequences. Consider the hand-coded definitions of PICKUP$(x, y, z)$—the definitions can roughly be viewed as 3-MA timelines of the form *begin*;*trans*;*end*.[14] State *begin* asserts facts that indicate $y$ is on $z$ and is not being held by $x$ and *end* asserts facts that indicate $y$ is being held by $x$ and is not on $z$. State *trans* is intended to model the fact that LEONARD's model-reconstruction process does not always handle the transition between *begin* and *end* smoothly (so the definition *begin*;*end* does not work well). We can make similar observations for PUTDOWN$(x, y, z)$.

Figure 15 gives the learned 3-AMA definitions of PICKUP$(x, y, z)$ and PUTDOWN$(x, y, z)$—the definitions contain six and two 3-MA timelines respectively. Since the definitions consists of multiple parallel timelines, they may at first not seem perspicuous. However, a closer examination reveals that, in each definition, there is a single timeline that is arguably perspicuous—we have placed these *perspicuous timelines* at the beginning of each definition. The perspicuous timelines have a natural *begin*;*trans*;*end* interpretation. In fact, they are practically equivalent to the definitions of PICKUP$(x, y, z)$ and PUTDOWN$(x, y, z)$ in $HD_2$.[15]

With this in mind, notice that the $HD_2$ definitions are overly general as indicated by significant false positive rates. The learned definitions, however, yield no false positives without a significant increase in false negatives. The learned definitions improve upon $HD_2$ by essentially specializing the $HD_2$ definitions (i.e., the perspicuous timelines) by conjoining them with the *non-perspicuous timelines*. While these non-perspicuous timelines are often not intuitive, they capture patterns in the events that help rule out non-events. For example, in the learned definition of PICKUP$(x, y, z)$ some of the non-perspicuous timelines indicate that ATTACHED$(y, z)$ is true during the transition period of the event. Such an attachment relationship does not make intuitive sense. Rather, it represents a systematic error made by the model reconstruction process for *pick up* events.

In summary, we see that the learned definitions of PICKUP$(x, y, z)$ and PUTDOWN$(x, y, z)$ each contain a perspicuous timeline and one or more non-perspicuous timelines. The perspicuous timelines give an intuitive definition of the events, whereas the non-perspicuous timelines capture non-intuitive aspects of the events and model reconstruction process that are important in practice. We note that, for experienced users, the primary difficulty of hand-coding definitions for LEONARD is

14. Note that the event-logic definition for PICKUP$(x, y, z)$ in $HD_2$ is written in a more compact form than 3-MA, but this definition can be converted to 3-MA (and hence 3-AMA). Rather, $HD_1$ cannot be translated exactly to 3-MA since it uses disjunction—it is the disjunction of two 3-MA timelines.

15. The primary difference is that the $HD_2$ definitions contain more negated propositions. The learner only considers a proposition and its negation if the proposition is true at some point during the training movies. Many of the negated propositions in $HD_2$ never appear positively, thus they are not included in the learned definitions.

to determining which non-perspicuous properties must be included. Typically this requires many iterations of trial and error. Our automated technique can relieve the user of this task. Alternatively, we could view the system as providing guidance for this task.

**Large Definitions.** The STACK$(w, x, y, z)$ and UNSTACK$(w, x, y, z)$ events are nearly identical to *put down* and *pick up* respectively. The only difference is that now we are picking up from and putting down onto a two block (rather than single block) tower (i.e., composed of blocks $y$ and $z$). Thus, here again we might expect there to be perspicuous 3-AMA definitions. However, we see that the learned definitions for STACK$(w, x, y, z)$ and UNSTACK$(w, x, y, z)$ in Figures 16 and 17 involve many more timelines than those for PICKUP$(w, x, y)$ and PUTDOWN$(w, x, y)$. Accordingly, the definitions are quite overwhelming and much less perspicuous.

Despite the large number of timelines, these definitions have the same general structure as those for *pick up* and *put down*. In particular, they each contain a distinguished perspicuous timeline, placed at the beginning of each definition, that is conjoined with many non-perspicuous timelines. It is clear that, as above, the perspicuous timelines have a natural *begin*;*trans*;*end* interpretation and, again, they are very similar to the definitions in $\mathrm{HD}_2$. In this case, however, the definitions in $\mathrm{HD}_2$ are not overly general (committing no false positives). Thus, here the inclusion of the non-perspicuous timelines has a detrimental effect since they unnecessarily specialize the definition resulting in more false negatives.

We suspect that a primary reason for the large number of non-perspicuous timelines relative to the definitions of *pick up* and *put down* stems from the increased difficulty of constructing force-dynamic models. The inclusion of the two block tower in these examples causes the model-reconstruction process to produce more unintended results, particularly during the transition periods of STACK and UNSTACK. The result is that often many unintuitive and physically incorrect patterns involving the three blocks and the hand are produced during the transition period. The learner captures these patterns roughly via the non-perspicuous timelines. It is likely that generalizing the definitions by including more training examples would filter out some of these timelines, making the overall definition more perspicuous. Alternatively, it is of interest to consider pruning the learned definitions. A straightforward way to do this is to generate negative examples. Then with these, we could remove timelines (generalizing the definition) that do not contribute toward rejecting the negative examples. It is unclear how to prune definitions without negative examples.

**Hierarchical Events.** MOVE$(w, x, y, z)$, ASSEMBLE$(w, x, y, z)$, and DISASSEMBLE$(w, x, y, z)$ are inherently hierarchical, being composed of the four simpler event types. The hand-coded definitions leverage this structure by utilizing the simpler definitions as macros. In this light, it should be clear that, when viewed non-hierarchically, (as our learner does) these events involve relatively long state sequences. Thus, 3-AMA is not adequate for writing down perspicuous definitions. In spite of this representational shortcoming, our learned 3-AMA definitions perform quite well. This performance supports one of our arguments for using AMA from section 3.2. Namely, given that it is easier to find short rather than long sequences, a practical approach to finding definitions for long events is to conjoin the short sequences within those events. Examining the timelines of the learned 3-AMA definitions reveals what we might expect. Each timeline captures an often understandable property of the long event sequence, but the conjunction of those timelines cannot be considered to be a perspicuous definition. A future direction is to utilize hierarchical learning techniques to improve the perspicuity of our definitions while maintaining accuracy.

| $N$ | *pick up* | *put down* | *stack* | *unstack* | *move* | *assemble* | *disassemble* |
|---|---|---|---|---|---|---|---|
| 29 | 0.0 | 0.20 | 0.45 | 0.10 | 0.03 | 0.07 | 0.10 |
| 25 | 0.0 | 0.20 | 0.47 | 0.16 | 0.05 | 0.09 | 0.10 |
| 20 | 0.01 | 0.21 | 0.50 | 0.17 | 0.08 | 0.12 | 0.12 |
| 15 | 0.01 | 0.22 | 0.53 | 0.26 | 0.14 | 0.20 | 0.16 |
| 10 | 0.07 | 0.27 | 0.60 | 0.36 | 0.23 | 0.32 | 0.26 |
| 5 | 0.22 | 0.43 | 0.77 | 0.54 | 0.35 | 0.57 | 0.43 |

Table 2: $\overline{\text{FN}}$ for $k = 3$, $D = \text{BN}$, and various values of $N$.

We note, however, that, at some level, the learned definition of $\text{MOVE}(w, x, y, z)$ given in Figure 18 is perspicuous. In particular, the first 3-MA timeline is naturally interpreted as giving the pre- and post-conditions for a move action. That is, initially $x$ is supported by $y$ and the hand $w$ is empty and finally $x$ is supported by $z$ and the hand $w$ is empty. Thus, if all we care about is pre- and post-conditions, we might consider this timeline to be perspicuous. The remaining timelines in the definition capture pieces of the internal event structure such as facts indicating that $x$ is moved by the hand. A weaker case can be made for *assemble* and *disassemble*. The first timeline in each of the learned definitions in Figures 19 and 20 can be interpreted as giving pre- and post-conditions. However, in these cases, the pre(post)-conditions for *assemble*(*disassemble*) are quite incomplete. The incompleteness is due to the inclusion of examples where the model-reconstruction process did not properly handle the initial(final) moments.

## 7. Related Work

Here we discuss two bodies of related work. First, we present previous work in visual event recognition and how it relates to our experiments here. Second, we discuss previous approaches to learning temporal patterns from positive data.

### 7.1 Visual Event Recognition

Our system is unique in that it combines positive-only learning with a temporal, relational, and force-dynamic representation to recognize events from real video. Prior work has investigated various subsets of the features of our system—but, to date, no system has combined all of these pieces together. Incorporating any one of these pieces into a system is a significant endeavor. In this respect, there are no competing approaches to directly compare our system against. Given this, the following is a representative list of systems that have common features with ours. It is not meant to be comprehensive and focuses on pointing out the primary differences between each of these systems and ours, as these primary differences actually render these systems only very loosely related to ours.

Borchardt (1985) presents a representation for temporal, relational, force-dynamic event definitions but these definitions are neither learned nor applied to video. Regier (1992) presents techniques for learning temporal event definitions but the learned definitions are neither relational, force dynamic, nor applied to video. In addition the learning technique is not truly positive-only—rather, it extracts implicit negative examples of an event type from positive examples of other event types.

Yamoto, Ohya, and Ishii (1992), Brand and Essa (1995), Siskind and Morris (1996), Brand, Oliver, and Pentland (1997), and Bobick and Ivanov (1998) present techniques for learning temporal event definitions from video but the learned definitions are neither relational nor force dynamic. Pinhanez and Bobick (1995) and Brand (1997a) present temporal, relational event definitions that recognize events in video but these definitions are neither learned nor force dynamic. Brand (1997b) and Mann and Jepson (1998) present techniques for analyzing force dynamics in video but neither formulate event definitions nor apply these techniques to recognizing events or learning event definitions.

### 7.2 Learning Temporal Patterns

We divide this body of work into three main categories: temporal data mining, inductive logic programming, and finite-state–machine induction.

**Temporal Data Mining.** The sequence-mining literature contains many general-to-specific ("levelwise") algorithms for finding frequent sequences (Agrawal & Srikant, 1995; Mannila, Toivonen, & Verkamo, 1995; Kam & Fu, 2000; Cohen, 2001; Hoppner, 2001). Here we explore a specific-to-general approach. In this previous work, researchers have studied the problem of mining temporal patterns using languages that are interpreted as placing constraints on partially or totally ordered sets of time points, e.g., sequential patterns (Agrawal & Srikant, 1995) and episodes (Mannila et al., 1995). These languages place constraints on time points rather than time intervals as in our work here. More recently there has been work on mining temporal patterns using interval-based pattern languages (Kam & Fu, 2000; Cohen, 2001; Hoppner, 2001).

Though the languages and learning frameworks vary among these approaches, they share two central features which distinguish them from our approach. First, they all typically have the goal of finding all frequent patterns (formulas) within a temporal data set—our approach is focused on finding patterns with a frequency of one (covering all positive examples). Our first learning application of visual-event recognition has not yet required us to find patterns with frequency less than one. However, there are a number of ways in which we can extend our method in that direction when it becomes necessary (e.g., to deal with noisy training data). Second, these approaches all use standard general-to-specific level-wise search techniques, whereas we chose to take a specific-to-general approach. One direction for future work is to develop a general-to-specific level-wise algorithm for finding frequent MA formulas and to compare it with our specific-to-general approach. Another direction is to design a level-wise version of our specific-to-general algorithm—where for example, the results obtained for the $k$-AMA LGG can be used to more efficiently calculate the $(k+1)$-AMA LGG. Whereas a level-wise approach is conceptually straightforward in a general-to-specific framework it is not so clear in the specific-to-general case. We are not familiar with other temporal data-mining systems that take a specific-to-general approach.

**First-Order Learning** In Section 3.3, we pointed out difficulties in using existing first-order clausal generalization techniques for learning AMA formulas. In spite of these difficulties, it is still possible to represent temporal events in first-order logic (either with or without capturing the AMA semantics precisely) and to apply general-purpose relational learning techniques, e.g., inductive logic programming (ILP) (Muggleton & De Raedt, 1994). Most ILP systems require both positive and negative training examples and hence are not suitable for our current positive-only framework. Exceptions include GOLEM (Muggleton & Feng, 1992), PROGOL (Muggleton, 1995), and CLAUDIEN (De Raedt & Dehaspe, 1997), among others. While we have not performed a full evaluation of these systems, our early experiments in the visual-event recognition domain confirmed our belief that horn clauses, lacking special handling of time, give a poor inductive bias. In particular, many of the learned clauses find patterns that simply do not make sense from a temporal perspective and, in turn, generalize poorly. We believe a reasonable alternative to our approach may be to incorporate syntactic biases into ILP systems as done, for example, in Cohen (1994), Dehaspe and De Raedt (1996), Klingspor, Morik, and Rieger (1996). In this work, however, we chose to work directly in a temporal logic representation.

| | Subsumption | | Semantic AMA LGG | | | Syntactic AMA LGG | | |
|---|---|---|---|---|---|---|---|---|
| Inputs | Semantic | Syntactic | Lower | Upper | Size | Lower | Upper | Size |
| MA | P | P | P | coNP | EXP | P | coNP | EXP |
| AMA | coNP-complete | P | coNP | NEXP | 2-EXP? | P | coNP | EXP |

Table 3: Complexity Results Summary. The LGG complexities are relative to *input plus output* size. The size column reports the worst-case smallest correct output size. The "?" indicates a conjecture.

**Finite-State Machines** Finally, we note there has been much theoretical and empirical research into learning finite-state machines (FSMs) (Angluin, 1987; Lang, Pearlmutter, & Price, 1998). We can view FSMs as describing properties of strings (symbol sequences). In our case, however, we are interested in describing sequences of propositional models rather than just sequences of symbols. This suggests learning a type of "factored" FSM where the arcs are labeled by sets of propositions rather than by single symbols. Factored FSMs may be a natural direction in which to extend the expressiveness of our current language, for example by allowing repetition. We are not aware of work concerned with learning factored FSMs; however, it is likely that inspiration can be drawn from symbol-based FSM-learning algorithms.

## 8. Conclusion

We have presented a simple logic for representing temporal events called AMA and have shown theoretical and empirical results for learning AMA formulas. Empirically, we've given the first system for learning temporal, relational, force-dynamic event definitions from positive-only input and we have applied that system to learn such definitions from real video input. The resulting performance matches that of event definitions that are hand-coded with substantial effort by human domain experts. On the theoretical side, Table 3 summarizes the upper and lower bounds that we have shown for the subsumption and generalization problems associated with this logic. In each case, we have provided a provably correct algorithm matching the upper bound shown. The table also shows the worst-case size that the smallest LGG could possibly take relative to the input size, for both AMA and MA inputs. The key results in this table are the polynomial-time MA subsumption and AMA syntactic subsumption, the coNP lower bound for AMA subsumption, the exponential size of LGGs in the worst case, and the apparently lower complexity of syntactic AMA LGG versus semantic LGG. We described how to build a learner based on these results and applied it to the visual-event learning domain. To date, however, the definitions we learn are neither cross-modal nor perspicuous. And while the performance of the learned definitions matches that of hand-coded ones, we wish to surpass hand coding. In the future, we intend to address cross-modality by applying our learning technique to the planning domain. We also believe that addressing perspicuity will lead to improved performance.

## Acknowledgments

The authors wish to thank our anonymous reviewers for helping to improve this paper. This work was supported in part by NSF grants 9977981-IIS and 0093100-IIS, an NSF Graduate Fellowship for Fern, and the Center for Education and Research in Information Assurance and Security at Purdue University. Part of this work was performed while Siskind was at NEC Research Institute, Inc.

## Appendix A. Internal Positive Event Logic

Here we give the syntax and semantics for an event logic called *Internal Positive Event Logic (IPEL)*. This logic is used in the main text only to motivate our choice of a small subset of this logic, AMA, by showing, in Proposition 4, that AMA can define any set of models that IPEL can define.

An event type (i.e., set of models) is said to be *internal* if whenever it contains any model $\mathcal{M} = \langle M, I \rangle$, it also contains any model that agrees with $\mathcal{M}$ on truth assignments $M[i]$ where $i \in I$. Full event logic allows the definition of non-internal events, for example, the formula $\Psi = \Diamond_{<} P$ is satisfied by $\langle M, I \rangle$ when there is some interval $I'$ *entirely preceding* $I$ such that $P$ is satisfied by $\langle M, I' \rangle$, thus $\Psi$ is not internal. The applications we are considering do not appear to require non-internal events, thus we currently only consider events that are internal.

Call an event type *positive* if it contains the model $\mathcal{M} = \langle M, [1,1] \rangle$ where $M(1)$ is the truth assignment assigning all propositions the value true. A positive event type cannot require any proposition to be false at any point in time.

IPEL is a fragment of full propositional event logic that can only describe positive internal events. We conjecture, but have not yet proven, that all positive internal events representable in the full event logic of Siskind (2001) can be represented by some IPEL formula. Formally, the syntax of IPEL formulas is given by

$$E ::= \mathbf{true} \mid \textit{prop} \mid E_1 \vee E_2 \mid \Diamond_{R'} E_1 \mid E_1 \wedge_R E_2,$$

where the $E_i$ are IPEL formulas, *prop* is a primitive proposition (sometimes called a primitive event type), $R$ is a subset of the thirteen Allen interval relations $\{\mathsf{s,f,d,b,m,o,=,si,fi,di,bi,ai,oi}\}$ (Allen, 1983), and $R'$ is a subset of the restricted set of Allen relations $\{\mathsf{s,f,d,=}\}$, the semantics for each Allen relation is given in Table 4. The difference between IPEL syntax and that of full propositional event logic is that event logic allows for a negation operator, and that, in full event logic, $R'$ can be any subset of all thirteen Allen relations. The operators $\wedge$ and ; used to define AMA formulas are merely abbreviations for the IPEL operators $\wedge_{\{=\}}$ and $\wedge_{\{\mathsf{m}\}}$ respectively, so AMA is a subset of IPEL (though a distinguished subset as indicated by Proposition 4).

Each of the thirteen Allen interval relations are binary relations on the set of closed natural-number intervals. Table 4 gives the definitions of these relations, defining $[m_1, m_2]\ r\ [n_1, n_2]$ for each Allen relation $r$. Satisfiability for IPEL formulas can now be defined as follows,

| $I_1$ | Relation | $I_2$ | English | Definition | Inverse |
|---|---|---|---|---|---|
| $[m_1, m_2]$ | s | $[n_1, n_2]$ | starts | $m_1 = n_1$ and $m_2 \leq n_2$ | si |
| $[m_1, m_2]$ | f | $[n_1, n_2]$ | finishes | $m_1 \leq n_1$ and $m_2 = n_2$ | fi |
| $[m_1, m_2]$ | d | $[n_1, n_2]$ | during | $m_1 \geq n_1$ and $m_2 \leq n_2$ | di |
| $[m_1, m_2]$ | b | $[n_1, n_2]$ | before | $m_2 \leq n_1$ | bi |
| $[m_1, m_2]$ | m | $[n_1, n_2]$ | meets | $m_2 = n_1$ or $m_2 + 1 = n_1$ | mi |
| $[m_1, m_2]$ | o | $[n_1, n_2]$ | overlaps | $m_1 \leq n_1 \leq m_2 \leq n_2$ | oi |
| $[m_1, m_2]$ | = | $[n_1, n_2]$ | equals | $m_1 = n_1$ and $m_2 = n_2$ | = |

Table 4: The Thirteen Allen Relations (adapted to our semantics).

- **true** is satisfied by every model.
- *prop* is satisfied by model $\langle M, I \rangle$ iff $M[x]$ assigns *prop* true for every $x \in I$.
- $E_1 \vee E_2$ is satisfied by a model $\mathcal{M}$ iff $\mathcal{M}$ satisfies $E_1$ or $\mathcal{M}$ satisfies $E_2$.
- $\Diamond_R E$ is satisfied by model $\langle M, I \rangle$ iff for some $r \in R$ there is an interval $I'$ such that $I'\ r\ I$ and $\langle M, I' \rangle$ satisfies $E$.
- $E_1 \wedge_R E_2$ is satisfied by model $\langle M, I \rangle$ iff for some $r \in R$ there exist intervals $I_1$ and $I_2$ such that $I_1\ r\ I_2$, $\text{SPAN}(I_1, I_2) = I$ and both $\langle M, I_1 \rangle$ satisfies $E_1$ and $\langle M, I_2 \rangle$ satisfies $E_2$.

where *prop* is a primitive proposition, $E$ and $E_i$ are IPEL formulas, $R$ is a set of Allen relations, and $\text{SPAN}(I_1, I_2)$ is the minimal interval that contains both $I_1$ and $I_2$. From this definition, it is easy to show, by induction on the number of operators and connectives in a formula, that all IPEL formulas define internal events. One can also verify that the definition of satisfiability given earlier for AMA formulas corresponds to the one we give here.

## Appendix B. Omitted Proofs

**Lemma 1.** *For any MA timeline $\Phi$ and any model $\mathcal{M}$, if $\mathcal{M}$ satisfies $\Phi$ then there is a witnessing interdigitation for MAP($\mathcal{M}$) $\leq \Phi$.*

**Proof**: Assume that $\mathcal{M} = \langle M, I \rangle$ satisfies the MA timeline $\Phi = s_1; \ldots; s_n$, and let $\Phi' =$ MAP($\mathcal{M}$). It is straightforward to argue, by induction on the length of $\Phi$, that there exists a mapping $V'$ from states of $\Phi$ to sub-intervals of $I$, such that

- for any $i \in V'(s)$, $M[i]$ satisfies $s$,
- $V'(s_1)$ includes the initial time point of $I$,
- $V'(s_n)$ includes the final time point of $I$, and
- for any $i \in [1, n-1]$, we have $V'(s_i)$ meets $V'(s_{i+1})$ (see Table 4).

Let $V$ be the relation between states $s \in \Phi$ and members $i \in I$ that is true when $i \in V'(s)$. Note that the conditions on $V'$ ensure that every $s \in \Phi$ and every $i \in I$ appear in some tuple in $V$ (not necessarily together). Below we use $V$ to construct a witnessing interdigitation $W$.

Let $R$ be the total, one-to-one, onto function from time-points in $I$ to corresponding states in $\Phi'$, noting that $\Phi'$ has one state for each time-point in $I$, as $\Phi' = \text{MAP}(\langle M, I\rangle)$. Note that $R$ preserves ordering in that, when $i \leq j$, $R(i)$ is no later than $R(j)$ in $\Phi'$. Let $W$ be the composition $V \circ R$ of the relations $V$ and $R$.

We show that $W$ is an interdigitation. We first show that each state from $\Phi$ or $\Phi'$ appears in a tuple in $W$, so $W$ is piecewise total. States from $\Phi$ must appear, trivially, because each appears in a tuple of $V$, and $R$ is total. States from $\Phi'$ appear because each $i \in I$ appears in a tuple of $V$, and $R$ is onto the states of $\Phi'$.

It now suffices to show that for any states $s$ before $t$ from $\Phi$, $W(s, s')$ and $W(t, t')$ implies that $s'$ is no later than $t'$ in $\Phi'$, so that $W$ is simultaneously consistent. The conditions defining $V'$ above imply that every number in $i \in V(s)$ is less than or equal to every $j \in V(t)$. The order-preservation property of $R$, noted above, then implies that every state $s' \in V \circ R(s)$ is no later than any state $t' \in V \circ R(t)$ in $\Phi'$, as desired. So $W$ is an interdigitation.

We now argue that $W$ witnesses $\Phi' \leq \Phi$. Consider $s \in \Phi$ and $t \in \Phi'$ such that $W(s, t)$. By the construction of $W$, there must be $i \in V'(s)$ for which $t$ is the $i$'th state of $\Phi'$. Since $\Phi' = \text{MAP}(\mathcal{M})$, it follows that $t$ is the set of true propositions in $M[i]$. Since $i \in V'(s)$, we know that $M[i]$ satisfies $s$. It follows that $s \subseteq t$, and so $t \leq s$. $\square$

**Lemma 3.** *For any $E \in$ IPEL, if model $\mathcal{M}$ embeds a model that satisfies $E$ then $\mathcal{M}$ satisfies $E$.*

**Proof**: Consider the models $\mathcal{M} = \langle M, I\rangle$ and $\mathcal{M}' = \langle M', I'\rangle$ such that $\mathcal{M}$ embeds $\mathcal{M}'$, let $\Phi = \text{MAP}(\mathcal{M})$ and $\Phi' = \text{MAP}(\mathcal{M}')$. Assume that $E \in$ IPEL is satisfied by $\mathcal{M}'$, we will show that $E$ is also satisfied by $\mathcal{M}$.

We know from the definition of embedding that $\Phi \leq \Phi'$ and thus there is a witnessing interdigitation $W$ for $\Phi \leq \Phi'$ by Proposition 2. We know there is a one-to-one correspondence between numbers in $I$ ($I'$) and states of $\Phi$ ($\Phi'$) and denote the state in $\Phi$ ($\Phi'$) corresponding to $i \in I$ ($i' \in I'$) as $s_i$ ($t_{i'}$). This correspondence allows us to naturally interpret $W$ as a mapping $V$ from subsets of $I'$ to subsets of $I$ as follows: for $I'_1 \subseteq I'$, $V(I'_1)$ equals the set of all $i \in I$ such that for some $i' \in I'_1$, $s_i$ co-occurs with $t_{i'}$ in $W$. We will use the following properties of $V$,

1. If $I'_1$ is a sub-interval of $I'$, then $V(I'_1)$ is a sub-interval of $I$.
2. If $I'_1$ is a sub-interval of $I'$, then $\langle M, V(I'_1)\rangle$ embeds $\langle M', I'_1\rangle$.
3. If $I'_1$ and $I'_2$ are sub-intervals of $I'$, and $r$ is an Allen relation, then $I'_1 r I'_2$ iff $V(I'_1) r V(I'_2)$.
4. If $I'_1$ and $I'_2$ are sub-intervals of $I'$, then $V(\text{SPAN}(I'_1, I'_2)) = \text{SPAN}(V(I'_1), V(I'_2))$.
5. $V(I') = I$.

We sketch the proofs of these properties. 1) Use induction on the length of $I'_1$, with the definition of interdigitation. 2) Since $V(I'_1)$ is an interval, $\text{MAP}(\langle M, V(I'_1)\rangle)$ is well defined. $\text{MAP}(\langle M, V(I'_1)\rangle) \leq \text{MAP}(\langle M', I'_1\rangle)$ follows from the assumption that $\mathcal{M}$ embeds $\mathcal{M}'$. 3) From Appendix A, we see that all Allen relations are defined in terms of the $\leq$ relation on the natural number endpoints of the intervals. We can show that $V$ preserves $\leq$ (but not $<$) on singleton sets (i.e., every member of $V(\{i'\})$ is $\leq$ every member of $V(\{j'\})$ when $i' \leq j'$) and that $V$ commutes with set union. It follows that $V$ preserves the Allen interval relations. 4) Use the fact that $V$ preserves $\leq$ in the sense just argued, along with the fact that $\textsc{Span}(I'_1, I'_2)$ depends only on the minimum and maximum numbers in $I'_1$ and $I'_2$. 5) Follows from the definition of interdigitation and the construction of $V$.

We now use induction on the number of operators and connectives in $E$ to prove that, if $\mathcal{M}'$ satisfies $E$, then so must $\mathcal{M}$. The base case is when $E = \mathit{prop}$, where *prop* is a primitive proposition, or **true**. Since $\mathcal{M}'$ satisfies $E$, we know that *prop* is true in all $M'[x']$ for $x' \in I'$. Since $W$ witnesses $\Phi \leq \Phi'$, we know that, if *prop* is true in $M'[x]$, then *prop* is true in all $M[x]$, where $x \in V(x')$. Therefore, since $V(I') = I$, *prop* is true for all $M'[x]$, where $x \in I$, hence $\mathcal{M}'$ satisfies $E$.

For the inductive case, assume that the claim holds for IPEL formulas with fewer than $N$ operators and connectives—let $E_1, E_2$ be two such formulas. When $E = E_1 \vee E_2$, the claim trivially holds. When $E = \Diamond_R E_1$, $R$ must be a subset of the set of relations $\{\mathsf{s},\mathsf{f},\mathsf{d},=\}$. Notice that $E$ can be written as a disjunction of $\Diamond_r E_1$ formulas, where $r$ is a single Allen relation from $R$. Thus, it suffices to handle the case where $R$ is a single Allen relation. Suppose $E = \Diamond_{\{\mathsf{s}\}} E_1$. Since $\mathcal{M}'$ satisfies $E$, there must be a sub-interval $I'_1$ of $I'$ such that $I'_1\ \mathsf{s}\ I'$ and $\langle M', I'_1 \rangle$ satisfies $E_1$. Let $I_1 = V(I'_1)$, we know from the properties of $V$ that $V(I') = I$, and, hence, that $I_1\ \mathsf{s}\ I$. Furthermore, we know that $\langle M, I_1 \rangle$ embeds $\langle M', I'_1 \rangle$, and, thus, by the inductive hypothesis, $\langle M, I_1 \rangle$ satisfies $E_1$. Combining these facts, we get that $E$ is satisfied by $\mathcal{M}$. Similar arguments hold for the remaining three Allen relations. Finally, consider the case when $E = E_1 \wedge_R E_2$, where $R$ can be any set of Allen relations. Again, it suffices to handle the case when $R$ is a single Allen relation $r$. Since $\mathcal{M}'$ satisfies $E = E_1 \wedge_r E_2$, we know that there are sub-intervals $I'_1$ and $I'_2$ of $I'$ such that $\textsc{Span}(I'_1, I'_2) = I'$, $I'_1\ r\ I'_2$, $\langle M', I'_1 \rangle$ satisfies $E_1$, and $\langle M', I'_2 \rangle$ satisfies $E_2$. From these facts, and the properties of $V$, it is easy to verify that $\mathcal{M}$ satisfies $E$. □

**Lemma 5.** *Given an MA formula $\Phi$ that subsumes each member of a set $\Sigma$ of MA formulas, $\Phi$ also subsumes some member $\Phi'$ of* IG($\Sigma$). *Dually, when $\Phi$ is subsumed by each member of $\Sigma$, we have that $\Phi$ is also subsumed by some member $\Phi'$ of* IS($\Sigma$). *In each case, the length of $\Phi'$ can be bounded by the size of $\Sigma$.*

**Proof**: We prove the result for IG($\Sigma$). The proof for IS($\Sigma$) follows similar lines. Let $\Sigma = \{\Phi_1, \ldots, \Phi_n\}$, $\Phi = s_1; \ldots; s_m$, and assume that for each $1 \leq i \leq n$, $\Phi_i \leq \Phi$. From Proposition 2, for each $i$, there is a witnessing interdigitation $W_i$ for $\Phi_i \leq \Phi$. We will combine the $W_i$ into an interdigitation of $\Sigma$, and show that the corresponding member of IG($\Sigma$) is subsumed by $\Phi$. To construct an interdigitation of $\Sigma$, first notice that, for each $s_j$, each $W_i$ specifies a set of states (possibly a single state but at least one) from $\Phi_i$ that all co-occur with $s_j$. Furthermore, since $W_i$ is an interdigitation, it is easy to show that this set of states corresponds to a consecutive subsequence of states from $\Phi_i$—let $\Phi_{j,i}$ be the MA timeline corresponding to this subsequence. Now let $\Sigma_j = \{\Phi_{j,i} \mid 1 \leq i \leq n\}$, and $\alpha_j$ be any interdigitation of $\Sigma_j$. We now take $I$ to be the union of all $\alpha_j$, for $1 \leq j \leq m$. We show that $I$ is an interdigitation of $\Sigma$. Since each state $s$ appearing in $\Sigma$ must co-occur with at least one state $s_j$ in $\Phi$ in at least one $W_i$, $s$ will be in at least one tuple of $\alpha_j$, and, hence, be in some tuple of $I$—so $I$ is piecewise total.

Now, define the restriction $I^{i,j}$ of $I$ to components $i$ and $j$, with $i < j$, to be the relation given by taking the set of all pairs formed by shortening tuples of $I$ by omitting all components except the $i$'th and the $j$'th. Likewise define $\alpha_k^{i,j}$ for each $k$. To show $I$ is an interdigitation, it now suffices to show that each $I^{i,j}$ is simultaneously consistent. Consider states $s_i$ and $s_j$ from timelines $\Phi_i$ and $\Phi_j$, respectively, such that $I^{i,j}(s_i, s_j)$. Suppose that $t_i$ occurs after $s_i$ in $\Phi_i$, and for some $t_j \in \Phi_j$, $I^{i,j}(t_i, t_j)$ holds. It suffices to show that $s_j$ is no later than $t_j$ in $\Phi_j$. Since $I^{i,j}(s_i, s_j)$ and $I^{i,j}(t_i, t_j)$, we must have $\alpha_k^{i,j}(s_i, s_j)$ and $\alpha_{k'}^{i,j}(t_i, t_j)$, respectively, for some $k$ and $k'$. We know $k \leq k'$ because $s_i$ is before $t_i$ in $\Phi_i$ and $W_i$ is simultaneously consistent. If $k = k'$, then $s_j$ is no later than $t_j$ in $\Phi_j$, because $\alpha_k$ must be simultaneously consistent, being an interdigitation. Otherwise, $k < k'$. Then $s_j$ is no later than $t_j$ in $\Phi_j$, as desired, because $W_j$ is simultaneously consistent. So $I$ is simultaneously consistent, and an interdigitation of $\Sigma$.

Let $\Phi'$ be the member of IG$(\Sigma)$ corresponding to $I$. We now show that $\Phi' \leq \Phi$. We know that each state $s' \in \Phi'$ is the intersection of the states in a tuple of some $\alpha_j$—we say that $s'$ derives from $\alpha_j$. Consider the interdigitation $I'$ between $\Phi$ and $\Phi'$, where $I'(s_j, s')$, for $s_j \in \Phi$ and $s' \in \Phi'$, if and only if $s'$ derives from $\alpha_j$. $I'$ is piecewise total, as every tuple of $I'$ derives from some $\alpha_j$, and no $\alpha_j$ is empty. $I'$ is simultaneously consistent because tuples of $I'$ deriving from later $\alpha_k$ must be later in the lexicographic ordering of $I$, given the simultaneous consistency of the $W_k$ interdigitations used to construct each $\alpha_j$. Finally, we know that $s_j$ subsumes (i.e., is a subset of) each state in each tuple of $\alpha_j$, because each $W_k$ is a witnessing interdigitation to $\Phi_k \leq \Phi$, and, hence, subsumes (is a subset of) the intersection of those states. Therefore, if $s_j \in \Phi$ co-occurs with $s' \in \Phi'$ in $I'$ we have that $s' \leq s_j$. Thus, $I'$ is a witnessing interdigitation for $\Phi' \leq \Phi$, and by Proposition 2 we have $\Phi' \leq \Phi$.

The size bound on $\Phi'$ follows, since, as pointed out in the main text, the size of any member of IG$(\Sigma)$ is upper-bounded by the number of states in $\Sigma$. □

**Lemma 8.** *Given MA timelines $\Phi_1 = s_1; \ldots; s_m$ and $\Phi_2 = t_1; \ldots; t_n$, there is a witnessing interdigitation for $\Phi_1 \leq \Phi_2$ iff there is a path in the subsumption graph $SG(\Phi_1, \Phi_2)$ from $v_{1,1}$ to $v_{m,n}$.*

**Proof**: Subsumption graph $SG(\Phi_1, \Phi_2)$ is equal to $\langle V, E\rangle$ with $V = \{v_{i,j} \mid 1 \leq i \leq m, 1 \leq j \leq n\}$ and $E = \{\langle v_{i,j}, v_{i',j'}\rangle \mid s_i \leq t_j,\ s_{i'} \leq t_{j'},\ i \leq i' \leq i+1, j \leq j' \leq j+1\}$. Note that there is a correspondence between vertices and state tuples—with vertex $v_{i,j}$ corresponding to $\langle s_i, t_j\rangle$.

For the forward direction, assume that $W$ is a witnessing interdigitation for $\Phi_1 \leq \Phi_2$. We know that, if the states $s_i$ and $t_j$ co-occur in $W$, then $s_i \leq t_j$ since $W$ witnesses $\Phi_1 \leq \Phi_2$. The vertices corresponding to the tuples of $W$ will be called co-occurrence vertices, and satisfy the first condition for belonging to some edge in $E$ (that $s_i \leq t_j$). It follows from the definition of interdigitation that both $v_{1,1}$ and $v_{m,n}$ are both co-occurrence vertices. Consider a co-occurrence vertex $v_{i,j}$ not equal to $v_{m,n}$, and the lexicographically least co-occurrence vertex $v_{i',j'}$ after $v_{i,j}$ (ordering vertices by ordering the pair of subscripts). We show that $i$, $j$, $i'$, and $j'$ satisfy the requirements for $\langle v_{i,j}, v_{i',j'}\rangle \in E$. If not, then either $i' > i+1$ or $j' > j+1$. If $i' > i+1$, then there can be no co-occurrence vertex $v_{i+1,j''}$, contradicting that $W$ is piecewise total. If $j' > j+1$, then since $W$ is piecewise total, there must be a co-occurrence vertex $v_{i'',j+1}$: but if $i'' < i$ or $i'' > i'$, this contradicts the simultaneous consistency of $W$, and if $i'' = i$, this contradicts the lexicographically least choice of $v_{i',j'}$. It follows that every co-occurrence vertex but $v_{m,n}$ has an edge to another co-occurrence vertex closer in Manhattan distance to $v_{m,n}$, and thus that there is a path from $v_{1,1}$ to $v_{m,n}$.

For the reverse direction assume there is a path of vertices in $SG(\Phi_1, \Phi_2)$ from $v_{1,1}$ to $v_{m,n}$ given by, $v_{i_1,j_1}, v_{i_2,j_2}, \ldots, v_{i_r,j_s}$ with $i_1 = j_1 = 1$, $i_r = m, j_s = n$. Let $W$ be the set of state tuples corresponding to the vertices along this path. $W$ must be simultaneously consistent with the $\Phi_i$ orderings because our directed edges are all non-decreasing in the $\Phi_i$ orderings. $W$ must be piecewise total because no edge can cross more than one state transition in either $\Phi_1$ or $\Phi_2$, by the edge set definition. So $W$ is an interdigitation. Finally, the definition of the edge set $E$ ensures that each tuple $\langle s_i, t_j \rangle$ in $W$ has the property $s_i \leq t_j$, so that $W$ is a witnessing interdigitation for $\Phi_1 \leq \Phi_2$, showing that $\Phi_1 \leq \Phi_2$, as desired. □

**Lemma 10.** *Given some $n$, let $\Psi$ be the conjunction of the timelines*

$$\bigcup_{i=1}^{n} \{(\mathit{PROP}_n; \mathit{True}_i; \mathit{False}_i; \mathit{PROP}_n), (\mathit{PROP}_n; \mathit{False}_i; \mathit{True}_i; \mathit{PROP}_n)\}.$$

*We have the following facts about truth assignments to the Boolean variables $p_1, \ldots, p_n$:*

1. *For any truth assignment $A$, $\mathit{PROP}_n; s_A; \mathit{PROP}_n$ is semantically equivalent to a member of* $\mathrm{IS}(\Psi)$.

2. *For each $\Phi \in \mathrm{IS}(\Psi)$ there is a truth assignment $A$ such that $\Phi \leq \mathit{PROP}_n; s_A; \mathit{PROP}_n$.*

**Proof**: To prove the first part of the lemma, we construct an interdigitation $I$ of $\Psi$ such that the corresponding member of $\mathrm{IS}(\Psi)$ is equivalent to $\mathrm{PROP}_n; s_A; \mathrm{PROP}_n$. Intuitively, we construct $I$ by ensuring that some tuple of $I$ consists only of states of the form $\mathrm{True}_k$ or $\mathrm{False}_k$ that agree with the truth assignment—the union of all the states in this tuple, taken by $\mathrm{IS}(\Psi)$ will equal $s_A$. Let $I = \{T_0, T_1, T_2, T_3, T_4\}$ be an interdigitation of $\Psi$ with exactly five state tuples $T_i$. We assign the states of each timeline of $\Psi$ to the tuples as follows:

1. For any $k$, such that $1 \leq k \leq n$ and $A(p_k)$ is true,

   - for the timeline $s_1; s_2; s_3; s_4 = Q; True_k; False_k; Q$, assign each state $s_i$ to tuple $T_i$, and assign state $s_1$ to $T_0$ as well, and
   - for the timeline $s'_1; s'_2; s'_3; s'_4 = Q; False_k; True_k; Q$, assign each state $s'_i$ to tuple $T_{i-1}$, and state $s'_4$ to tuple $T_4$ as well.

2. For any $k$, such that $1 \leq k \leq n$ and $A(p_k)$ is false, assign states to tuples as in item 1 while interchanging the roles of $True_k$ and $False_k$.

It should be clear that $I$ is piecewise total and simultaneously consistent with the state orderings in $\Psi$, and so is an interdigitation. The union of the states in each of $T_0$, $T_1$, $T_3$, and $T_4$ is equal to $\mathrm{PROP}_n$, since $\mathrm{PROP}_n$ is included as a state in each of those tuples. Furthermore, we see that the union of the states in $T_2$ is equal to $s_A$. Thus, the member of $\mathrm{IS}(\Psi)$ corresponding to $I$ is equal to $\mathrm{PROP}_n; \mathrm{PROP}_n; s_A; \mathrm{PROP}_n; \mathrm{PROP}_n$, which is semantically equivalent to $\mathrm{PROP}_n; s_A; \mathrm{PROP}_n$, as desired.

To prove the second part of the lemma, let $\Phi$ be any member of $\mathrm{IS}(\Psi)$. We first argue that every state in $\Phi$ must contain either $\mathrm{True}_k$ or $\mathrm{False}_k$ for each $1 \leq k \leq n$. For any $k$, since $\Psi$ contains $\mathrm{PROP}_n; \mathrm{True}_k; \mathrm{False}_k; \mathrm{PROP}_n$, every member of $\mathrm{IS}(\Psi)$ must be subsumed by $\mathrm{PROP}_n; \mathrm{True}_k; \mathrm{False}_k; \mathrm{PROP}_n$. So, $\Phi$ is subsumed by $\mathrm{PROP}_n; \mathrm{True}_k; \mathrm{False}_k; \mathrm{PROP}_n$. But every state in $\mathrm{PROP}_n; \mathrm{True}_k; \mathrm{False}_k; \mathrm{PROP}_n$ contains either $\mathrm{True}_k$ or $\mathrm{False}_k$, implying that so does $\Phi$, as desired.

Next, we claim that for each $1 \le k \le n$, either $\Phi \le \text{True}_k$ or $\Phi \le \text{False}_k$—i.e., either all states in $\Phi$ include $\text{True}_k$, or all states in $\Phi$ include $\text{False}_k$ (and possibly both). To prove this claim, assume, for the sake of contradiction, that, for some $k$, $\Phi \not\le \text{True}_k$ and $\Phi \not\le \text{False}_k$. Combining this assumption with our first claim, we see there must be states $s$ and $s'$ in $\Phi$ such that $s$ contains $True_k$ but not $False_k$, and $s'$ contains $False_k$ but not $True_k$, respectively. Consider the interdigitation $I$ of $\Psi$ that corresponds to $\Phi$ as a member of $\text{IS}(\Psi)$. We know that $s$ and $s'$ are each equal to the union of states in tuples $T$ and $T'$, respectively, of $I$. $T$ and $T'$ must each include one state from each timeline $s_1; s_2; s_3; s_4 = \text{PROP}_n; \text{True}_k; \text{False}_k; \text{PROP}_n$ and $s'_1; s'_2; s'_3; s'_4 = \text{PROP}_n; \text{False}_k; \text{True}_k; \text{PROP}_n$. Clearly, since $s$ does not include $\text{False}_k$, $T$ includes the states $s_1$ and $s'_2$, and likewise $T'$ includes the states $s_2$ and $s'_1$. It follows that $I$ is not simultaneously consistent with the state orderings in $s_1; s_2; s_3; s_4$ and $s'_1; s'_2; s'_3; s'_4$, contradicting our choice of $I$ as an interdigitation. This shows that either $\Phi \le \text{True}_k$ or $\Phi \le \text{False}_k$.

Define the truth assignment $A$ such that for all $1 \le k \le n$, $A(p_k)$ if and only if $\Phi \le \text{True}_k$. Since,for each $k$, $\Phi \le \text{True}_k$ or $\Phi \le \text{False}_k$, it follows that each state of $\Phi$ is subsumed by $s_A$. Furthermore, since $\Phi$ begins and ends with $\text{PROP}_n$, it is easy to give an interdigitation of $\Phi$ and $\text{PROP}_n; s_A; \text{PROP}_n$ that witnesses $\Phi \le \text{PROP}_n; s_A; \text{PROP}_n$. Thus, we have that $\Phi \le \text{PROP}_n; s_A; \text{PROP}_n$. □

**Lemma 16.** *Let $\Phi_1$ and $\Phi_2$ be as given on page 402, in the proof of Theorem 17, and let $\Psi = \bigwedge \text{IG}(\{\Phi_1, \Phi_2\})$. For any $\Psi'$ whose timelines are a subset of those in $\Psi$ that omits some square timeline, we have $\Psi < \Psi'$.*

**Proof**: Since the timelines in $\Psi'$ are a subset of the timelines in $\Psi$, we know that $\Psi \le \Psi'$. It remains to show that $\Psi' \not\le \Psi$. We show this by constructing a timeline that is covered by $\Psi'$, but not by $\Psi$.

Let $\Phi = s_1; s_2; \ldots; s_{2n-1}$ be a square timeline in $\Psi$ that is not included in $\Psi'$. Recall that each $s_i$ is a single proposition from the proposition set $P = \{p_{i,j} \mid 1 \le i \le n,\ 1 \le j \le n\}$, and that, for consecutive states $s_i$ and $s_{i+1}$, if $s_i = p_{i,j}$, then $s_{i+1}$ is either $p_{i+1,j}$ or $p_{i,j+1}$. Define a new timeline $\overline{\Phi} = \overline{s}_2; \overline{s}_3; \ldots; \overline{s}_{2n-2}$ with $\overline{s}_i = (P - s_i)$. We now show that $\overline{\Phi} \not\le \Phi$ (so that $\overline{\Phi} \not\le \Psi$), and that, for any $\Phi'$ in $\Psi - \{\Phi\}$, $\overline{\Phi} \le \Phi'$ (so that $\overline{\Phi} \le \Psi'$).

For the sake of contradiction, assume that $\overline{\Phi} \le \Phi$—then there must be a interdigitation $W$ witnessing $\overline{\Phi} \le \Phi$. We show by induction on $i$ that, for $i \ge 2$, $W(s_i, \overline{s}_j)$ implies $j > i$. For the base case, when $i = 2$, we know that $\overline{s}_2 \not\le s_2$, since $s_2 \not\subseteq \overline{s}_2$, and so $W(s_2, \overline{s}_2)$ is false, since $W$ witnesses subsumption. For the inductive case, assume the claim holds for all $i' < i$, and that $W(s_i, \overline{s}_j)$. We know that $\overline{s}_i \not\le s_i$, and thus $i \ne j$. Because $W$ is piecewise total, we must have $W(s_{i-1}, \overline{s}_{j'})$ for some $j'$, and, by the induction hypothesis, we must have $j' > i - 1$. Since $W$ is simultaneously consistent with the $s_k$ and $\overline{s}_{k'}$ state orderings, and $i - 1 < i$, we have $j' \le j$. It follows that $j > i$ as desired. Given this claim, we see that $s_{2n-2}$ cannot co-occur in $W$ with any state in $\overline{\Phi}$, contradicting the fact that $W$ is piecewise total. Thus we have that $\overline{\Phi} \not\le \Phi$.

Let $\Phi' = s'_1; \ldots; s'_m$ be any timeline in $\Psi - \{\Phi\}$, we now construct an interdigitation that witnesses $\overline{\Phi} \le \Phi'$. Note that while $\Phi$ is assumed to be square, $\Phi'$ need not be. Let $j$ be the smallest index where $s_j \ne s'_j$— since $s_1 = s'_1 = p_{1,1}$, and $\Phi \ne \Phi'$, we know that such a $j$ must exist, and is in the range $2 \le j \le m$. We use the index $j$ to guide our construction of an interdigitation. Let $W$ be an interdigitation of $\overline{\Phi}$ and $\Phi'$, with exactly the following co-occurring states (i.e., state tuples):

1. For $1 \le i \le j - 1$, $\overline{s}_{i+1}$ co-occurs with $s'_i$.

2. For $j \leq i \leq m$, $\overline{s}_j$ co-occurs with $s'_i$.
3. For $j+1 \leq i \leq 2n-2$, $\overline{s}_i$ co-occurs with $s'_m$.

It is easy to check that $W$ is both piecewise total and simultaneously consistent with the state orderings in $\overline{\Phi}$ and $\Phi$, and so is an interdigitation. We now show that $W$ witnesses $\overline{\Phi} \leq \Phi'$ by showing that all states in $\overline{\Phi}$ are subsumed by the states they co-occur with in $W$. For co-occurring states $\overline{s}_{i+1}$ and $s'_i$ corresponding to the first item above we have that $s'_i = s_i$—this implies that $s'_i$ is contained in $\overline{s}_{i+1}$, giving that $\overline{s}_{i+1} \leq s'_i$. Now consider co-occurring states $\overline{s}_j$ and $s'_i$ from the second item above. Since $\Phi$ is square, choose $k$ and $l$ so that $s_{j-1} = p_{k,l}$, we have that $s_j$ is either $p_{k+1,l}$ or $p_{k,l+1}$. In addition, since $s_{j-1} = s'_{j-1}$ we have that $s'_j$ is either $p_{k+1,l}, p_{k,l+1}$ or $p_{k+1,l+1}$ but that $s_j \neq s'_j$. In any of these cases, we find that no state in $\Phi'$ after $s'_j$ can equal $s_j$—this follows by noting that the proposition indices never decrease across the timeline $\Phi'$[16]. We therefore have that, for $i \geq j$, $\overline{s}_j \leq s'_i$. Finally, for co-occurring states $\overline{s}_i$ and $s'_m$ from item three above, we have $\overline{s}_i \leq s'_m$, since $s'_m = p_{n,n}$, which is in all states of $\overline{\Phi}$. Thus, we have shown that for all co-occurring states in $W$, the state from $\overline{\Phi}$ is subsumed by the co-occurring state in $\Phi'$. Therefore, $W$ witnesses $\overline{\Phi} \leq \Phi'$, which implies that $\overline{\Phi} \leq \Phi'$. □

**Lemma 26.** *For any model $\langle M, I\rangle \in \mathcal{M}$ and any $\Psi \in$ AMA$^-$, $\Psi$ covers $\langle M, I\rangle$ iff $F[\Psi]$ covers $T[\langle M, I\rangle]$.*

**Proof**: Recall that $\mathcal{M}$ is the set of models over propositions in the set $P = \{p_1, \ldots, p_n\}$ and that we assume AMA$^-$ uses only primitive propositions from $P$ (possibly negated). We also have the set of propositions $\overline{P} = \{\bar{p}_1, \ldots, \bar{p}_n\}$, and assume that formulas in AMA use only propositions in $P \cup \overline{P}$ and that $\overline{\mathcal{M}}$ is the set of models over $P \cup \overline{P}$, where for each $i$, exactly one of $p_i$ and $\bar{p}_i$ is true at any time. Note that $F[\Psi]$ is in AMA and that $T[\langle M, I\rangle]$ is in $\overline{\mathcal{M}}$. We prove the lemma via straightforward induction on the structure of $\Psi$—proving the result for literals, then for states, then for timelines, and finally for AMA$^-$ formulas.

To prove the result for literals, we consider two cases (the third case of **true** is trivial). First, $\Psi$ can be a single proposition $p_i$, so that $\Psi' = F[p_i] = p_i$. Consider any model $\langle M, I\rangle \in \mathcal{M}$ and let $\langle M', I\rangle = T[\langle M, I\rangle]$. The following relationships yield the desired result.

| | | | |
|---|---|---|---|
| $\Psi$ covers $\langle M, I\rangle$ | iff | for each $i \in I$, $M[i]$ assigns $p_i$ true | (by definition of satisfiability) |
| | iff | for each $i \in I$, $M'[i]$ assigns $p_i$ true | (by definition of $T$) |
| | iff | $\Psi' = p_i$ covers $T[\langle M, I\rangle]$ | (by definition of satisfiability) |

The second case is when $\Psi$ is a negated proposition $\neg\Diamond p_i$—here, we get that $\Psi' = \bar{p}_i$. Let $\langle M, I\rangle \in \mathcal{M}$ and $\langle M', I\rangle = T[\langle M, I\rangle]$. The following relationships yield the desired result.

| | | | |
|---|---|---|---|
| $\Psi$ covers $\langle M, I\rangle$ | iff | for each $i \in I$, $M[i]$ assigns $p_i$ false | (by definition of satisfiability) |
| | iff | for each $i \in I$, $M'[i]$ assigns $\bar{p}_i$ true | (by definition of $T$) |
| | iff | $\Psi' = \bar{p}_i$ covers $T[\langle M, I\rangle]$ | (by definition of satisfiability) |

This proves the lemma for literals.

16. Note that if $\Phi$ were not required to be square then it is possible for $s'_{j+1}$ to equal $s_j$—i.e., they could both equal $p_{k+1,l+1}$.

To prove the result for states, we use induction on the number $k$ of literals in a state. The base case is when $k = 1$ (the state is a single literal) and was proven above. Now assume that the lemma holds for states with $k$ or fewer literals and let $\Psi = l_1 \wedge \cdots \wedge l_{k+1}$ and $\langle M, I\rangle \in \mathcal{M}$. From the inductive assumption we know that $\Phi = l_1 \wedge \cdots \wedge l_k$ covers $\langle M, I\rangle$ iff $F[\Phi]$ covers $T[\langle M, I\rangle]$. From our base case we also know that $l_{k+1}$ covers $\langle M, I\rangle$ iff $F[l_{k+1}]$ covers $T[\langle M, I\rangle]$. From these facts and the definition of satisfiability for states, we get that $\Psi$ covers $\langle M, I\rangle$ iff $F[\Phi] \wedge F[l_{k+1}]$ covers $T[\langle M, I\rangle]$. Clearly $F$ has the property that $F[\Phi] \wedge F[l_{k+1}] = F[\Psi]$, showing that the lemma holds for states.

To prove the result for timelines, we use induction on the number $k$ of states in the timeline. The base case is when $k = 1$ (the timeline is a single state) and was proven above. Now assume that the lemma holds for timelines with $k$ or fewer states. Let $\Psi = s_1; \ldots ; s_{k+1}$ and $\langle M, [t, t']\rangle \in \mathcal{M}$ with $\langle M', [t, t']\rangle = T[\langle M, [t, t']\rangle]$. We have the following relationships.

| | | |
|---|---|---|
| $\Psi$ covers $\langle M, [t, t']\rangle$ | iff | there exists some $t'' \in [t, t']$, such that $s_1$ covers $\langle M, [t, t'']\rangle$ and $\Phi = s_2; \ldots ; s_{k+1}$ covers either $\langle M, [t'', t']\rangle$ or $\langle M, [t''+1, t']\rangle$ |
| | iff | there exists some $t'' \in [t, t']$, such that $F[s_1]$ covers $\langle M', [t, t'']\rangle$ and $F[\Phi]$ covers either $\langle M', [t'', t']\rangle$ or $\langle M', [t''+1, t']\rangle$ |
| | iff | $F[s_1]; F[\Phi]$ covers $\langle M', [t, t']\rangle$ |
| | iff | $F[\Psi]$ covers $\langle M', [t, t']\rangle$ |

Where the first iff follows from the definition of satisfiability; the second follows from our inductive hypothesis, our base case, and the fact that for $I \subseteq [t, t']$ we have $T[\langle M, I\rangle] = \langle M', I\rangle$; the third follows from the definition of satisfiability; and the fourth follows from the fact that $F[s_1]; F[\Phi] = F[\Psi]$.

Finally, we prove the result for AMA$^-$ formulas, by induction on the number $k$ of timelines in the formula. The base case is when $k = 1$ (the formula is a single timeline) and was proven above. Now assume that the lemma holds for AMA$^-$ formulas with with $k$ or fewer timelines and let $\Psi = \Phi_1 \wedge \cdots \wedge \Phi_{k+1}$ and $\langle M, I\rangle \in \mathcal{M}$. From the inductive assumption, we know that $\Psi' = \Phi_1 \wedge \cdots \wedge \Phi_k$ covers $\langle M, I\rangle$ iff $F[\Psi']$ covers $T[\langle M, I\rangle]$. From our base case, we also know that $\Phi_{k+1}$ covers $\langle M, I\rangle$ iff $F[\Phi_{k+1}]$ covers $T[\langle M, I\rangle]$. From these facts and the definition of satisfiability, we get that $\Psi$ covers $\langle M, I\rangle$ iff $F[\Psi'] \wedge F[\Phi_{k+1}]$ covers $T[\langle M, I\rangle]$. Clearly $F$ has the property that $F[\Psi'] \wedge F[\Phi_{k+1}] = F[\Psi]$, showing that the lemma holds for AMA$^-$ formulas. This completes the proof. □

## Appendix C. Hand-coded and Learned Definitions Used in Our Experiments

Below we give the two sets of hand-coded definitions, $\text{HD}_1$ and $\text{HD}_2$, used in our experimental evaluation. We also give a set of learned AMA event definitions for the same seven event types. The learned definitions correspond to the output of our $k$-AMA learning algorithm, given all available training examples (30 examples per event type), with $k = 3$ and $D = \text{BN}$. All the event definitions are written in event logic, where $\neg\Diamond p$ denotes the negation of proposition $p$.

$$
\textsc{PickUp}(x,y,z) \triangleq \left(\begin{array}{l}
\neg\Diamond x = y \wedge \neg\Diamond z = x \wedge \neg\Diamond z = y \wedge \\
\textsc{Supported}(y) \wedge \neg\Diamond\textsc{Attached}(x,z) \wedge \\
\left\{\begin{array}{l}
\left[\begin{array}{l}
\neg\Diamond\textsc{Attached}(x,y) \wedge \neg\Diamond\textsc{Supports}(x,y) \wedge \\
\textsc{Supports}(z,y) \wedge \\
\neg\Diamond\textsc{Supported}(x) \wedge \neg\Diamond\textsc{Attached}(y,z) \wedge \\
\neg\Diamond\textsc{Supports}(y,x) \wedge \neg\Diamond\textsc{Supports}(y,z) \wedge \\
\neg\Diamond\textsc{Supports}(x,z) \wedge \neg\Diamond\textsc{Supports}(z,x)
\end{array}\right]; \\
[\textsc{Attached}(x,y) \vee \textsc{Attached}(y,z)]\,; \\
\left[\begin{array}{l}
\textsc{Attached}(x,y) \wedge \textsc{Supports}(x,y) \wedge \\
\neg\Diamond\textsc{Supports}(z,y) \wedge \\
\neg\Diamond\textsc{Supported}(x) \wedge \neg\Diamond\textsc{Attached}(y,z) \wedge \\
\neg\Diamond\textsc{Supports}(y,x) \wedge \neg\Diamond\textsc{Supports}(y,z) \wedge \\
\neg\Diamond\textsc{Supports}(x,z) \wedge \neg\Diamond\textsc{Supports}(z,x)
\end{array}\right]
\end{array}\right\}
\end{array}\right)
$$

$$
\textsc{PutDown}(x,y,z) \triangleq \left(\begin{array}{l}
\neg\Diamond x = y \wedge \neg\Diamond z = x \wedge \neg\Diamond z = y \wedge \\
\textsc{Supported}(y) \wedge \neg\Diamond\textsc{Attached}(x,z) \wedge \\
\left\{\begin{array}{l}
\left[\begin{array}{l}
\textsc{Attached}(x,y) \wedge \textsc{Supports}(x,y) \wedge \\
\neg\Diamond\textsc{Supports}(z,y) \wedge \\
\neg\Diamond\textsc{Supported}(x) \wedge \neg\Diamond\textsc{Attached}(y,z) \wedge \\
\neg\Diamond\textsc{Supports}(y,x) \wedge \neg\Diamond\textsc{Supports}(y,z) \wedge \\
\neg\Diamond\textsc{Supports}(x,z) \wedge \neg\Diamond\textsc{Supports}(z,x)
\end{array}\right]; \\
[\textsc{Attached}(x,y) \vee \textsc{Attached}(y,z)]\,; \\
\left[\begin{array}{l}
\neg\Diamond\textsc{Attached}(x,y) \wedge \neg\Diamond\textsc{Supports}(x,y) \wedge \\
\textsc{Supports}(z,y) \wedge \\
\neg\Diamond\textsc{Supported}(x) \wedge \neg\Diamond\textsc{Attached}(y,z) \wedge \\
\neg\Diamond\textsc{Supports}(y,x) \wedge \neg\Diamond\textsc{Supports}(y,z) \wedge \\
\neg\Diamond\textsc{Supports}(x,z) \wedge \neg\Diamond\textsc{Supports}(z,x)
\end{array}\right]
\end{array}\right\}
\end{array}\right)
$$

$$
\textsc{Stack}(w,x,y,z) \triangleq \left[\begin{array}{l}
\neg\Diamond z = w \wedge \neg\Diamond z = x \wedge \neg\Diamond z = y \wedge \\
\textsc{PutDown}(w,x,y) \wedge \textsc{Supports}(z,y) \wedge \\
\neg\textsc{Attached}(z,y)
\end{array}\right]
$$

$$
\textsc{Unstack}(w,x,y,z) \triangleq \left[\begin{array}{l}
\neg\Diamond z = w \wedge \neg\Diamond z = x \wedge \neg\Diamond z = y \wedge \\
\textsc{PickUp}(w,x,y) \wedge \textsc{Supports}(z,y) \wedge \neg\textsc{Attached}(z,y)
\end{array}\right]
$$

$$
\textsc{Move}(w,x,y,z) \triangleq \neg\Diamond y = z \wedge [\textsc{PickUp}(w,x,y); \textsc{PutDown}(w,x,z)]
$$

$$
\textsc{Assemble}(w,x,y,z) \triangleq \textsc{PutDown}(w,y,z) \wedge_{\{<\}} \textsc{Stack}(w,x,y,z)
$$

$$
\textsc{Disassemble}(w,x,y,z) \triangleq \textsc{Unstack}(w,x,y,z) \wedge_{\{<\}} \textsc{PickUp}(x,y,z)
$$

Figure 12: The $HD_1$ event-logic definitions for all seven event types.

$$\textsc{PickUp}(x,y,z) \triangleq \left(\begin{array}{l} \neg\Diamond x=y \wedge \neg\Diamond z=x \wedge \neg\Diamond z=y \wedge \\ \textsc{Supported}(y) \wedge \neg\Diamond\textsc{Attached}(x,z) \wedge \\ \left\{\begin{array}{l} \left[\begin{array}{l} \neg\Diamond\textsc{Attached}(x,y) \wedge \neg\Diamond\textsc{Supports}(x,y) \wedge \\ \textsc{Supports}(z,y) \wedge \textsc{Contacts}(z,y) \wedge \\ \neg\Diamond\textsc{Supported}(x) \wedge \neg\Diamond\textsc{Attached}(y,z) \wedge \\ \neg\Diamond\textsc{Supports}(y,x) \wedge \neg\Diamond\textsc{Supports}(y,z) \wedge \\ \neg\Diamond\textsc{Supports}(x,z) \wedge \neg\Diamond\textsc{Supports}(z,x) \end{array}\right] \wedge_{\{<,\mathsf{m}\}} \\ \left[\begin{array}{l} \textsc{Attached}(x,y) \wedge \textsc{Supports}(x,y) \wedge \\ \neg\Diamond\textsc{Supports}(z,y) \wedge \\ \neg\Diamond\textsc{Supported}(x) \wedge \neg\Diamond\textsc{Attached}(y,z) \wedge \\ \neg\Diamond\textsc{Supports}(y,x) \wedge \neg\Diamond\textsc{Supports}(y,z) \wedge \\ \neg\Diamond\textsc{Supports}(x,z) \wedge \neg\Diamond\textsc{Supports}(z,x) \end{array}\right] \end{array}\right\} \end{array}\right)$$

$$\textsc{PutDown}(x,y,z) \triangleq \left(\begin{array}{l} \neg\Diamond x=y \wedge \neg\Diamond z=x \wedge \neg\Diamond z=y \wedge \\ \textsc{Supported}(y) \wedge \neg\Diamond\textsc{Attached}(x,z) \wedge \\ \left\{\begin{array}{l} \left[\begin{array}{l} \textsc{Attached}(x,y) \wedge \textsc{Supports}(x,y) \wedge \\ \neg\Diamond\textsc{Supports}(z,y) \wedge \\ \neg\Diamond\textsc{Supported}(x) \wedge \neg\Diamond\textsc{Attached}(y,z) \wedge \\ \neg\Diamond\textsc{Supports}(y,x) \wedge \neg\Diamond\textsc{Supports}(y,z) \wedge \\ \neg\Diamond\textsc{Supports}(x,z) \wedge \neg\Diamond\textsc{Supports}(z,x) \end{array}\right] \wedge_{\{<,\mathsf{m}\}} \\ \left[\begin{array}{l} \neg\Diamond\textsc{Attached}(x,y) \wedge \neg\Diamond\textsc{Supports}(x,y) \wedge \\ \textsc{Supports}(z,y) \wedge \textsc{Contacts}(z,y) \wedge \\ \neg\Diamond\textsc{Supported}(x) \wedge \neg\Diamond\textsc{Attached}(y,z) \wedge \\ \neg\Diamond\textsc{Supports}(y,x) \wedge \neg\Diamond\textsc{Supports}(y,z) \wedge \\ \neg\Diamond\textsc{Supports}(x,z) \wedge \neg\Diamond\textsc{Supports}(z,x) \end{array}\right] \end{array}\right\} \end{array}\right)$$

Figure 13: Part I of the $\mathrm{HD}_2$ event-logic definitions.

$$
\textsc{Stack}(w,x,y,z) \;\triangleq\; \left(\begin{array}{l}
\neg\Diamond w = x \wedge \neg\Diamond y = w \wedge \neg\Diamond y = x\wedge \\
\neg\Diamond z = w \wedge \neg\Diamond z = x \wedge \neg\Diamond z = y\wedge \\
\textsc{Supported}(x) \wedge \neg\Diamond\textsc{Attached}(w,y)\wedge \\
\left\{\begin{array}{l}
\left[\begin{array}{l}
\textsc{Attached}(w,x) \wedge \textsc{Supports}(w,x)\wedge \\
\neg\Diamond\textsc{Supports}(y,x)\wedge \\
\textsc{Supports}(z,y) \wedge \textsc{Contacts}(z,y)\wedge \\
\neg\Diamond\textsc{Attached}(z,y)\wedge \\
\neg\Diamond\textsc{Supported}(w) \wedge \neg\Diamond\textsc{Attached}(x,y)\wedge \\
\neg\Diamond\textsc{Supports}(x,w) \wedge \neg\Diamond\textsc{Supports}(x,y)\wedge \\
\neg\Diamond\textsc{Supports}(w,y) \wedge \neg\Diamond\textsc{Supports}(y,w)
\end{array}\right] \wedge_{\{<,\mathsf{m}\}} \\
\left[\begin{array}{l}
\neg\Diamond\textsc{Attached}(w,x) \wedge \neg\Diamond\textsc{Supports}(w,x)\wedge \\
\textsc{Supports}(y,x) \wedge \textsc{Contacts}(y,x)\wedge \\
\textsc{Supports}(z,y) \wedge \textsc{Contacts}(z,y)\wedge \\
\neg\Diamond\textsc{Attached}(z,y)\wedge \\
\neg\Diamond\textsc{Supported}(w) \wedge \neg\Diamond\textsc{Attached}(x,y)\wedge \\
\neg\Diamond\textsc{Supports}(x,w) \wedge \neg\Diamond\textsc{Supports}(x,y)\wedge \\
\neg\Diamond\textsc{Supports}(w,y) \wedge \neg\Diamond\textsc{Supports}(y,w)
\end{array}\right]
\end{array}\right\}
\end{array}\right)
$$

$$
\textsc{Unstack}(w,x,y,z) \;\triangleq\; \left(\begin{array}{l}
\neg\Diamond w = x \wedge \neg\Diamond y = w \wedge \neg\Diamond y = x\wedge \\
\neg\Diamond z = w \wedge \neg\Diamond z = x \wedge \neg\Diamond z = y\wedge \\
\textsc{Supported}(x) \wedge \neg\Diamond\textsc{Attached}(w,y)\wedge \\
\left\{\begin{array}{l}
\left[\begin{array}{l}
\neg\Diamond\textsc{Attached}(w,x) \wedge \neg\Diamond\textsc{Supports}(w,x)\wedge \\
\textsc{Supports}(y,x) \wedge \textsc{Contacts}(y,x)\wedge \\
\textsc{Supports}(z,y) \wedge \textsc{Contacts}(z,y)\wedge \\
\neg\Diamond\textsc{Attached}(z,y)\wedge \\
\neg\Diamond\textsc{Supported}(w) \wedge \neg\Diamond\textsc{Attached}(x,y)\wedge \\
\neg\Diamond\textsc{Supports}(x,w) \wedge \neg\Diamond\textsc{Supports}(x,y)\wedge \\
\neg\Diamond\textsc{Supports}(w,y) \wedge \neg\Diamond\textsc{Supports}(y,w)
\end{array}\right] \wedge_{\{<,\mathsf{m}\}} \\
\left[\begin{array}{l}
\textsc{Attached}(w,x) \wedge \textsc{Supports}(w,x)\wedge \\
\neg\Diamond\textsc{Supports}(y,x)\wedge \\
\textsc{Supports}(z,y) \wedge \textsc{Contacts}(z,y)\wedge \\
\neg\Diamond\textsc{Attached}(z,y)\wedge \\
\neg\Diamond\textsc{Supported}(w) \wedge \neg\Diamond\textsc{Attached}(x,y)\wedge \\
\neg\Diamond\textsc{Supports}(x,w) \wedge \neg\Diamond\textsc{Supports}(x,y)\wedge \\
\neg\Diamond\textsc{Supports}(w,y) \wedge \neg\Diamond\textsc{Supports}(y,w)
\end{array}\right]
\end{array}\right\}
\end{array}\right)
$$

$$
\textsc{Move}(w,x,y,z) \;\triangleq\; \neg\Diamond y = z \wedge [\textsc{PickUp}(w,x,y); \textsc{PutDown}(w,x,z)]
$$

$$
\textsc{Assemble}(w,x,y,z) \;\triangleq\; \textsc{PutDown}(w,y,z) \wedge_{\{<\}} \textsc{Stack}(w,x,y,z)
$$

$$
\textsc{Disassemble}(w,x,y,z) \;\triangleq\; \textsc{Unstack}(w,x,y,z) \wedge_{\{<\}} \textsc{PickUp}(x,y,z)
$$

Figure 14: Part II of the $HD_2$ event-logic definitions.

$$
\textsc{PickUp}(x,y,z) \;\triangleq\;
\left(
\begin{array}{l}
\left\{
\begin{array}{l}
\left[\begin{array}{l}
\textsc{Supported}(y) \wedge \textsc{Supports}(z,y) \wedge \\
\textsc{Contacts}(y,z) \wedge \neg\Diamond\textsc{Supports}(x,y) \wedge \\
\neg\Diamond\textsc{Attached}(x,y) \wedge \neg\Diamond\textsc{Attached}(y,z)
\end{array}\right] ; \\
\textsc{Supported}(y); \\
\left[\begin{array}{l}
\textsc{Supported}(y) \wedge \textsc{Supports}(x,y) \wedge \\
\textsc{Attached}(x,y) \wedge \neg\Diamond\textsc{Supports}(z,y) \wedge \\
\neg\Diamond\textsc{Contacts}(y,z) \wedge \neg\Diamond\textsc{Attached}(y,z)
\end{array}\right]
\end{array}
\right\} \wedge \\
\left\{
\begin{array}{l}
\textsc{Supported}(y); \\
\left[\begin{array}{l}
\textsc{Supported}(y) \wedge \textsc{Attached}(x,y) \wedge \\
\textsc{Attached}(y,z)
\end{array}\right] ; \\
[\textsc{Supported}(y) \wedge \textsc{Attached}(x,y)]
\end{array}
\right\} \wedge \\
\left\{
\begin{array}{l}
[\textsc{Supported}(y) \wedge \textsc{Contacts}(y,z)] ; \\
[\textsc{Supported}(y) \wedge \textsc{Attached}(y,z)] ; \\
[\textsc{Supported}(y) \wedge \textsc{Attached}(x,y)]
\end{array}
\right\} \wedge \\
\left\{
\begin{array}{l}
\left[\begin{array}{l}
\textsc{Supported}(y) \wedge \textsc{Supports}(z,y) \wedge \\
\textsc{Contacts}(y,z) \wedge \neg\Diamond\textsc{Supports}(x,y) \wedge \\
\neg\Diamond\textsc{Attached}(x,y) \wedge \neg\Diamond\textsc{Attached}(y,z)
\end{array}\right] ; \\
[\textsc{Supported}(y) \wedge \textsc{Supports}(z,y)] ; \\
[\textsc{Supported}(y) \wedge \textsc{Attached}(x,y)]
\end{array}
\right\} \wedge \\
\left\{
\begin{array}{l}
[\textsc{Supported}(y) \wedge \textsc{Supports}(z,y)] ; \\
[\textsc{Supported}(y) \wedge \textsc{Attached}(x,y)] ; \\
\left[\begin{array}{l}
\textsc{Supported}(y) \wedge \textsc{Supports}(x,y) \wedge \\
\textsc{Attached}(x,y) \wedge \neg\Diamond\textsc{Supports}(z,y) \wedge \\
\neg\Diamond\textsc{Contacts}(y,z) \wedge \neg\Diamond\textsc{Attached}(y,z)
\end{array}\right]
\end{array}
\right\}
\end{array}
\right)
$$

$$
\textsc{PutDown}(x,y,z) \;\triangleq\;
\left(
\begin{array}{l}
\left\{
\begin{array}{l}
\left[\begin{array}{l}
\textsc{Supported}(y) \wedge \textsc{Supports}(x,y) \wedge \textsc{Attached}(x,y) \wedge \\
\neg\Diamond\textsc{Supports}(z,y) \wedge \neg\Diamond\textsc{Contacts}(y,z) \wedge \\
\neg\Diamond\textsc{Attached}(y,z)
\end{array}\right] ; \\
\textsc{Supported}(y); \\
\left[\begin{array}{l}
\textsc{Supported}(y) \wedge \textsc{Supports}(z,y) \wedge \textsc{Contacts}(z,y) \wedge \\
\neg\Diamond\textsc{Supports}(x,y) \wedge \neg\Diamond\textsc{Attached}(x,y)
\end{array}\right]
\end{array}
\right\} \wedge \\
\left\{
\begin{array}{l}
[\textsc{Supported}(y) \wedge \textsc{Attached}(x,y)] ; \\
[\textsc{Supported}(y) \wedge \textsc{Attached}(x,y) \wedge \textsc{Attached}(y,z)] ; \\
\textsc{Supported}(y)
\end{array}
\right\}
\end{array}
\right)
$$

Figure 15: The learned 3-AMA definitions for $\textsc{PickUp}(x,y,z)$ and $\textsc{PutDown}(x,y,z)$.

$$
\left(\begin{array}{l}
\left\{\begin{array}{l}
\left[\begin{array}{l}\textsc{Supported}(y) \wedge \textsc{Attached}(w,x) \wedge \textsc{Supports}(z,y) \wedge \textsc{Contacts}(y,z) \wedge \\ \neg\Diamond\textsc{Supports}(x,y) \wedge \neg\Diamond\textsc{Supports}(y,x) \wedge \neg\Diamond\textsc{Contacts}(x,y) \wedge \neg\Diamond\textsc{Attached}(x,y)\end{array}\right]; \\
[\textsc{Supported}(y)]; \\
\left[\begin{array}{l}\textsc{Supported}(y) \wedge \textsc{Supported}(x) \wedge \textsc{Supports}(y,x) \wedge \textsc{Contacts}(x,y) \wedge \textsc{Contacts}(y,z) \wedge \\ \neg\Diamond\textsc{Supports}(x,y) \wedge \neg\Diamond\textsc{Attached}(w,x) \wedge \neg\Diamond\textsc{Attached}(x,y) \wedge \neg\Diamond\textsc{Attached}(y,z)\end{array}\right]
\end{array}\right\} \wedge \\
\left\{\begin{array}{l}
[\textsc{Supported}(y) \wedge \textsc{Attached}(w,x)]; \\
[\textsc{Supported}(y) \wedge \textsc{Attached}(x,y)]; \\
[\textsc{Supported}(y) \wedge \textsc{Supported}(x) \wedge \textsc{Supports}(y,x) \wedge \textsc{Contacts}(x,y)]
\end{array}\right\} \wedge \\
\left\{\begin{array}{l}
[\textsc{Supported}(y) \wedge \textsc{Attached}(w,x)]; \\
[\textsc{Supported}(y) \wedge \textsc{Supports}(x,y) \wedge \textsc{Attached}(w,x) \wedge \textsc{Attached}(x,y) \wedge \textsc{Attached}(y,z)]; \\
[\textsc{Supported}(y) \wedge \textsc{Supported}(x)\textsc{Supports}(y,x)]
\end{array}\right\} \wedge \\
\left\{\begin{array}{l}
[\textsc{Supported}(y) \wedge \textsc{Attached}(w,x)]; \\
[\textsc{Supported}(y) \wedge \textsc{Supported}(x) \wedge \textsc{Supports}(x,y) \wedge \textsc{Supports}(y,x) \wedge \textsc{Attached}(w,x)]; \\
[\textsc{Supported}(y) \wedge \textsc{Supported}(x) \wedge \textsc{Supports}(y,x)]
\end{array}\right\} \wedge \\
\left\{\begin{array}{l}
[\textsc{Supported}(y) \wedge \textsc{Attached}(w,x) \wedge \textsc{Supports}(z,y) \wedge \textsc{Contacts}(y,z)]; \\
[\textsc{Supported}(y) \wedge \textsc{Attached}(y,z)]; \\
[\textsc{Supported}(y) \wedge \textsc{Supported}(x) \wedge \textsc{Supports}(y,x) \wedge \textsc{Contacts}(y,z)]
\end{array}\right\} \wedge \\
\left\{\begin{array}{l}
[\textsc{Supported}(y) \wedge \textsc{Attached}(w,x) \wedge \textsc{Supports}(z,y) \wedge \textsc{Contacts}(y,z)]; \\
[\textsc{Supported}(y) \wedge \textsc{Attached}(w,x) \wedge \textsc{Attached}(y,z)]; \\
[\textsc{Supported}(y) \wedge \textsc{Supported}(x) \wedge \textsc{Supports}(y,x)]
\end{array}\right\} \wedge \\
\left\{\begin{array}{l}
\left[\begin{array}{l}\textsc{Supported}(y) \wedge \textsc{Attached}(w,x) \wedge \textsc{Supports}(z,y) \wedge \textsc{Contacts}(y,z) \wedge \\ \neg\Diamond\textsc{Supports}(x,y) \wedge \neg\Diamond\textsc{Supports}(y,x) \wedge \neg\Diamond\textsc{Contacts}(x,y) \wedge \neg\Diamond\textsc{Attached}(x,y)\end{array}\right]; \\
[\textsc{Supported}(y) \wedge \textsc{Attached}(w,x)]; \\
[\textsc{Supported}(y) \wedge \textsc{Supported}(x) \wedge \textsc{Supports}(y,x)]
\end{array}\right\} \wedge \\
\left\{\begin{array}{l}
[\textsc{Supported}(y) \wedge \textsc{Attached}(w,x)]; \\
[\textsc{Supported}(y) \wedge \textsc{Attached}(w,x) \wedge \textsc{Supports}(z,y) \wedge \textsc{Contacts}(y,z)]; \\
[\textsc{Supported}(y) \wedge \textsc{Supported}(x)]
\end{array}\right\} \wedge \\
\left\{\begin{array}{l}
[\textsc{Supported}(y) \wedge \textsc{Attached}(w,x)]; \\
[\textsc{Supported}(y) \wedge \textsc{Attached}(w,x) \wedge \textsc{Supports}(z,y) \wedge \textsc{Supported}(x)]; \\
[\textsc{Supported}(y) \wedge \textsc{Supported}(x)]
\end{array}\right\} \wedge \\
\left\{\begin{array}{l}
[\textsc{Supported}(y) \wedge \textsc{Attached}(w,x)]; \\
\left[\begin{array}{l}\textsc{Supported}(y) \wedge \textsc{Contacts}(y,z) \wedge \textsc{Supports}(z,y) \wedge \textsc{Supported}(x) \wedge \\ \neg\Diamond\textsc{Supports}(x,y) \wedge \neg\Diamond\textsc{Attached}(x,y)\end{array}\right]; \\
[\textsc{Supported}(y) \wedge \textsc{Supported}(x)]
\end{array}\right\} \wedge \\
\left\{\begin{array}{l}
\textsc{Supported}(y); \\
\left[\begin{array}{l}\textsc{Supported}(y) \wedge \textsc{Contacts}(y,z) \wedge \textsc{Supports}(z,y) \wedge \textsc{Supported}(x) \wedge \\ \neg\Diamond\textsc{Supports}(x,y) \wedge \neg\Diamond\textsc{Attached}(x,y) \wedge \neg\Diamond\textsc{Attached}(y,z)\end{array}\right]; \\
[\textsc{Supported}(y) \wedge \textsc{Supported}(x) \wedge \textsc{Supports}(y,x)]
\end{array}\right\} \wedge \\
\left\{\begin{array}{l}
[\textsc{Supported}(y) \wedge \textsc{Attached}(w,x)]; \\
[\textsc{Supported}(y) \wedge \textsc{Contacts}(y,z) \wedge \textsc{Supported}(x)]; \\
[\textsc{Supported}(y) \wedge \textsc{Supported}(x) \wedge \textsc{Supported}(y)x]
\end{array}\right\} \wedge \\
\left\{\begin{array}{l}
[\textsc{Supported}(y) \wedge \textsc{Attached}(w,x)]; \\
[\textsc{Supported}(y) \wedge \textsc{Supported}(x) \wedge \textsc{Supports}(y,x)]; \\
\left[\begin{array}{l}\textsc{Supported}(y) \wedge \textsc{Supported}(x) \wedge \textsc{Supports}(y,x) \wedge \textsc{Contacts}(x,y) \wedge \textsc{Contacts}(y,z) \wedge \\ \neg\Diamond\textsc{Supports}(x,y) \wedge \neg\Diamond\textsc{Attached}(w,x) \wedge \neg\Diamond\textsc{Attached}(x,y) \wedge \neg\Diamond\textsc{Attached}(y,z)\end{array}\right]
\end{array}\right\} \wedge \\
\left\{\begin{array}{l}
\textsc{Supported}(y); \\
\left[\begin{array}{l}\textsc{Supported}(y) \wedge \textsc{Supported}(x) \wedge \textsc{Supports}(y,x) \wedge \textsc{Supports}(z,y) \wedge \\ \textsc{Contacts}(x,y) \wedge \textsc{Contacts}(y,z)\end{array}\right]; \\
\left[\begin{array}{l}\textsc{Supported}(y) \wedge \textsc{Supported}(x) \wedge \textsc{Supports}(y,x) \wedge \textsc{Contacts}(x,y) \wedge \textsc{Contacts}(y,z) \wedge \\ \neg\Diamond\textsc{Supports}(x,y) \wedge \neg\Diamond\textsc{Attached}(w,x) \wedge \neg\Diamond\textsc{Attached}(x,y) \wedge \neg\Diamond\textsc{Attached}(y,z)\end{array}\right]
\end{array}\right\}
\end{array}\right)
$$

Figure 16: The learned 3-AMA definition for $\textsc{Stack}(w,x,y,z)$.

$$
\left(\begin{array}{l}
\left\{\begin{array}{l}
\left[\begin{array}{l}
\textsc{Supported}(x) \wedge \textsc{Supported}(y) \wedge \textsc{Supports}(y,x) \wedge \\
\textsc{Contacts}(x,y) \wedge \textsc{Contacts}(y,z) \wedge \neg\Diamond\textsc{Supports}(w,x) \wedge \\
\neg\Diamond\textsc{Supports}(x,y) \wedge \neg\Diamond\textsc{Attached}(w,x) \wedge \neg\Diamond\textsc{Attached}(x,y)
\end{array}\right]; \\
[\textsc{Supported}(x) \wedge \textsc{Supported}(y)]; \\
\left[\begin{array}{l}
\textsc{Supported}(x) \wedge \textsc{Supported}(y) \wedge \textsc{Attached}(w,x) \wedge \textsc{Supports}(z,y) \wedge \\
\textsc{Contacts}(y,z) \wedge \textsc{Attached}(w,x) \wedge \neg\Diamond\textsc{Supports}(x,y) \wedge \\
\neg\Diamond\textsc{Supports}(y,x) \wedge \neg\Diamond\textsc{Contacts}(x,y) \wedge \\
\neg\Diamond\textsc{Attached}(x,y) \wedge \neg\Diamond\textsc{Attached}(y,z)
\end{array}\right]
\end{array}\right\} \wedge \\
\left\{\begin{array}{l}
[\textsc{Supported}(x) \wedge \textsc{Supported}(y) \wedge \textsc{Supports}(y,x)]; \\
[\textsc{Supported}(x) \wedge \textsc{Supported}(y) \wedge \textsc{Attached}(w,x) \wedge \textsc{Attached}(y,z)]; \\
[\textsc{Supported}(x) \wedge \textsc{Supported}(y) \wedge \textsc{Attached}(w,x) \wedge \textsc{Contacts}(y,z)]
\end{array}\right\} \wedge \\
\left\{\begin{array}{l}
[\textsc{Supported}(x) \wedge \textsc{Supported}(y) \wedge \textsc{Supports}(y,x) \wedge \textsc{Contacts}(y,z)]; \\
[\textsc{Supported}(x) \wedge \textsc{Supported}(y) \wedge \textsc{Attached}(y,z)]; \\
[\textsc{Supported}(x) \wedge \textsc{Supported}(y) \wedge \textsc{Attached}(w,x) \wedge \textsc{Contacts}(y,z)]
\end{array}\right\} \wedge \\
\left\{\begin{array}{l}
[\textsc{Supported}(x) \wedge \textsc{Supported}(y) \wedge \textsc{Supports}(y,x) \wedge \textsc{Contacts}(x,y)]; \\
[\textsc{Supported}(x) \wedge \textsc{Supported}(y) \wedge \textsc{Supports}(y,x) \wedge \textsc{Attached}(x,y)]; \\
[\textsc{Supported}(x) \wedge \textsc{Supported}(y) \wedge \textsc{Attached}(w,x)]
\end{array}\right\} \wedge \\
\left\{\begin{array}{l}
[\textsc{Supported}(x) \wedge \textsc{Supported}(y) \wedge \textsc{Supports}(y,x)]; \\
[\textsc{Supported}(x) \wedge \textsc{Supported}(y) \wedge \textsc{Contacts}(y,z)]; \\
[\textsc{Supported}(x) \wedge \textsc{Supported}(y) \wedge \textsc{Attached}(w,x)]
\end{array}\right\} \wedge \\
\left\{\begin{array}{l}
[\textsc{Supported}(x) \wedge \textsc{Supported}(y) \wedge \textsc{Supports}(y,x)]; \\
[\textsc{Supported}(x) \wedge \textsc{Supported}(y) \wedge \textsc{Attached}(w,x)]; \\
\left[\begin{array}{l}
\textsc{Supported}(x) \wedge \textsc{Supported}(y) \wedge \textsc{Attached}(w,x) \wedge \textsc{Supports}(z,y) \wedge \\
\textsc{Contacts}(y,z) \wedge \textsc{Attached}(w,x) \wedge \neg\Diamond\textsc{Supports}(x,y) \wedge \\
\neg\Diamond\textsc{Supports}(y,x) \wedge \neg\Diamond\textsc{Contacts}(x,y) \wedge \\
\neg\Diamond\textsc{Attached}(x,y) \wedge \neg\Diamond\textsc{Attached}(y,z)
\end{array}\right]
\end{array}\right\} \wedge \\
\left\{\begin{array}{l}
\left[\begin{array}{l}
\textsc{Supported}(x) \wedge \textsc{Supported}(y) \wedge \textsc{Supports}(y,x) \wedge \\
\textsc{Contacts}(x,y) \wedge \textsc{Contacts}(y,z) \wedge \\
\neg\Diamond\textsc{Supports}(w,x) \wedge \neg\Diamond\textsc{Supports}(x,y) \wedge \\
\neg\Diamond\textsc{Attached}(w,x) \wedge \neg\Diamond\textsc{Attached}(x,y)
\end{array}\right]; \\
[\textsc{Supported}(x) \wedge \textsc{Supported}(y) \wedge \textsc{Supports}(y,x)]; \\
[\textsc{Supported}(x) \wedge \textsc{Supported}(y) \wedge \textsc{Attached}(w,x)]
\end{array}\right\} \wedge \\
\left\{\begin{array}{l}
[\textsc{Supported}(x) \wedge \textsc{Supported}(y) \wedge \textsc{Supports}(y,x) \wedge \textsc{Contacts}(y,z)]; \\
[\textsc{Supported}(x) \wedge \textsc{Supported}(y) \wedge \textsc{Supports}(y,x) \wedge \textsc{Attached}(y,z)]; \\
[\textsc{Supported}(x) \wedge \textsc{Supported}(y) \wedge \textsc{Attached}(w,x)]
\end{array}\right\} \wedge \\
\left\{\begin{array}{l}
[\textsc{Supported}(x) \wedge \textsc{Supported}(y) \wedge \textsc{Supports}(y,x)]; \\
\left[\begin{array}{l}
\textsc{Supported}(x) \wedge \textsc{Supported}(y) \wedge \textsc{Supports}(y,x) \wedge \textsc{Attached}(y,z) \wedge \\
\textsc{Supports}(x,y) \wedge \textsc{Attached}(w,x) \wedge \textsc{Attached}(x,y)
\end{array}\right]; \\
[\textsc{Supported}(x) \wedge \textsc{Supported}(y) \wedge \textsc{Attached}(w,x)]
\end{array}\right\} \wedge \\
\left\{\begin{array}{l}
[\textsc{Supported}(x) \wedge \textsc{Supported}(y)]; \\
[\textsc{Supported}(x) \wedge \textsc{Supported}(y) \wedge \textsc{Supports}(y,x) \wedge \textsc{Attached}(w,x)]; \\
[\textsc{Supported}(x) \wedge \textsc{Supported}(y) \wedge \textsc{Supports}(w,x) \wedge \textsc{Attached}(w,x)]
\end{array}\right\} \wedge \\
\left\{\begin{array}{l}
[\textsc{Supported}(x) \wedge \textsc{Supported}(y) \wedge \textsc{Supports}(y,x)]; \\
[\textsc{Supported}(x) \wedge \textsc{Supported}(y) \wedge \textsc{Supports}(w,x) \wedge \textsc{Attached}(w,x)]; \\
[\textsc{Supported}(x) \wedge \textsc{Supported}(y) \wedge \textsc{Attached}(w,x)]
\end{array}\right\} \wedge \\
\left\{\begin{array}{l}
[\textsc{Supported}(x) \wedge \textsc{Supported}(y) \wedge \textsc{Supports}(y,x)]; \\
\left[\begin{array}{l}
\textsc{Supported}(x) \wedge \textsc{Supported}(y) \wedge \textsc{Contacts}(y,z) \wedge \\
\neg\Diamond\textsc{Supports}(x,y) \wedge \neg\Diamond\textsc{Attached}(x,y) \wedge \neg\Diamond\textsc{Attached}(y,z)
\end{array}\right]; \\
[\textsc{Supported}(x) \wedge \textsc{Supported}(y)]
\end{array}\right\}
\end{array}\right)
$$

Figure 17: The learned 3-AMA definition for $\textsc{Unstack}(w,x,y,z)$.

$$
\left(
\begin{array}{l}
\left\{
\begin{array}{l}
\left[
\begin{array}{l}
\textsc{Supported}(x) \wedge \textsc{Supports}(y,x) \wedge \textsc{Contacts}(y,x) \wedge \\
\neg\Diamond\textsc{Supports}(w,x) \wedge \neg\Diamond\textsc{Supports}(z,x) \wedge \neg\Diamond\textsc{Contacts}(x,z) \wedge \\
\neg\Diamond\textsc{Attached}(w,x) \wedge \neg\Diamond\textsc{Attached}(y,x) \wedge \neg\Diamond\textsc{Attached}(x,z)
\end{array}
\right]; \\
\textsc{Supported}(x); \\
\left[
\begin{array}{l}
\textsc{Supported}(x) \wedge \textsc{Supports}(z,x) \wedge \textsc{Contacts}(x,z) \wedge \\
\neg\Diamond\textsc{Supports}(w,x) \wedge \neg\Diamond\textsc{Supports}(y,x) \wedge \neg\Diamond\textsc{Contacts}(y,x) \wedge \\
\neg\Diamond\textsc{Attached}(w,x) \wedge \neg\Diamond\textsc{Attached}(y,x) \wedge \neg\Diamond\textsc{Attached}(x,z)
\end{array}
\right]
\end{array}
\right\} \wedge \\
\left\{
\begin{array}{l}
[\textsc{Supported}(x) \wedge \textsc{Supports}(y,x)]\,; \\
[\textsc{Supported}(x) \wedge \textsc{Attached}(w,x)]\,; \\
\textsc{Supported}(x)
\end{array}
\right\} \wedge \\
\left\{
\begin{array}{l}
\textsc{Supported}(x); \\
[\textsc{Supported}(x) \wedge \textsc{Attached}(w,x) \wedge \textsc{Attached}(x,z)]\,; \\
\textsc{Supported}(x)
\end{array}
\right\} \wedge \\
\left\{
\begin{array}{l}
[\textsc{Supported}(x)]\,; \\
[\textsc{Supported}(x) \wedge \textsc{Attached}(x,z)]\,; \\
[\textsc{Supported}(x) \wedge \textsc{Contacts}(x,z)]
\end{array}
\right\} \wedge \\
\left\{
\begin{array}{l}
\textsc{Supported}(x); \\
[\textsc{Supported}(x) \wedge \textsc{Attached}(w,x) \wedge \textsc{Supports}(w,x)]\,; \\
\textsc{Supported}(x)
\end{array}
\right\} \wedge \\
\left\{
\begin{array}{l}
\textsc{Supported}(x); \\
[\textsc{Supported}(x) \wedge \textsc{Attached}(w,x) \wedge \textsc{Attached}(y,x)]\,; \\
\textsc{Supported}(x)
\end{array}
\right\} \wedge \\
\left\{
\begin{array}{l}
[\textsc{Supported}(x) \wedge \textsc{Contacts}(y,x)]\,; \\
[\textsc{Supported}(x) \wedge \textsc{Attached}(y,x)]\,; \\
\textsc{Supported}(x)
\end{array}
\right\}
\end{array}
\right)
$$

Figure 18: The learned 3-AMA definition for $\textsc{Move}(w,x,y,z)$.

$$
\left(
\begin{array}{l}
\left\{
\begin{array}{l}
\left[
\begin{array}{l}
\neg\Diamond\textsc{Supported}(x) \wedge \neg\Diamond\textsc{Supports}(z,y) \wedge \neg\Diamond\textsc{Supports}(y,x)\wedge \\
\neg\Diamond\textsc{Contacts}(x,y) \wedge \neg\Diamond\textsc{Contacts}(z,y)\wedge \\
\neg\Diamond\textsc{Attached}(w,x) \wedge \neg\Diamond\textsc{Attached}(z,y)
\end{array}
\right]; \\
\mathbf{true}; \\
\left[
\begin{array}{l}
\textsc{Supported}(x) \wedge \textsc{Supported}(y) \wedge \textsc{Supports}(z,y)\wedge \\
\textsc{Supports}(y,x) \wedge \textsc{Contacts}(x,y)\wedge \\
\textsc{Contacts}(z,y) \wedge \neg\Diamond\textsc{Attached}(w,y)
\end{array}
\right]
\end{array}
\right\} \wedge \\
\left\{
\begin{array}{l}
\left[
\begin{array}{l}
\neg\Diamond\textsc{Supported}(x) \wedge \neg\Diamond\textsc{Supports}(z,y) \wedge \neg\Diamond\textsc{Supports}(y,x)\wedge \\
\neg\Diamond\textsc{Contacts}(x,y) \wedge \neg\Diamond\textsc{Contacts}(z,y)\wedge \\
\neg\Diamond\textsc{Attached}(w,x) \wedge \neg\Diamond\textsc{Attached}(z,y)
\end{array}
\right]; \\
\textsc{Attached}(w,y); \\
\textsc{Supported}(y)
\end{array}
\right\} \wedge \\
\left\{
\begin{array}{l}
\mathbf{true}; \\
[\textsc{Supported}(y) \wedge \neg\Diamond\textsc{Attached}(w,x) \wedge \neg\Diamond\textsc{Attached}(z,y)]; \\
\textsc{Supported}(y)
\end{array}
\right\} \wedge \\
\left\{
\begin{array}{l}
\mathbf{true}; \\
[\textsc{Supported}(y) \wedge \textsc{Attached}(z,y)]; \\
[\textsc{Supported}(y) \wedge \textsc{Contacts}(z,y)]
\end{array}
\right\} \wedge \\
\left\{
\begin{array}{l}
\mathbf{true}; \\
[\textsc{Supported}(y) \wedge \textsc{Supports}(z,y)\textsc{Contacts}(z,y) \wedge \textsc{Attached}(w,x)]; \\
\textsc{Supported}(y)
\end{array}
\right\} \wedge \\
\left\{
\begin{array}{l}
\mathbf{true}; \\
[\textsc{Supported}(y) \wedge \textsc{Attached}(w,y)\textsc{Attached}(z,y)]; \\
\textsc{Supported}(y)
\end{array}
\right\}
\end{array}
\right)
$$

Figure 19: The learned 3-AMA definition for $\textsc{Assemble}(w,x,y,z)$.

$$
\left(\begin{array}{l}
\left\{\begin{array}{l}
\left[\begin{array}{l}
\textsc{Supported}(x) \wedge \textsc{Supported}(y) \wedge \textsc{Supports}(y,x) \wedge \textsc{Supports}(z,y) \wedge \\
\textsc{Contacts}(x,y) \wedge \textsc{Contacts}(z,y) \wedge \neg\Diamond\textsc{Supports}(w,x) \wedge \\
\neg\Diamond\textsc{Supports}(w,y) \wedge \neg\Diamond\textsc{Supports}(x,y) \wedge \neg\Diamond\textsc{Attached}(x,w) \wedge \\
\neg\Diamond\textsc{Attached}(w,y) \wedge \neg\Diamond\textsc{Attached}(x,y) \wedge \neg\Diamond\textsc{Attached}(z,y)
\end{array}\right]; \\
\textsc{Supported}(y); \\
\left[\begin{array}{l}
\textsc{Supported}(y) \wedge \neg\Diamond\textsc{Supported}(x) \wedge \neg\Diamond\textsc{Supports}(w,x) \wedge \\
\neg\Diamond\textsc{Supports}(z,y) \wedge \neg\Diamond\textsc{Supports}(y,x) \wedge \neg\Diamond\textsc{Contacts}(x,y) \wedge \\
\neg\Diamond\textsc{Contacts}(z,y) \wedge \neg\Diamond\textsc{Attached}(x,w) \wedge \neg\Diamond\textsc{Attached}(z,y)
\end{array}\right];
\end{array}\right\} \wedge \\
\left\{\begin{array}{l}
[\textsc{Supported}(x) \wedge \textsc{Supported}(y)]; \\
\left[\begin{array}{l}
\textsc{Supported}(x) \wedge \textsc{Supported}(y) \wedge \textsc{Supports}(w,x) \wedge \\
\textsc{Supports}(z,y) \wedge \textsc{Contacts}(z,y) \wedge \textsc{Attached}(x,w)
\end{array}\right]; \\
\textsc{Supported}(y)
\end{array}\right\} \wedge \\
\left\{\begin{array}{l}
\left[\begin{array}{l}
\textsc{Supported}(x) \wedge \textsc{Supported}(y) \wedge \textsc{Supports}(z,y) \wedge \\
\textsc{Supports}(y,x) \wedge \textsc{Contacts}(x,y) \wedge \textsc{Contacts}(z,y)
\end{array}\right]; \\
[\textsc{Supported}(x) \wedge \textsc{Supported}(y) \wedge \textsc{Supports}(y,x) \wedge \textsc{Attached}(x,y)]; \\
\textsc{Supported}(y)
\end{array}\right\} \wedge \\
\left\{\begin{array}{l}
[\textsc{Supported}(x) \wedge \textsc{Supported}(y) \wedge \textsc{Supports}(y,x) \wedge \textsc{Contacts}(z,y)]; \\
\left[\begin{array}{l}
\textsc{Supported}(x) \wedge \textsc{Supported}(y) \wedge \textsc{Supports}(x,y) \wedge \\
\textsc{Supports}(y,z) \wedge \textsc{Attached}(x,y) \wedge \textsc{Attached}(z,y)
\end{array}\right]; \\
\textsc{Supported}(y)
\end{array}\right\} \wedge \\
\left\{\begin{array}{l}
[\textsc{Supported}(x) \wedge \textsc{Supported}(y) \wedge \textsc{Supports}(y,x)]; \\
\left[\begin{array}{l}
\textsc{Supported}(x) \wedge \textsc{Supported}(y) \wedge \textsc{Supports}(x,y) \wedge \\
\textsc{Supports}(y,z) \wedge \textsc{Attached}(x,y) \wedge \textsc{Attached}(z,y) \wedge \textsc{Attached}(x,w)
\end{array}\right]; \\
\textsc{Supported}(y)
\end{array}\right\} \wedge \\
\left\{\begin{array}{l}
\textsc{Supported}(y); \\
[\textsc{Supported}(y) \wedge \textsc{Attached}(w,y) \wedge \textsc{Attached}(z,y)]; \\
\textsc{Supported}(y)
\end{array}\right\} \wedge \\
\left\{\begin{array}{l}
\textsc{Supported}(y); \\
[\textsc{Supported}(y) \wedge \textsc{Supports}(w,y) \wedge \textsc{Attached}(w,y)]; \\
\textsc{Supported}(y)
\end{array}\right\}
\end{array}\right)
$$

Figure 20: The learned 3-AMA definition for $\textsc{Disassemble}(w,x,y,z)$.